\documentclass{book}
\usepackage{roboto}

\usepackage[english]{babel}
\usepackage[letterpaper,top=2cm,bottom=2cm,left=3cm,right=3cm,marginparwidth=1.75cm]{geometry}

\usepackage{pdflscape}
\usepackage{rotating}
\usepackage{longtable}
\usepackage{array}
\usepackage{float}
\usepackage{amsmath}
\usepackage{graphicx}
\usepackage{inconsolata}
\usepackage[colorlinks=true, allcolors=blue]{hyperref}
\usepackage{enumitem}
\usepackage{tikz} % TikZ package from first code
\usepackage{listings}
\usepackage{xcolor}
\usepackage[T1]{fontenc}
\usepackage{IEEEtrantools}  % IEEE package for formatting author blocks
\usepackage{epigraph}  % For block quotes

\definecolor{bgcolor}{rgb}{0.97,0.97,0.97}
\definecolor{codeblue}{rgb}{0.1,0.1,0.8}
\definecolor{codegreen}{rgb}{0,0.4,0}
\definecolor{codegray}{rgb}{0.4,0.4,0.4}
\definecolor{codepurple}{rgb}{0.5,0,0.5}
\definecolor{codered}{rgb}{0.6,0.2,0.2}
\definecolor{lightgray}{rgb}{0.9,0.9,0.9}
\definecolor{darkgray}{rgb}{0.6,0.6,0.6} % Darker gray for Python frames

\makeatletter
\renewcommand{\paragraph}{%
  \@startsection{paragraph}{4}{\z@}{1ex}{-1em}{\normalfont\normalsize\bfseries\color{gray}}}
\makeatother

\lstdefinestyle{python}{
    language=Python,
    basicstyle=\ttfamily\small\color{black}\usefont{T1}{zi4}{m}{n},  % Inconsolata for code
    keywordstyle=\bfseries\color{codeblue},  % Bold keywords
    stringstyle=\color{codegreen},  % Strings in green
    commentstyle=\slshape\color{codegray},  % Comments in gray and slanted (not italics)
    showstringspaces=false,
    numbers=left,
    numberstyle=\tiny\color{codegray},  % Line numbers in tiny gray
    stepnumber=1,
    numbersep=8pt,
    frame=single,
    rulecolor=\color{darkgray},  % Darker frame for Python code
    breaklines=true,
    backgroundcolor=\color{bgcolor},
    tabsize=4,
    captionpos=b,
    morekeywords={self}, % Add more keywords if needed
}

\lstdefinestyle{cmd}{
    language=bash,
    basicstyle=\ttfamily\small\color{black}\usefont{T1}{zi4}{m}{n},  % Inconsolata
    keywordstyle=\bfseries\color{blue},
    stringstyle=\color{codegreen},
    commentstyle=\itshape\color{gray},
    showstringspaces=false,
    numbers=none,
    frame=single,
    rulecolor=\color{darkgray},  % Darker frame for cmd
    breaklines=true,
    backgroundcolor=\color{bgcolor},
    tabsize=4,
    captionpos=b,
}

\title{Deep Learning and Machine Learning, Advancing Big Data Analytics and Management: Tensorflow Pretrained Models}

\author{
    Keyu Chen\textsuperscript{*} \\ 
    \textit{Georgia Institute of Technology} \\
    kchen637@gatech.edu
    \and
    Ziqian Bi\textsuperscript{*} \\
    \textit{Indiana University} \\
    bizi@iu.edu
    \and
    Qian Niu \\ 
    \textit{Kyoto University} \\
    niu.qian.f44@kyoto-u.jp
    \and   
    Junyu Liu \\ 
    \textit{Kyoto University} \\
    liu.junyu.82w@st.kyoto-u.ac.jp
    \and
    Benji Peng \\ 
    \textit{AppCubic} \\
    benji@appcubic.com
    \and
    Sen Zhang \\ 
    \textit{Rutgers University} \\
    sen.z@rutgers.edu
    \and
    Ming Liu \\ 
    \textit{Purdue University} \\
    liu3183@purdue.edu
    \and
    Xinyuan Song \\
    \textit{Emory University} \\
    xsong30@emory.edu
    \and
    Zekun Jiang \\
    \textit{Sichuan University} \\
    zekun\_jiang@163.com
    \and
    Tianyang Wang \\ 
    \textit{Xi’an Jiaotong-Liverpool University} \\
    Tianyang.Wang21@student.xjtlu.edu.cn
    \and 
    Ming Li \\ 
    \textit{Georgia Institute of Technology} \\
    mli694@gatech.edu
    \and
    Xuanhe Pan \\ 
    \textit{University of Wisconsin-Madison} \\
    xpan73@wisc.edu
    \and
    Jiawei Xu \\ 
    \textit{Purdue University} \\
    xu1644@purdue.edu
    \and
    Jinlang Wang \\ 
    \textit{University of Wisconsin-Madison} \\
    jinlang.wang@wisc.edu
    \and
    Pohsun Feng\textsuperscript{$\dagger$} \\
    \textit{National Taiwan Normal University} \\
    41075018h@ntnu.edu.tw
}

\date{} % Remove the date

\begin{document}

\maketitle

% Add footnotes for co-first author and corresponding author
\begingroup
\renewcommand\thefootnote{}\footnote{
    \textsuperscript{*} Equal contribution \\
    \textsuperscript{$\dagger$} Corresponding author
}
\addtocounter{footnote}{0}
\endgroup

% Insert the quote right after the title and authors
\epigraph{"It may be that our role on this planet is not to worship God but to create him."}{\textit{Arthur C. Clarke}}

\tableofcontents  % Table of contents

% Add your sections/chapters here
\setcounter{part}{2} % 从3开始标号
\part{Advancing Your Skills}

\setcounter{chapter}{40} % 从41开始标号
\chapter{TensorFlow pre-trained models}

\section{What is TensorFlow (TF) for Deep Learning}

\subsection{Introduction to TensorFlow in Deep Learning}
TensorFlow (TF) is an open-source platform developed by Google for machine learning and deep learning applications. It provides a flexible architecture for building machine learning models, especially neural networks, making it an essential tool for both beginners and experienced practitioners in deep learning. TensorFlow allows developers to perform computations efficiently on both CPUs and GPUs, supporting high-performance machine learning applications \cite{DBLP:journals/corr/AbadiBCCDDDGIIK16}.

At its core, TensorFlow enables the creation of computational graphs, which represent mathematical operations in the form of a directed graph. These graphs make it easier to visualize and optimize the performance of deep learning models. TensorFlow can handle large datasets and perform complex operations with its optimized execution engine.

\subsection{TensorFlow Architecture for Deep Learning}
TensorFlow's architecture is designed to provide flexibility and scalability. The core components of its architecture include:

\begin{enumerate}
    \item \textbf{TensorFlow Core}: The foundation of TensorFlow, providing low-level API functions for handling tensors and operations. It allows full control over model design and execution.
    \item \textbf{Tensors}: The fundamental data structure in TensorFlow, representing n-dimensional arrays. Tensors can be scalars (0-D), vectors (1-D), matrices (2-D), or higher-dimensional objects.
    \item \textbf{Graph}: TensorFlow uses computational graphs to represent operations. Nodes in the graph represent operations (like addition or multiplication), and the edges between them represent tensors being passed as inputs and outputs.
    \item \textbf{Session}: To execute a computational graph, TensorFlow uses sessions. A session manages the resources and execution of operations within the graph.
    \item \textbf{Eager Execution}: While TensorFlow traditionally relied on constructing computational graphs and then executing them, TensorFlow now supports eager execution. This mode allows operations to be executed immediately, simplifying debugging and interaction.
    \item \textbf{Estimators and Keras}: TensorFlow provides high-level APIs like Estimators and Keras to simplify the process of building and training deep learning models.
\end{enumerate}

\subsection{Key Features of TensorFlow for Pretrained Models}
TensorFlow provides several key features that make it a powerful tool for working with pretrained models:

\begin{itemize}
    \item \textbf{Model Zoo}: TensorFlow offers access to a large collection of pretrained models in the TensorFlow Model Garden, ranging from image classification, object detection, to natural language processing.
    \item \textbf{TensorFlow Hub}: TensorFlow Hub is a repository of reusable machine learning modules that can be easily integrated into new models. These modules include pretrained models and can be fine-tuned for specific tasks.
    \item \textbf{Transfer Learning}: TensorFlow supports transfer learning, a technique where you can take a pretrained model and adapt it to a new, related task by retraining only certain layers. This reduces the need for large amounts of labeled data and computational resources.
    \item \textbf{TensorFlow Serving}: TensorFlow Serving is designed for serving machine learning models in production environments. It provides a flexible and efficient system to serve trained models in real-time.
    \item \textbf{Model Optimization}: TensorFlow has built-in tools for optimizing pretrained models, such as quantization and pruning, which can reduce the model size and improve performance without sacrificing accuracy.
\end{itemize}

\subsection{TensorFlow in Pretrained Model Workflows}
TensorFlow plays a critical role in workflows that utilize pretrained models, enabling developers to streamline their processes. Here are the steps typically involved in using TensorFlow with pretrained models:

\begin{enumerate}
    \item \textbf{Model Selection}: Begin by selecting a pretrained model from TensorFlow Hub or the TensorFlow Model Garden. These models have been trained on large datasets and can be fine-tuned for specific tasks.
    
    \item \textbf{Loading the Model}: TensorFlow makes it easy to load pretrained models using its high-level API. For example, in TensorFlow Hub, you can load a model using the following Python code:

    \begin{lstlisting}[style=python]
    import tensorflow_hub as hub
    
    # Load a pretrained model from TensorFlow Hub
    model = hub.KerasLayer("https://tfhub.dev/google/imagenet/mobilenet_v2_100_224/classification/5")
    \end{lstlisting}

    \item \textbf{Preprocessing Data}: Pretrained models often require specific input formats, so it’s important to preprocess your data accordingly. TensorFlow provides many tools to help with this, such as the \texttt{tf.image} module for image data manipulation.
    
    \item \textbf{Fine-tuning the Model}: You can modify the pretrained model by adding new layers or freezing some of the existing layers. The following code snippet demonstrates how to add new layers to a pretrained model in TensorFlow:

    \begin{lstlisting}[style=python]
    base_model = hub.KerasLayer("https://tfhub.dev/google/imagenet/mobilenet_v2_100_224/classification/5", trainable=False)
    
    # Add new layers on top of the base model
    model = tf.keras.Sequential([
        base_model,
        tf.keras.layers.Dense(128, activation='relu'),
        tf.keras.layers.Dense(10, activation='softmax')
    ])
    \end{lstlisting}
    
    \item \textbf{Training the Model}: Once your model is set up, you can train it using TensorFlow’s high-level Keras API. The training process typically involves specifying the optimizer, loss function, and evaluation metrics.
    
    \begin{lstlisting}[style=python]
    model.compile(optimizer='adam', loss='sparse_categorical_crossentropy', metrics=['accuracy'])
    
    # Train the model
    model.fit(train_data, train_labels, epochs=5)
    \end{lstlisting}

    \item \textbf{Model Evaluation}: After training, the model can be evaluated on a test dataset to measure its performance.
    
    \begin{lstlisting}[style=python]
    # Evaluate the model on the test dataset
    test_loss, test_acc = model.evaluate(test_data, test_labels)
    print(f"Test accuracy: {test_acc}")
    \end{lstlisting}

    \item \textbf{Deployment}: Once the model is trained and evaluated, it can be deployed using TensorFlow Serving or exported as a SavedModel format for future use.
\end{enumerate}

\section{What is a Pretrained Model}

\subsection{Definition of Pretrained Models}
A pretrained model is a machine learning model that has been previously trained on a large dataset, typically using a task that is related to the problem you are trying to solve. This training helps the model to learn general patterns, which can then be fine-tuned or adapted to new, specific tasks. Instead of training a model from scratch, you can leverage the knowledge that the pretrained model has already gained and apply it to your own problem, saving both time and computational resources.

Pretrained models are especially popular in deep learning, particularly in areas like computer vision and natural language processing (NLP), where large datasets and substantial computational power are required to train complex models like convolutional neural networks (CNNs) or transformers.

\subsection{Advantages of Pretrained Models}
Pretrained models offer several advantages, especially for beginners and those with limited computational resources. Some of the key benefits include:

\begin{itemize}
    \item \textbf{Faster development:} Since the model has already learned many of the basic patterns, you don't need to spend as much time training it from scratch. You can fine-tune the model on your specific task, which is generally much faster.
    \item \textbf{Better performance with limited data:} In many cases, you might not have enough data to train a deep learning model from scratch. Pretrained models are helpful because they have already been trained on large datasets and can generalize well even with smaller datasets.
    \item \textbf{Reduced computational cost:} Training deep learning models from scratch can be very computationally expensive. Pretrained models allow you to leverage the power of complex models without the need for high-end hardware or long training times.
    \item \textbf{Access to state-of-the-art techniques:} Many pretrained models are based on cutting-edge research and have been fine-tuned to achieve high accuracy in a variety of tasks. By using these models, you can implement state-of-the-art solutions without needing deep expertise in model design or training.
\end{itemize}

\subsection{Common Use Cases of Pretrained Models}
Pretrained models are used in a wide range of applications across different fields. Some common use cases include:

\begin{itemize}
    \item \textbf{Image classification:} Pretrained models like ResNet, VGG, or MobileNet are commonly used for classifying images into different categories. These models have been trained on large image datasets like ImageNet.
    \item \textbf{Object detection:} Models such as YOLO (You Only Look Once) or Faster R-CNN are used for identifying and locating objects within an image.
    \item \textbf{Natural Language Processing (NLP):} Pretrained models such as BERT (Bidirectional Encoder Representations from Transformers) and GPT (Generative Pretrained Transformer) are used for tasks like text classification, sentiment analysis, and language translation.
    \item \textbf{Transfer learning:} Pretrained models are often used in transfer learning, where the knowledge gained from one task (e.g., image classification) is transferred to a different but related task (e.g., object detection).
    \item \textbf{Feature extraction:} Pretrained models are also used as feature extractors, where the learned features of the model are used as inputs for other models or algorithms.
\end{itemize}

\subsection{Pretrained Models in TensorFlow}
TensorFlow provides easy access to a wide range of pretrained models, which can be used for tasks such as image classification, object detection, and text analysis. You can load these models from the TensorFlow Hub, a repository of pretrained models that are ready to use.

Here's an example of how you can load and use a pretrained model for image classification in TensorFlow:

\begin{lstlisting}[style=python]
import tensorflow as tf
import tensorflow_hub as hub
import numpy as np
from PIL import Image

# Load a pretrained MobileNetV2 model from TensorFlow Hub
model = hub.KerasLayer("https://tfhub.dev/google/tf2-preview/mobilenet_v2/classification/4")

# Load and preprocess an image
image = Image.open("example_image.jpg")
image = image.resize((224, 224))
image = np.array(image) / 255.0  # Normalize the image
image = image[np.newaxis, ...]

# Make a prediction
predictions = model(image)
predicted_class = np.argmax(predictions)

print("Predicted class:", predicted_class)
\end{lstlisting}

In this example, we load a pretrained MobileNetV2 model from TensorFlow Hub and use it to classify an image. The image is first preprocessed by resizing it to the required input size and normalizing the pixel values. Finally, the model makes a prediction, and we output the predicted class.

By using TensorFlow Hub and pretrained models, you can quickly get started with complex machine learning tasks, even if you're a beginner.

\section{How to Use Pretrained Models}
Pretrained models are models that have already been trained on a large dataset, often for a similar task. Instead of training a new model from scratch, we can leverage these pretrained models for tasks such as image classification, object detection, natural language processing, etc. This approach saves both time and computational resources.

In this section, we will cover three main methods for utilizing pretrained models: Transfer Learning, Linear Probe, and Fine-Tuning.

    \subsection{Transfer Learning}
    Transfer learning is a machine learning technique where a model trained on one task is reused on a different, but related task. It allows us to leverage the knowledge a model has acquired from a large dataset to apply it to a smaller dataset or a new task. This reduces the need for large amounts of data and decreases the training time.

        \subsubsection{Feature Extraction}
        Feature extraction refers to using the pretrained model to extract useful features from the input data, and then using these features in a new model. Typically, only the top layers of the pretrained model (the feature extraction layers) are used, while a new classifier is trained on top of them.

        \begin{lstlisting}[style=python]
        from tensorflow.keras.applications import VGG16
        from tensorflow.keras.models import Model
        from tensorflow.keras.layers import Dense, Flatten

        # Load the pretrained model (e.g., VGG16) without the classifier layers
        base_model = VGG16(weights='imagenet', include_top=False, input_shape=(224, 224, 3))

        # Freeze the base model
        base_model.trainable = False

        # Add new classifier layers on top
        model = Model(inputs=base_model.input, 
                      outputs=Dense(10, activation='softmax')(Flatten()(base_model.output)))

        # Compile the model
        model.compile(optimizer='adam', loss='categorical_crossentropy', metrics=['accuracy'])

        # Train the model on your dataset
        model.fit(train_data, train_labels, epochs=10)
        \end{lstlisting}

        \subsubsection{Using Pretrained Weights}
        In many cases, pretrained models come with weights that have been learned on large datasets such as ImageNet. These weights can be used directly in new models to enhance performance. Here’s an example of how to load pretrained weights.

        \begin{lstlisting}[style=python]
        from tensorflow.keras.applications import ResNet50

        # Load the pretrained ResNet50 model with ImageNet weights
        model = ResNet50(weights='imagenet')

        # Use the model for prediction
        predictions = model.predict(new_data)
        \end{lstlisting}
        
    \subsection{Linear Probe}
    A Linear Probe is a lightweight approach to transfer learning where only the final classifier layer is trained, while all other layers are frozen. This allows for very fast training and serves as a good baseline to determine whether transfer learning is a viable approach for your task.

        \subsubsection{Training Only the Classifier Layer}
        In this case, all of the layers except for the final classifier are kept frozen during training. The classifier layer is initialized randomly and trained on your specific dataset.

        \begin{lstlisting}[style=python]
        from tensorflow.keras.applications import MobileNetV2
        from tensorflow.keras.layers import Dense
        from tensorflow.keras.models import Model

        # Load pretrained MobileNetV2 model
        base_model = MobileNetV2(weights='imagenet', include_top=False, pooling='avg')

        # Freeze all layers
        base_model.trainable = False

        # Add a new classifier layer
        classifier = Dense(10, activation='softmax')(base_model.output)

        # Create the model
        model = Model(inputs=base_model.input, outputs=classifier)

        # Compile the model
        model.compile(optimizer='adam', loss='categorical_crossentropy', metrics=['accuracy'])

        # Train only the classifier layer
        model.fit(train_data, train_labels, epochs=5)
        \end{lstlisting}
        
        \subsubsection{Advantages and Disadvantages of Linear Probing}
        The main advantage of using a linear probe is its simplicity and speed. Since we only train the final layer, the training process is fast and computationally inexpensive. However, it may not be as powerful as fine-tuning the entire model, especially when the pretrained model is not closely related to the target task.

\textbf{Linear Probe: Advantages and Disadvantages}

\begin{itemize}
    \item \textbf{Advantages}
    \begin{itemize}
        \item Faster Training
        \item Less Computationally Expensive
    \end{itemize}
    \item \textbf{Disadvantages}
    \begin{itemize}
        \item May Not Generalize Well to New Tasks
        \item Less Accurate on Complex Datasets
    \end{itemize}
\end{itemize}

    \subsection{Fine-Tuning}
    Fine-tuning is a more advanced transfer learning technique where you unfreeze some or all of the layers in the pretrained model and retrain them on the new dataset. This allows the model to adapt more fully to the new task, improving accuracy, especially if the pretrained model’s dataset is not very similar to the new one.

        \subsubsection{Fine-Tuning the Entire Model}
        Fine-tuning the entire model involves unfreezing all layers of the pretrained model and training the entire model on the new data. This allows the weights in all layers to adjust to the specifics of the new task.

        \begin{lstlisting}[style=python]
        # Unfreeze all layers in the base model
        base_model.trainable = True

        # Recompile the model (necessary after unfreezing layers)
        model.compile(optimizer='adam', loss='categorical_crossentropy', metrics=['accuracy'])

        # Fine-tune the entire model on your data
        model.fit(train_data, train_labels, epochs=10)
        \end{lstlisting}

        \subsubsection{Fine-Tuning Specific Layers}
        Sometimes it’s beneficial to fine-tune only certain layers of the model, especially deeper layers. Early layers often capture general features (such as edges in images) that are useful across a variety of tasks, while deeper layers capture more task-specific features.

        \begin{lstlisting}[style=python]
        # Unfreeze specific layers (e.g., the last few layers)
        for layer in base_model.layers[:-5]:
            layer.trainable = False

        # Recompile the model
        model.compile(optimizer='adam', loss='categorical_crossentropy', metrics=['accuracy'])

        # Fine-tune selected layers
        model.fit(train_data, train_labels, epochs=10)
        \end{lstlisting}

        \subsubsection{When to Use Fine-Tuning}
        Fine-tuning is most useful when:
        \begin{itemize}
            \item Your new dataset is large and somewhat different from the dataset used to pretrain the model.
            \item You need the model to be highly specialized for your new task.
            \item You have the computational resources and time to fine-tune multiple layers.
        \end{itemize}

\section{Dataset}

In this experiment, we use two classic datasets: CIFAR-10 and ImageNet.

\subsection{CIFAR-10}

CIFAR-10 is a widely used small image dataset containing 10 classes, with 6,000 images per class, resulting in a total of 60,000 images. The images are of size 32x32 pixels and are colored. The dataset includes categories such as airplanes, cars, birds, cats, deer, dogs, frogs, horses, ships, and trucks. Due to its small size and low image resolution, CIFAR-10 is well-suited for rapid experimentation and prototyping. In this experiment, we use CIFAR-10 as an example to demonstrate the process of training and testing models \cite{krizhevsky2009learning}.

\subsection{ImageNet}

ImageNet is a large-scale image dataset that spans over 1,000 classes and contains millions of high-resolution images. It plays a central role in the field of computer vision and is widely used for tasks such as image classification, object detection, and feature extraction. Models pre-trained on ImageNet have a broad understanding of the world due to the diversity of the images, making them useful for transfer learning and feature extraction tasks \cite{5206848}.

\subsection{Why CIFAR-10 and ImageNet?}

We chose CIFAR-10 as the demonstration dataset primarily because it is small and manageable. This makes it ideal for quick iteration and verifying the feasibility of algorithms in resource-constrained environments. For beginners and researchers, CIFAR-10 serves as a very practical platform for experimentation \cite{krizhevsky2009learning}.

On the other hand, ImageNet is used due to its vast scale and diverse range of images, making it a common choice for pre-training models. Models pre-trained on ImageNet provide rich, generalizable features that significantly enhance downstream tasks. In this experiment, we utilize ImageNet pre-trained models to extract image features, reducing training cost and improving model generalization \cite{5206848}.

\section{Comparison Between Linear Probe and Fine-tuning}

In this section, we will compare two approaches for adapting deep learning models to new data: \textbf{Linear Probe} and \textbf{Fine-tuning}. We will use the CIFAR-10 dataset and a pre-trained ResNet-152 model to illustrate the differences between these methods through visualization techniques such as PCA, t-SNE, and UMAP.

\subsection{Introduction to Dimensionality Reduction}

When working with high-dimensional data, visualizing patterns and relationships can be challenging. \textbf{Dimensionality reduction} techniques help simplify data by reducing the number of features while preserving important information.

Common methods include:

\begin{itemize}
    \item \textbf{Principal Component Analysis (PCA)}: Projects data onto directions that maximize variance.
    \item \textbf{t-Distributed Stochastic Neighbor Embedding (t-SNE)}: Converts similarities between data points into probabilities to preserve local structures.
    \item \textbf{Uniform Manifold Approximation and Projection (UMAP)}: Preserves both local and global data structures, often faster than t-SNE.
\end{itemize}

These techniques allow us to visualize complex datasets in two or three dimensions, making them easier to interpret.

\subsection{Linear Probe on 20\% of CIFAR-10 Features Extracted by ResNet-152}

In the \textbf{Linear Probe} approach, we use a pre-trained model to extract features from the dataset without updating the model's weights. We then analyze these features to understand how well the pre-trained model represents our data \cite{alain2018understandingintermediatelayersusing}.

\subsubsection{Installing Required Libraries}

Before running the code, install the necessary libraries:

\begin{lstlisting}[style=cmd]
pip install tensorflow matplotlib scikit-learn umap-learn
\end{lstlisting}

\subsubsection{Python Code for Feature Extraction and Visualization}

The following code performs feature extraction using a pre-trained ResNet-152 model in TensorFlow and visualizes the features using PCA, t-SNE, and UMAP \cite{he2015deepresiduallearningimage, DBLP:journals/corr/AbadiBCCDDDGIIK16, MACKIEWICZ1993303, JMLR:v9:vandermaaten08a, lel2018umap}.

\begin{lstlisting}[style=python]
import tensorflow as tf
import numpy as np
import matplotlib.pyplot as plt
from tensorflow.keras.applications import ResNet152
from tensorflow.keras.applications.resnet import preprocess_input
from tensorflow.keras.preprocessing.image import ImageDataGenerator
from sklearn.decomposition import PCA
from sklearn.manifold import TSNE
import umap
import os

# Set random seed for reproducibility
np.random.seed(42)
tf.random.set_seed(42)

# Create directory to save visualizations
os.makedirs('visualizations', exist_ok=True)

# Load CIFAR-10 dataset
(x_train_full, y_train_full), _ = tf.keras.datasets.cifar10.load_data()
y_train_full = y_train_full.flatten()

# Normalize images and resize to match ResNet input size
def preprocess_images(images):
    images_resized = tf.image.resize(images, [224, 224])
    return preprocess_input(images_resized)

# Split dataset: 20% for Linear Probe, 80% for Fine-tuning
num_samples = x_train_full.shape[0]
indices = np.arange(num_samples)
np.random.shuffle(indices)

split_idx = int(0.2 * num_samples)
linear_probe_idx = indices[:split_idx]
fine_tune_idx = indices[split_idx:]

x_linear_probe = x_train_full[linear_probe_idx]
y_linear_probe = y_train_full[linear_probe_idx]

x_fine_tune = x_train_full[fine_tune_idx]
y_fine_tune = y_train_full[fine_tune_idx]

# Preprocess images
x_linear_probe_preprocessed = preprocess_images(x_linear_probe)

# Load pre-trained ResNet-152 model without the top classification layer
base_model = ResNet152(weights='imagenet', include_top=False, pooling='avg')

# Extract features
features_linear_probe = base_model.predict(x_linear_probe_preprocessed, batch_size=32, verbose=1)

# Visualization function
def visualize(features, labels, method='pca', title='', filename=''):
    if method == 'pca':
        reducer = PCA(n_components=2)
    elif method == 'tsne':
        reducer = TSNE(n_components=2, random_state=42)
    elif method == 'umap':
        reducer = umap.UMAP(n_components=2, random_state=42)

    reduced_features = reducer.fit_transform(features)

    plt.figure(figsize=(10, 8))
    scatter = plt.scatter(reduced_features[:, 0], reduced_features[:, 1], c=labels, cmap='tab10', alpha=0.7)
    plt.legend(*scatter.legend_elements(), title="Classes")
    plt.title(title)
    plt.savefig(f'visualizations/{filename}.png', dpi=200)
    plt.close()

# Visualize and save Linear Probe results
visualize(features_linear_probe, y_linear_probe, method='pca', title='PCA - Linear Probe', filename='PCA_Linear_Probe')
visualize(features_linear_probe, y_linear_probe, method='tsne', title='t-SNE - Linear Probe', filename='tSNE_Linear_Probe')
visualize(features_linear_probe, y_linear_probe, method='umap', title='UMAP - Linear Probe', filename='UMAP_Linear_Probe')
\end{lstlisting}

\subsubsection{Understanding the Code}

In the code above:

\begin{itemize}
    \item We load and preprocess the CIFAR-10 dataset to match the input size of ResNet-152 (224x224 pixels) and apply the necessary preprocessing function.
    \item The dataset is split into 20\% for feature extraction (Linear Probe) and 80\% for fine-tuning.
    \item We load the pre-trained ResNet-152 model without the top classification layer to use it as a feature extractor.
    \item Features are extracted by passing the preprocessed images through the base model.
    \item The \texttt{visualize} function reduces the features to two dimensions using PCA, t-SNE, or UMAP and saves the plots with the specified filenames.
\end{itemize}

\subsubsection{PCA Visualization}

PCA reduces the dimensionality of the data by projecting it onto directions (principal components) that maximize variance \cite{MACKIEWICZ1993303}.

\begin{figure}[H]
    \centering
    \includegraphics[width=0.8\textwidth]{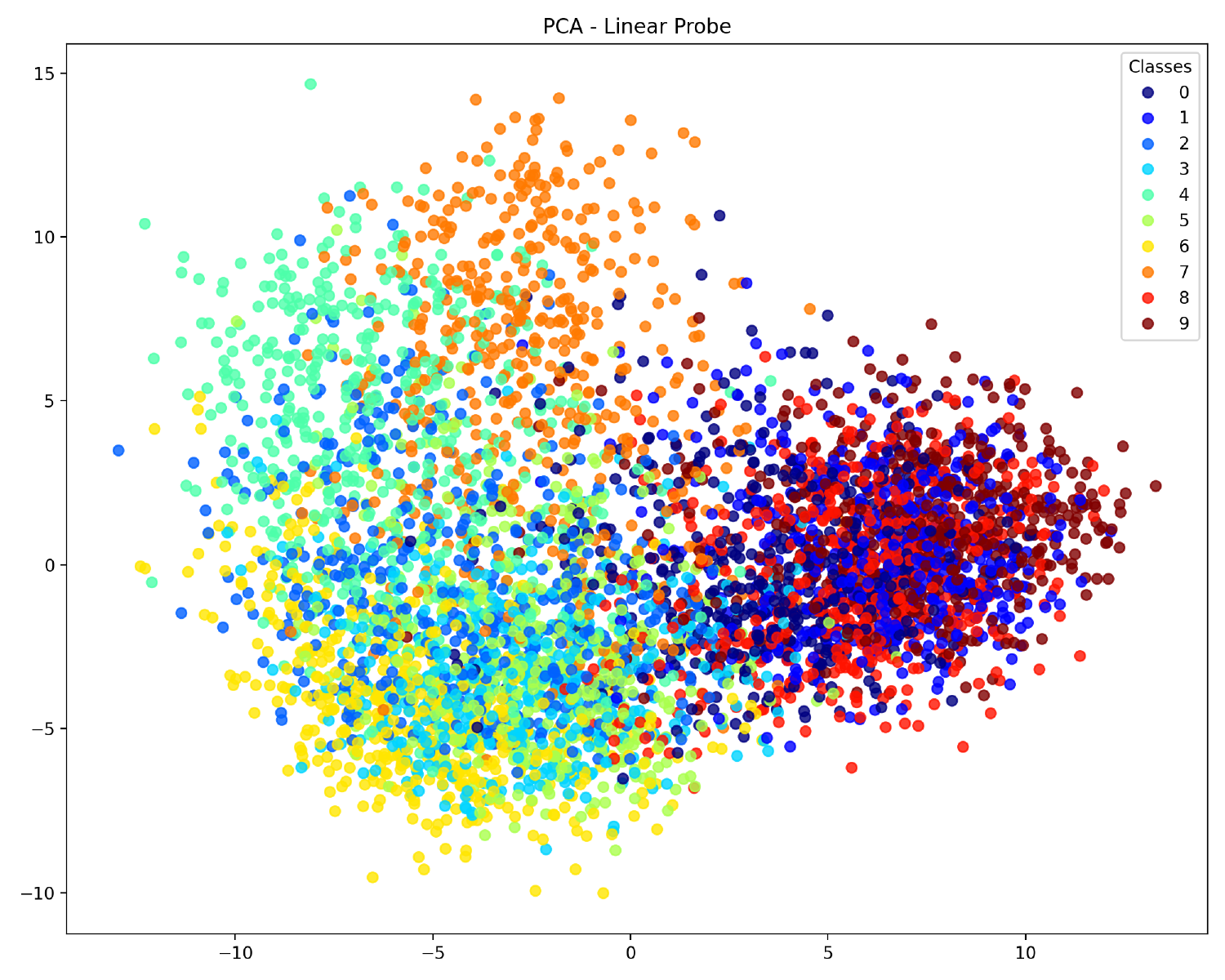}
    \caption{PCA visualization of features from 20\% CIFAR-10 (Linear Probe)}
    \label{fig:pca_linear_probe}
\end{figure}

In Figure~\ref{fig:pca_linear_probe}, we observe how the pre-trained model's features are distributed across different classes after applying PCA.

\subsubsection{t-SNE Visualization}

t-SNE is a non-linear technique that preserves local structures and is particularly good at visualizing clusters \cite{JMLR:v9:vandermaaten08a}.

\begin{figure}[H]
    \centering
    \includegraphics[width=0.8\textwidth]{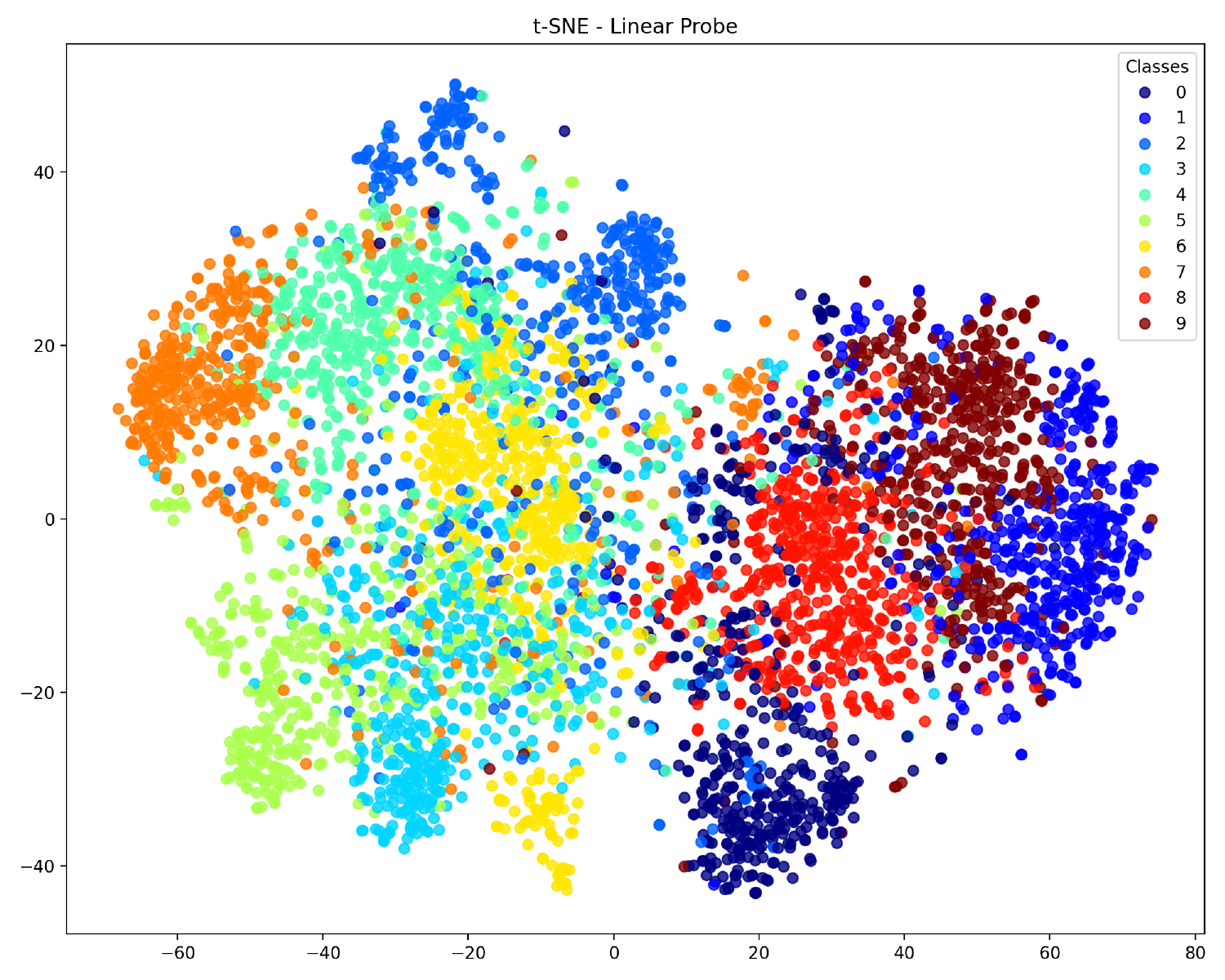}
    \caption{t-SNE visualization of features from 20\% CIFAR-10 (Linear Probe)}
    \label{fig:tsne_linear_probe}
\end{figure}

Figure~\ref{fig:tsne_linear_probe} shows the t-SNE visualization, where clusters corresponding to different classes can be observed.

\subsubsection{UMAP Visualization}

UMAP aims to preserve both local and global structures of the data \cite{lel2018umap}.

\begin{figure}[H]
    \centering
    \includegraphics[width=0.8\textwidth]{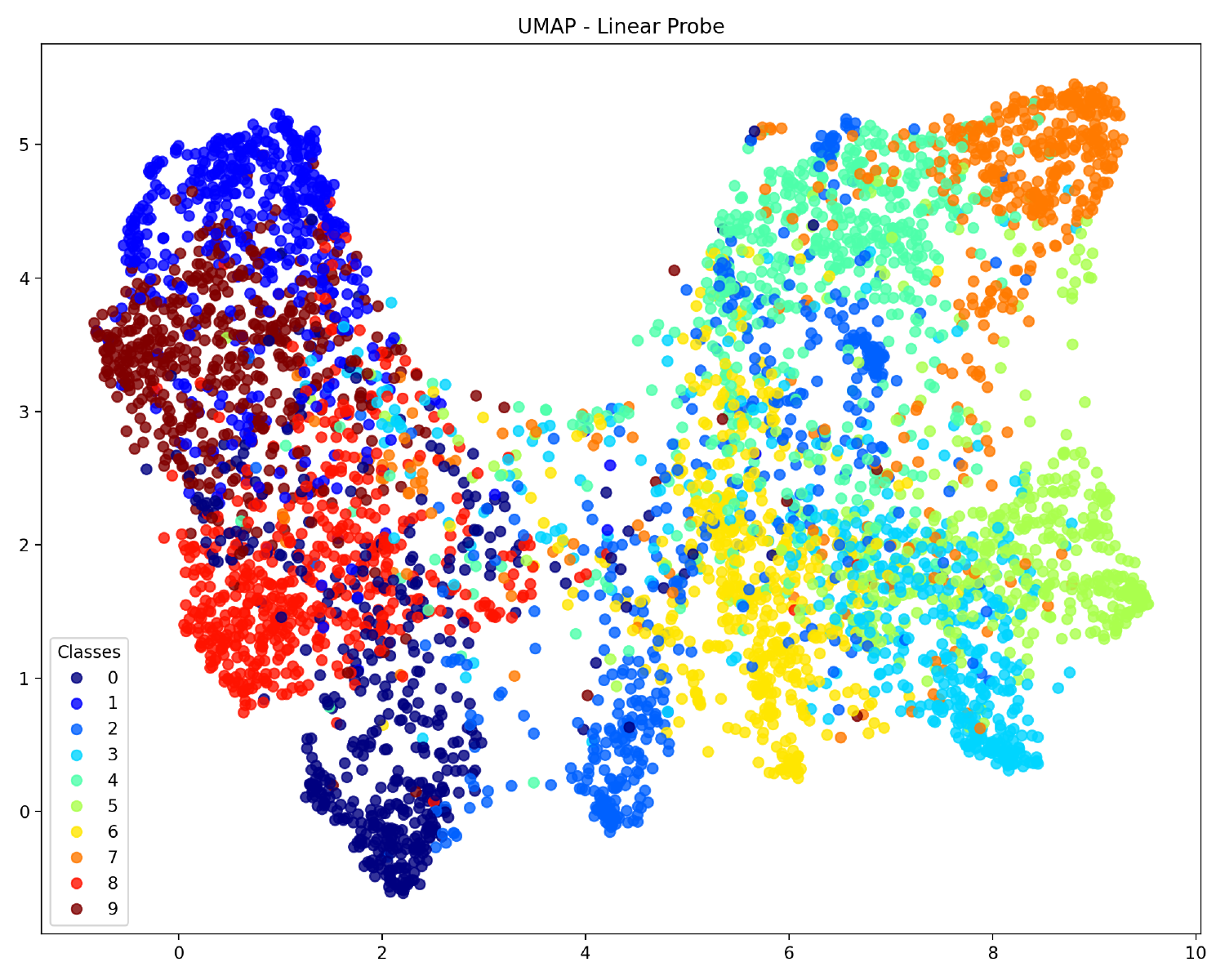}
    \caption{UMAP visualization of features from 20\% CIFAR-10 (Linear Probe)}
    \label{fig:umap_linear_probe}
\end{figure}

In Figure~\ref{fig:umap_linear_probe}, UMAP provides another perspective on the feature distribution, capturing more of the global data structure.

\subsection{Fine-tuning ResNet-152 on the Remaining 80\% of CIFAR-10 Data}

Fine-tuning involves updating the weights of a pre-trained model on a new dataset, allowing it to adapt to the specific features of the new data and potentially improve performance.

\subsubsection{Python Code for Fine-tuning and Visualization}

The following code fine-tunes the ResNet-152 model on the remaining 80\% of the CIFAR-10 data and visualizes the updated features.

\begin{lstlisting}[style=python]
import tensorflow as tf
import numpy as np
import matplotlib.pyplot as plt
from tensorflow.keras.applications import ResNet152
from tensorflow.keras.applications.resnet import preprocess_input
from tensorflow.keras.preprocessing.image import ImageDataGenerator
from tensorflow.keras.models import Model
from tensorflow.keras.layers import Dense, GlobalAveragePooling2D
from sklearn.decomposition import PCA
from sklearn.manifold import TSNE
import umap
import os

# Set random seed for reproducibility
np.random.seed(42)
tf.random.set_seed(42)

# Create directory to save visualizations
os.makedirs('visualizations', exist_ok=True)

# Load CIFAR-10 dataset
(x_train_full, y_train_full), _ = tf.keras.datasets.cifar10.load_data()
y_train_full = y_train_full.flatten()

# Normalize images and resize to match ResNet input size
def preprocess_images(images):
    images_resized = tf.image.resize(images, [224, 224])
    return preprocess_input(images_resized)

# Split dataset: 20% for Linear Probe, 80% for Fine-tuning
num_samples = x_train_full.shape[0]
indices = np.arange(num_samples)
np.random.shuffle(indices)

split_idx = int(0.2 * num_samples)
linear_probe_idx = indices[:split_idx]
fine_tune_idx = indices[split_idx:]

x_fine_tune = x_train_full[fine_tune_idx]
y_fine_tune = y_train_full[fine_tune_idx]

# Preprocess images
x_fine_tune_preprocessed = preprocess_images(x_fine_tune)

# Convert labels to categorical
y_fine_tune_categorical = tf.keras.utils.to_categorical(y_fine_tune, num_classes=10)

# Load pre-trained ResNet-152 model
base_model = ResNet152(weights='imagenet', include_top=False)

# Add new classification layers
x = base_model.output
x = GlobalAveragePooling2D()(x)
predictions = Dense(10, activation='softmax')(x)

# Define the fine-tuned model
fine_tuned_model = Model(inputs=base_model.input, outputs=predictions)

# Compile the model
fine_tuned_model.compile(optimizer=tf.keras.optimizers.SGD(lr=0.001, momentum=0.9),
                         loss='categorical_crossentropy',
                         metrics=['accuracy'])

# Fine-tune the model
fine_tuned_model.fit(x_fine_tune_preprocessed, y_fine_tune_categorical,
                     batch_size=32, epochs=5, verbose=1)

# Extract features from fine-tuned model
feature_extractor = Model(inputs=fine_tuned_model.input, outputs=fine_tuned_model.layers[-2].output)
features_fine_tuned = feature_extractor.predict(x_fine_tune_preprocessed, batch_size=32, verbose=1)

# Visualization function
def visualize(features, labels, method='pca', title='', filename=''):
    if method == 'pca':
        reducer = PCA(n_components=2)
    elif method == 'tsne':
        reducer = TSNE(n_components=2, random_state=42)
    elif method == 'umap':
        reducer = umap.UMAP(n_components=2, random_state=42)

    reduced_features = reducer.fit_transform(features)

    plt.figure(figsize=(10, 8))
    scatter = plt.scatter(reduced_features[:, 0], reduced_features[:, 1], c=labels, cmap='tab10', alpha=0.7)
    plt.legend(*scatter.legend_elements(), title="Classes")
    plt.title(title)
    plt.savefig(f'visualizations/{filename}.png', dpi=200)
    plt.close()

# Visualize and save Fine-tuned results
visualize(features_fine_tuned, y_fine_tune, method='pca', title='PCA - Fine-tuned', filename='PCA_Fine_tuned')
visualize(features_fine_tuned, y_fine_tune, method='tsne', title='t-SNE - Fine-tuned', filename='tSNE_Fine_tuned')
visualize(features_fine_tuned, y_fine_tune, method='umap', title='UMAP - Fine-tuned', filename='UMAP_Fine_tuned')
\end{lstlisting}

\subsubsection{Understanding the Code}

In this code:

\begin{itemize}
    \item We prepare the 80\% dataset for fine-tuning by preprocessing the images and converting labels to categorical format.
    \item We load the pre-trained ResNet-152 model and add new classification layers suitable for CIFAR-10.
    \item The entire model is compiled and fine-tuned on the 80\% dataset for a specified number of epochs.
    \item After training, we define a new model (\texttt{feature\_extractor}) to extract features from the fine-tuned model.
    \item The extracted features are then visualized using PCA, t-SNE, and UMAP, and the plots are saved with matching filenames.
\end{itemize}

\subsubsection{PCA Visualization After Fine-tuning}

\begin{figure}[H]
    \centering
    \includegraphics[width=0.8\textwidth]{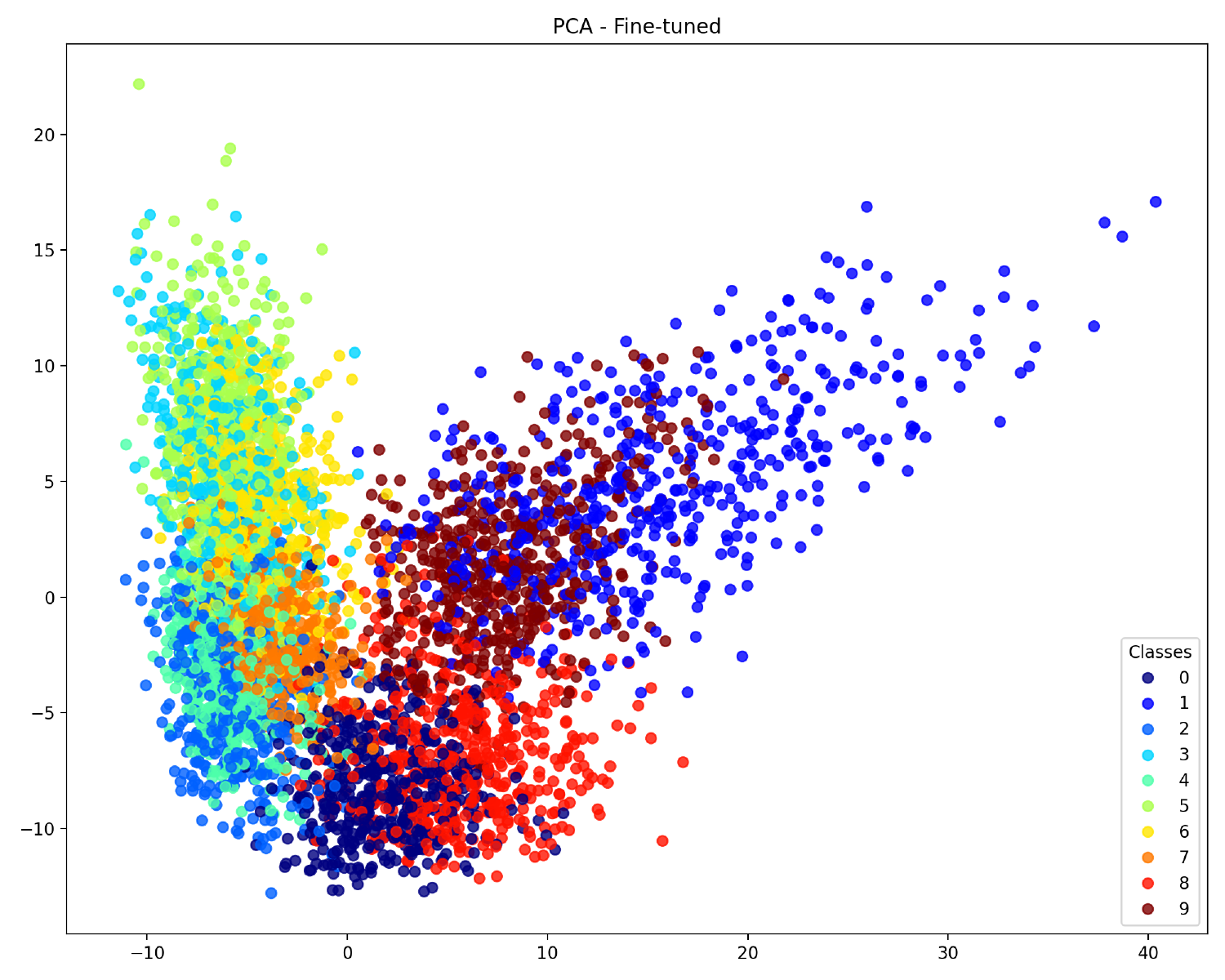}
    \caption{PCA visualization of fine-tuned features from 80\% CIFAR-10}
    \label{fig:pca_fine_tuned}
\end{figure}

Figure~\ref{fig:pca_fine_tuned} shows the PCA visualization after fine-tuning. Comparing this to the earlier PCA plot, we can see how the feature representation has changed.

\subsubsection{t-SNE Visualization After Fine-tuning}

\begin{figure}[H]
    \centering
    \includegraphics[width=0.8\textwidth]{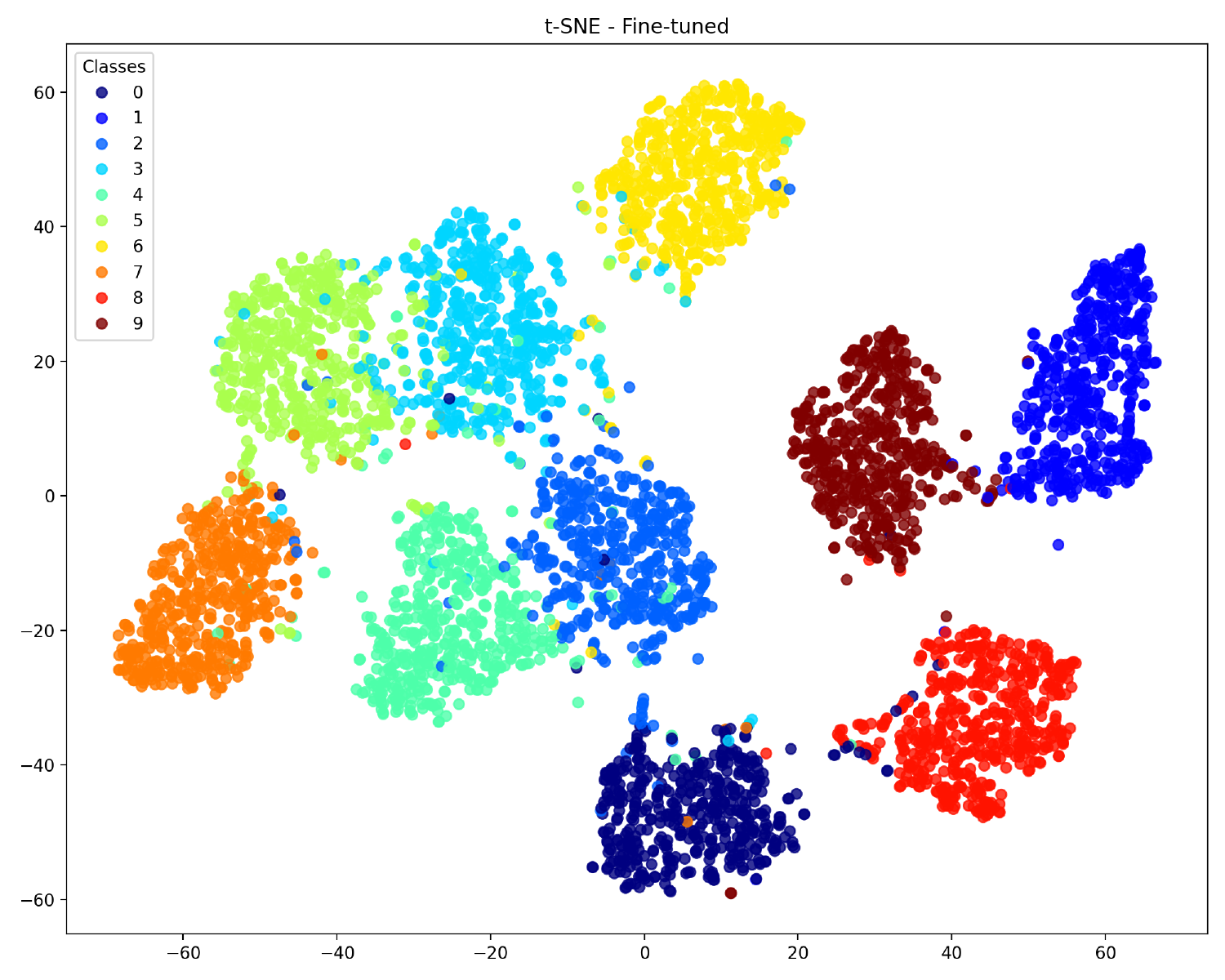}
    \caption{t-SNE visualization of fine-tuned features from 80\% CIFAR-10}
    \label{fig:tsne_fine_tuned}
\end{figure}

In Figure~\ref{fig:tsne_fine_tuned}, the t-SNE visualization shows more distinct clusters, indicating that the model has learned more specific features of CIFAR-10.

\subsubsection{UMAP Visualization After Fine-tuning}

\begin{figure}[H]
    \centering
    \includegraphics[width=0.8\textwidth]{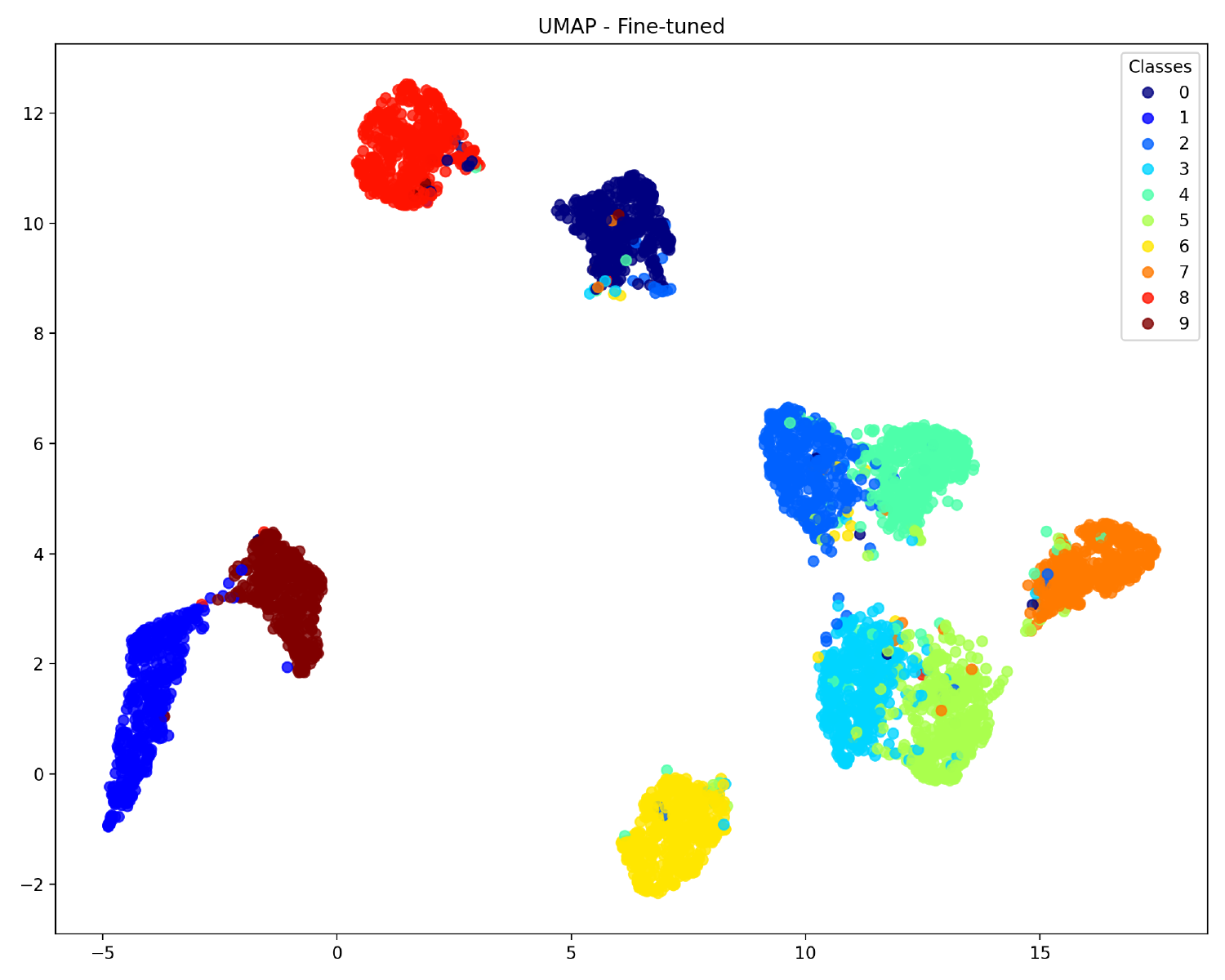}
    \caption{UMAP visualization of fine-tuned features from 80\% CIFAR-10}
    \label{fig:umap_fine_tuned}
\end{figure}

Figure~\ref{fig:umap_fine_tuned} provides the UMAP visualization after fine-tuning, offering another perspective on how the model's feature representation has improved.

\subsection{Conclusion}

By comparing the visualizations before and after fine-tuning, we observe that:

\begin{itemize}
    \item Fine-tuning helps the model adapt to the specific features of the CIFAR-10 dataset.
    \item Clusters in the visualizations become more distinct after fine-tuning, indicating better class separation.
    \item Dimensionality reduction techniques like PCA, t-SNE, and UMAP are valuable tools for analyzing and understanding high-dimensional data.
\end{itemize}

This exercise demonstrates the importance of model adaptation techniques and how visualization can aid in interpreting model performance.

\section{VGG (2014)}

The VGG network, introduced in 2014 by the Visual Geometry Group at Oxford University, is a highly influential deep convolutional neural network (CNN). The key feature of VGG is its use of small 3x3 convolutional filters and the depth of the network. The deeper the network, the better its capacity to learn complex patterns from images. In this section, we will discuss the structure of the VGG family (VGG-11, VGG-13, VGG-16, and VGG-19), explain their components, and provide code examples for VGG-16 \cite{simonyan2015deepconvolutionalnetworkslargescale}.

\textbf{Comparison of VGG Architectures}

Below is a comparison of the main components in each VGG architecture. The table compares the number of convolutional layers, max-pooling layers, and fully connected layers (classification head) in each model. This table follows the format from the original VGG paper.

\begin{center}
\begin{tabular}{|c|c|c|c|c|}
\hline
\textbf{Component} & \textbf{VGG-11} & \textbf{VGG-13} & \textbf{VGG-16} & \textbf{VGG-19} \\
\hline
Conv3-64 & 1 & 2 & 2 & 2 \\
Conv3-128 & 1 & 2 & 2 & 2 \\
Conv3-256 & 2 & 2 & 3 & 4 \\
Conv3-512 & 2 & 2 & 3 & 4 \\
MaxPooling & 5 & 5 & 5 & 5 \\
\hline
FC-4096 & \multicolumn{4}{c|}{1} \\
FC-4096 & \multicolumn{4}{c|}{1} \\
FC-1000 (Softmax) & \multicolumn{4}{c|}{1} \\
\hline
\end{tabular}
\end{center}

\paragraph{Explanation of the Components}

\textbf{Conv3-64}: The first block consists of 3x3 convolutional filters applied to the input image (or feature map). The number "64" refers to the number of feature maps (or channels) output by the layer. Deeper models (VGG-13, VGG-16, VGG-19) have more convolutional layers to capture finer details in the data.

\textbf{Conv3-128}: In the second block, the convolutional filters continue to extract features from the previous layer's output, increasing the number of channels to 128, allowing the network to capture more complex patterns.

\textbf{Conv3-256}: The third block increases the number of feature maps to 256, further refining the information captured from the image. Models like VGG-16 and VGG-19 use additional layers here to extract more detailed information.

\textbf{Conv3-512}: The fourth and fifth blocks increase the number of feature maps to 512, capturing very high-level abstract features from the input. Deeper models like VGG-16 and VGG-19 have more layers here, making them more powerful at recognizing complex patterns in images.

\textbf{MaxPooling}: Max-pooling layers are inserted after each set of convolutional layers. These layers reduce the spatial size of the feature maps, which helps in reducing the computational cost and the number of parameters, while retaining important spatial features.

\textbf{Classification Head}: All VGG models have an identical classification head. This consists of three fully connected layers:
- First fully connected layer: 4096 units
- Second fully connected layer: 4096 units
- Final fully connected layer: 1000 units with softmax activation for classification into 1000 categories.

\paragraph{Design Philosophy of VGG}

The VGG architecture is built with simplicity in mind. Instead of using large convolutional filters, VGG opts for small 3x3 filters, which allows the network to increase depth (number of layers) while keeping the computational complexity manageable. This depth gives the network greater capacity to learn more intricate patterns in the data.

\textbf{TensorFlow Code for VGG-16}

Now that we have explained the components, let's implement the VGG-16 model using TensorFlow:

\begin{lstlisting}[style=python]
import tensorflow as tf

def VGG16(input_shape=(224, 224, 3), num_classes=1000):
    model = tf.keras.Sequential()
    
    # Block 1
    model.add(tf.keras.layers.Conv2D(64, (3, 3), padding='same', activation='relu', input_shape=input_shape))
    model.add(tf.keras.layers.Conv2D(64, (3, 3), padding='same', activation='relu'))
    model.add(tf.keras.layers.MaxPooling2D((2, 2), strides=(2, 2)))
    
    # Block 2
    model.add(tf.keras.layers.Conv2D(128, (3, 3), padding='same', activation='relu'))
    model.add(tf.keras.layers.Conv2D(128, (3, 3), padding='same', activation='relu'))
    model.add(tf.keras.layers.MaxPooling2D((2, 2), strides=(2, 2)))
    
    # Block 3
    model.add(tf.keras.layers.Conv2D(256, (3, 3), padding='same', activation='relu'))
    model.add(tf.keras.layers.Conv2D(256, (3, 3), padding='same', activation='relu'))
    model.add(tf.keras.layers.Conv2D(256, (3, 3), padding='same', activation='relu'))
    model.add(tf.keras.layers.MaxPooling2D((2, 2), strides=(2, 2)))
    
    # Block 4
    model.add(tf.keras.layers.Conv2D(512, (3, 3), padding='same', activation='relu'))
    model.add(tf.keras.layers.Conv2D(512, (3, 3), padding='same', activation='relu'))
    model.add(tf.keras.layers.Conv2D(512, (3, 3), padding='same', activation='relu'))
    model.add(tf.keras.layers.MaxPooling2D((2, 2), strides=(2, 2)))
    
    # Block 5
    model.add(tf.keras.layers.Conv2D(512, (3, 3), padding='same', activation='relu'))
    model.add(tf.keras.layers.Conv2D(512, (3, 3), padding='same', activation='relu'))
    model.add(tf.keras.layers.Conv2D(512, (3, 3), padding='same', activation='relu'))
    model.add(tf.keras.layers.MaxPooling2D((2, 2), strides=(2, 2)))
    
    # Fully connected layers
    model.add(tf.keras.layers.Flatten())
    model.add(tf.keras.layers.Dense(4096, activation='relu'))
    model.add(tf.keras.layers.Dense(4096, activation='relu'))
    model.add(tf.keras.layers.Dense(num_classes, activation='softmax'))
    
    return model

# Create the VGG16 model
model = VGG16()
model.summary()
\end{lstlisting}

\paragraph{Key Insights for Beginners}

\textbf{Why 3x3 filters?} Small filters like 3x3 allow VGG to increase depth, capturing more complex patterns while keeping computational costs under control.

\textbf{Why use MaxPooling?} Pooling reduces the spatial size of the data progressively, preventing the model from becoming too large and reducing the chances of overfitting.

\textbf{Fully Connected Layers}: The fully connected layers at the end are responsible for interpreting the high-level features and making the final classification.

\subsection{VGG16}
The VGG16 model is widely used in image classification tasks. Here, we will use the VGG16 model pre-trained on the ImageNet dataset and apply transfer learning to the CIFAR-10 dataset using two approaches: Linear Probe and Fine-tuning. The CIFAR-10 dataset contains images of size 32x32 pixels, so we will resize them to 224x224 pixels to match the input size expected by VGG16 \cite{simonyan2015deepconvolutionalnetworkslargescale}.

\paragraph{Linear Probe} 
In the Linear Probe approach, we freeze the pre-trained VGG16 model's convolutional layers and train only the classification layers on CIFAR-10.

\begin{lstlisting}[style=python]
import tensorflow as tf
from tensorflow.keras import layers, models
from tensorflow.keras.applications import VGG16
from tensorflow.keras.datasets import cifar10
from tensorflow.keras.utils import to_categorical

# Load CIFAR-10 dataset and preprocess images
(x_train, y_train), (x_test, y_test) = cifar10.load_data()
x_train = tf.image.resize(x_train, (224, 224)) / 255.0
x_test = tf.image.resize(x_test, (224, 224)) / 255.0
y_train = to_categorical(y_train, 10)
y_test = to_categorical(y_test, 10)

# Load pre-trained VGG16 model without the top classification layer
base_model = VGG16(weights='imagenet', include_top=False, input_shape=(224, 224, 3))

# Freeze all layers of the base model
for layer in base_model.layers:
    layer.trainable = False

# Add classification layers on top of the base model
model = models.Sequential([
    base_model,
    layers.Flatten(),
    layers.Dense(256, activation='relu'),
    layers.Dense(10, activation='softmax')
])

# Compile the model
model.compile(optimizer='adam', loss='categorical_crossentropy', metrics=['accuracy'])

# Train the model for linear probe
history = model.fit(x_train, y_train, epochs=10, validation_data=(x_test, y_test))

# Evaluate the model
results = model.evaluate(x_test, y_test)
print(f"VGG16 Linear Probe Test Accuracy: {results[1]}")
\end{lstlisting}

In this code, we load the CIFAR-10 dataset, resize the images to 224x224 pixels, and normalize the pixel values. The VGG16 model is loaded without its top classification layers, and we freeze all of its convolutional layers. A custom classification head is added, consisting of a Flatten layer and two Dense layers. The model is trained for 10 epochs, and the test accuracy after training is printed.

\paragraph{Fine-tuning} 
Fine-tuning involves unfreezing some or all of the pre-trained layers and training them along with the classification layers. This allows the model to better adapt the pre-trained features to the new dataset.

\begin{lstlisting}[style=python]
# Load and preprocess CIFAR-10 dataset (same as before)
(x_train, y_train), (x_test, y_test) = cifar10.load_data()
x_train = tf.image.resize(x_train, (224, 224)) / 255.0
x_test = tf.image.resize(x_test, (224, 224)) / 255.0
y_train = to_categorical(y_train, 10)
y_test = to_categorical(y_test, 10)

# Load pre-trained VGG16 model without the top classification layer
base_model = VGG16(weights='imagenet', include_top=False, input_shape=(224, 224, 3))

# Unfreeze the last 4 layers of the base model for fine-tuning
for layer in base_model.layers[-4:]:
    layer.trainable = True

# Add classification layers
model = models.Sequential([
    base_model,
    layers.Flatten(),
    layers.Dense(256, activation='relu'),
    layers.Dense(10, activation='softmax')
])

# Compile the model with a smaller learning rate for fine-tuning
model.compile(optimizer=tf.keras.optimizers.Adam(1e-5), loss='categorical_crossentropy', metrics=['accuracy'])

# Fine-tune the model
fine_tune_history = model.fit(x_train, y_train, epochs=10, validation_data=(x_test, y_test))

# Evaluate the model
fine_tune_results = model.evaluate(x_test, y_test)
print(f"VGG16 Fine-tuning Test Accuracy: {fine_tune_results[1]}")
\end{lstlisting}

In this fine-tuning approach, we unfreeze the last four layers of the VGG16 model, allowing them to be updated during training. The model is compiled with a smaller learning rate (`1e-5') to ensure that the pre-trained weights are updated gradually. After fine-tuning, we print the model's test accuracy, which typically improves compared to the linear probe approach.

\subsection{VGG19}
VGG19 is a deeper version of VGG16, with more convolutional layers. Here, we apply both the Linear Probe and Fine-tuning methods to VGG19, similar to what we did with VGG16 \cite{simonyan2015deepconvolutionalnetworkslargescale}.

\paragraph{Linear Probe} 
In the Linear Probe approach for VGG19, we freeze all the pre-trained layers and only train the classification layers.

\begin{lstlisting}[style=python]
from tensorflow.keras.applications import VGG19

# Load CIFAR-10 dataset and preprocess images
(x_train, y_train), (x_test, y_test) = cifar10.load_data()
x_train = tf.image.resize(x_train, (224, 224)) / 255.0
x_test = tf.image.resize(x_test, (224, 224)) / 255.0
y_train = to_categorical(y_train, 10)
y_test = to_categorical(y_test, 10)

# Load pre-trained VGG19 model without the top classification layer
base_model_vgg19 = VGG19(weights='imagenet', include_top=False, input_shape=(224, 224, 3))

# Freeze all layers of the base model
for layer in base_model_vgg19.layers:
    layer.trainable = False

# Add classification layers
model_vgg19 = models.Sequential([
    base_model_vgg19,
    layers.Flatten(),
    layers.Dense(256, activation='relu'),
    layers.Dense(10, activation='softmax')
])

# Compile the model
model_vgg19.compile(optimizer='adam', loss='categorical_crossentropy', metrics=['accuracy'])

# Train the model for linear probe
vgg19_history = model_vgg19.fit(x_train, y_train, epochs=10, validation_data=(x_test, y_test))

# Evaluate the model
vgg19_results = model_vgg19.evaluate(x_test, y_test)
print(f"VGG19 Linear Probe Test Accuracy: {vgg19_results[1]}")
\end{lstlisting}

In this code, we use the VGG19 model, pre-trained on ImageNet, and freeze all of its layers. After adding a new classification head, we train the model on the CIFAR-10 dataset for 10 epochs and evaluate the test accuracy. This serves as the linear probe approach for VGG19.

\paragraph{Fine-tuning} 
Similar to VGG16, fine-tuning for VGG19 involves unfreezing the last few layers of the model and training them along with the new classification layers.

\begin{lstlisting}[style=python]
# Load and preprocess CIFAR-10 dataset (same as before)
(x_train, y_train), (x_test, y_test) = cifar10.load_data()
x_train = tf.image.resize(x_train, (224, 224)) / 255.0
x_test = tf.image.resize(x_test, (224, 224)) / 255.0
y_train = to_categorical(y_train, 10)
y_test = to_categorical(y_test, 10)

# Load pre-trained VGG19 model without the top classification layer
base_model_vgg19 = VGG19(weights='imagenet', include_top=False, input_shape=(224, 224, 3))

# Unfreeze the last 4 layers of the base model for fine-tuning
for layer in base_model_vgg19.layers[-4:]:
    layer.trainable = True

# Add classification layers
model_vgg19 = models.Sequential([
    base_model_vgg19,
    layers.Flatten(),
    layers.Dense(256, activation='relu'),
    layers.Dense(10, activation='softmax')
])

# Compile the model with a smaller learning rate for fine-tuning
model_vgg19.compile(optimizer=tf.keras.optimizers.Adam(1e-5), loss='categorical_crossentropy', metrics=['accuracy'])

# Fine-tune the model
vgg19_fine_tune_history = model_vgg19.fit(x_train, y_train, epochs=10, validation_data=(x_test, y_test))

# Evaluate the model
vgg19_fine_tune_results = model_vgg19.evaluate(x_test, y_test)
print(f"VGG19 Fine-tuning Test Accuracy: {vgg19_fine_tune_results[1]}")
\end{lstlisting}

In this approach, we unfreeze the last four layers of VGG19 and fine-tune the model for another 10 epochs. By using a lower learning rate, we carefully update the pre-trained weights. The test accuracy after fine-tuning is printed to show how this method improves performance compared to the linear probe approach.

\section{Inception (2015)}

The Inception network, also known as GoogLeNet, was introduced in 2015 by Szegedy et al. in the paper "Going Deeper with Convolutions." Inception’s key contribution was to allow for increased depth and width of neural networks without excessively increasing computational complexity. This was achieved by introducing the \textbf{Inception module}, a structure that applies multiple convolution filters of different sizes in parallel, allowing the network to capture features at various scales \cite{szegedy2015rethinkinginceptionarchitecturecomputer}.

Inception was highly successful, winning the ImageNet Large-Scale Visual Recognition Challenge (ILSVRC) in 2014, and laid the foundation for later versions such as Inception-v3 and Inception-v4.

\paragraph{The Inception Module}

The core idea of the Inception module is to apply multiple types of operations to the same input and concatenate the outputs. These operations include:
- \(1 \times 1\) convolutions, which are used for dimensionality reduction.
- \(3 \times 3\) and \(5 \times 5\) convolutions, which capture spatial features at different scales.
- \(3 \times 3\) max pooling, which reduces spatial dimensions and adds more robust features.

The benefit of the Inception module is that it allows the network to look at the input in different ways (i.e., at different resolutions) without having to decide in advance what size of convolution filter to use. This multi-scale approach improves the model's ability to capture diverse types of patterns in the input data.

\paragraph{Mathematical Explanation of the Inception Module}

For a given input \(x\), the output of an Inception module consists of the concatenation of four parallel paths:
1. \(1 \times 1\) convolution: This reduces the depth (number of channels) of the input before applying larger convolutions.
2. \(3 \times 3\) convolution: This captures medium-scale features.
3. \(5 \times 5\) convolution: This captures larger-scale features.
4. \(3 \times 3\) max pooling: This reduces spatial size and captures prominent features.

Let \(f_{1 \times 1}(x)\), \(f_{3 \times 3}(x)\), and \(f_{5 \times 5}(x)\) represent the outputs of the \(1 \times 1\), \(3 \times 3\), and \(5 \times 5\) convolutions, and \(f_{\text{pool}}(x)\) the output of the max pooling layer. The overall output \(y\) of the Inception module is the concatenation of these outputs:

\[
y = \text{concat}(f_{1 \times 1}(x), f_{3 \times 3}(x), f_{5 \times 5}(x), f_{\text{pool}}(x))
\]

The concatenation step allows the network to combine information from different scales.

\paragraph{Inception Module Diagram}

Below is an updated diagram illustrating the structure of the Inception module, which includes parallel convolution and pooling operations:

\usetikzlibrary{positioning}

\begin{tikzpicture}[node distance=1.0cm, auto]
    % Input node
    \node (input) [rectangle, draw, text centered, minimum height=1.5em, minimum width=3cm] {Input};
    
    % 1x1 conv branch
    \node (conv1x1) [rectangle, draw, below left=of input, xshift=-2.0cm, text centered, minimum height=1.5em, minimum width=3cm] {1x1 Conv};
    
    % 3x3 conv branch
    \node (conv3x3reduce) [rectangle, draw, below=of input, xshift=-2.0cm, text centered, minimum height=1.5em, minimum width=3cm] {1x1 Conv (reduce)};
    \node (conv3x3) [rectangle, draw, below=of conv3x3reduce, text centered, minimum height=1.5em, minimum width=3cm] {3x3 Conv};
    
    % 5x5 conv branch
    \node (conv5x5reduce) [rectangle, draw, below right=of input, xshift=-2.0cm, text centered, minimum height=1.5em, minimum width=3cm] {1x1 Conv (reduce)};
    \node (conv5x5) [rectangle, draw, below=of conv5x5reduce, text centered, minimum height=1.5em, minimum width=3cm] {5x5 Conv};
    
    % Max pooling branch
    \node (pool) [rectangle, draw, below right=of input, xshift=-3.0cm, xshift=5cm, text centered, minimum height=1.5em, minimum width=3cm] {3x3 Max Pooling};
    \node (pool1x1) [rectangle, draw, below=of pool, text centered, minimum height=1.5em, minimum width=3cm] {1x1 Conv};
    
    % Concatenate node
    \node (concat) [rectangle, draw, below=1.0cm of conv3x3, text centered, minimum height=1.5em, minimum width=3cm] {Concatenation};
    
    % Connections
    \draw[->] (input) -- (conv1x1);
    \draw[->] (input) -- (conv3x3reduce);
    \draw[->] (input) -- (conv5x5reduce);
    \draw[->] (input) -- (pool);
    \draw[->] (conv3x3reduce) -- (conv3x3);
    \draw[->] (conv5x5reduce) -- (conv5x5);
    \draw[->] (pool) -- (pool1x1);
    
    \draw[->] (conv1x1) -- (concat);
    \draw[->] (conv3x3) -- (concat);
    \draw[->] (conv5x5) -- (concat);
    \draw[->] (pool1x1) -- (concat);
\end{tikzpicture}

\paragraph{Comparison of Inception Variants}

The Inception architecture has evolved through several versions (v1, v2, v3, v4). Each version introduces modifications to improve accuracy and efficiency. Below is a table comparing the key components of Inception-v1, Inception-v3, and Inception-v4.

\begin{center}
\begin{tabular}{|c|c|c|c|}
\hline
\textbf{Component} & \textbf{Inception-v1} & \textbf{Inception-v3} & \textbf{Inception-v4} \\
\hline
Number of Layers & 22 & 48 & 55 \\
1x1 Convolutions & Yes & Yes & Yes \\
3x3 Convolutions & Yes & Yes (Factorized) & Yes (Factorized) \\
5x5 Convolutions & Yes & No (Replaced by 2x 3x3) & No (Replaced by 2x 3x3) \\
MaxPooling & Yes & Yes & Yes \\
AveragePooling & Yes & Yes & Yes \\
Auxiliary Classifiers & Yes & Yes & Yes \\
\hline
\end{tabular}
\end{center}

\paragraph{Explanation of the Components}

\textbf{1x1 Convolutions}: Used primarily for dimensionality reduction, allowing the larger \(3 \times 3\) and \(5 \times 5\) convolutions to be applied efficiently without drastically increasing the number of parameters.

\textbf{3x3 Convolutions}: Captures medium-sized spatial features. In later versions like Inception-v3 and Inception-v4, the \(3 \times 3\) convolutions are factorized into two consecutive \(3 \times 3\) convolutions for better computational efficiency.

\textbf{5x5 Convolutions}: Used in Inception-v1 to capture larger-scale spatial features. In later versions, this is replaced by two consecutive \(3 \times 3\) convolutions, which are more computationally efficient but capture the same effective receptive field.

\textbf{MaxPooling and AveragePooling}: Pooling layers help reduce the spatial size of the feature maps, making the network more computationally efficient and robust to spatial variance in the input.

\textbf{Auxiliary Classifiers}: These are intermediate classifiers added during training to help with gradient propagation and to regularize the network. They are not used during inference but improve training stability.

\paragraph{Design Philosophy of Inception}

The design philosophy behind Inception is to handle computational complexity by allowing the network to process information at multiple scales simultaneously. The use of \(1 \times 1\) convolutions helps reduce the dimensionality of the input before applying larger convolution filters, thereby reducing the overall number of parameters. This efficiency allowed Inception to build deeper networks with greater accuracy without a corresponding increase in computational cost.

\paragraph{TensorFlow Code for Inception Module}

Below is the implementation of an Inception module using TensorFlow \texttt{tf.keras}.

\begin{lstlisting}[style=python]
import tensorflow as tf

def inception_module(input_tensor, filters):
    # 1x1 convolution branch
    branch1x1 = tf.keras.layers.Conv2D(filters, (1, 1), padding='same', activation='relu')(input_tensor)
    
    # 3x3 convolution branch
    branch3x3 = tf.keras.layers.Conv2D(filters, (1, 1), padding='same', activation='relu')(input_tensor)
    branch3x3 = tf.keras.layers.Conv2D(filters, (3, 3), padding='same', activation='relu')(branch3x3)
    
    # 5x5 convolution branch
    branch5x5 = tf.keras.layers.Conv2D(filters, (1, 1), padding='same', activation='relu')(input_tensor)
    branch5x5 = tf.keras.layers.Conv2D(filters, (5, 5), padding='same', activation='relu')(branch5x5)
    
    # Max pooling branch
    branch_pool = tf.keras.layers.MaxPooling2D((3, 3), strides=(1, 1), padding='same')(input_tensor)
    branch_pool = tf.keras.layers.Conv2D(filters, (1, 1), padding='same', activation='relu')(branch_pool)
    
    # Concatenate all branches
    output = tf.keras.layers.concatenate([branch1x1, branch3x3, branch5x5, branch_pool], axis=-1)
    
    return output

# Example Inception model
def inception_v1(input_shape=(224, 224, 3), num_classes=1000):
    input_tensor = tf.keras.Input(shape=input_shape)
    
    # Initial convolution and pooling
    x = tf.keras.layers.Conv2D(64, (7, 7), strides=2, padding='same', activation='relu')(input_tensor)
    x = tf.keras.layers.MaxPooling2D((3, 3), strides=2, padding='same')(x)
    
    # Apply Inception modules
    x = inception_module(x, 64)
    x = inception_module(x, 128)
    x = inception_module(x, 256)
    
    # Global average pooling and output
    x = tf.keras.layers.GlobalAveragePooling2D()(x)
    output_tensor = tf.keras.layers.Dense(num_classes, activation='softmax')(x)
    
    model = tf.keras.Model(inputs=input_tensor, outputs=output_tensor)
    
    return model

# Create the Inception-v1 model
model = inception_v1()
model.summary()
\end{lstlisting}

\paragraph{Key Insights for Beginners}

\textbf{Why use multiple convolution filters in parallel?} The Inception module applies multiple convolution filters of different sizes (e.g., \(1 \times 1\), \(3 \times 3\), and \(5 \times 5\)) in parallel. This allows the network to capture features at different spatial scales, improving its ability to recognize patterns in the data.

\textbf{What is dimensionality reduction with \(1 \times 1\) convolutions?} \(1 \times 1\) convolutions are used to reduce the number of input channels before applying larger filters, which significantly reduces the computational complexity of the network.

\subsection{InceptionV3}
InceptionV3 is a convolutional neural network architecture widely used for image classification. It improves upon earlier Inception models by using factorized convolutions, which reduce computational cost. In this section, we will explore how to apply Linear Probe and Fine-tuning on InceptionV3 using the CIFAR-10 dataset \cite{szegedy2015rethinkinginceptionarchitecturecomputer}.

\paragraph{Linear Probe}
In the Linear Probe approach, we freeze the convolutional layers of the pre-trained InceptionV3 model and train only the classification layers.

\begin{lstlisting}[style=python]
import tensorflow as tf
from tensorflow.keras import layers, models
from tensorflow.keras.applications import InceptionV3
from tensorflow.keras.datasets import cifar10
from tensorflow.keras.utils import to_categorical

# Load and preprocess CIFAR-10 dataset
(x_train, y_train), (x_test, y_test) = cifar10.load_data()
x_train = tf.image.resize(x_train, (299, 299)) / 255.0
x_test = tf.image.resize(x_test, (299, 299)) / 255.0
y_train = to_categorical(y_train, 10)
y_test = to_categorical(y_test, 10)

# Load pre-trained InceptionV3 model without the top classification layer
base_model_inceptionv3 = InceptionV3(weights='imagenet', include_top=False, input_shape=(299, 299, 3))

# Freeze all layers of the base model
for layer in base_model_inceptionv3.layers:
    layer.trainable = False

# Add classification layers
model_inceptionv3 = models.Sequential([
    base_model_inceptionv3,
    layers.GlobalAveragePooling2D(),
    layers.Dense(256, activation='relu'),
    layers.Dense(10, activation='softmax')
])

# Compile the model
model_inceptionv3.compile(optimizer='adam', loss='categorical_crossentropy', metrics=['accuracy'])

# Train the model for linear probe
history_inceptionv3 = model_inceptionv3.fit(x_train, y_train, epochs=10, validation_data=(x_test, y_test))

# Evaluate the model
results_inceptionv3 = model_inceptionv3.evaluate(x_test, y_test)
print(f"InceptionV3 Linear Probe Test Accuracy: {results_inceptionv3[1]}")
\end{lstlisting}

In this code, we resize CIFAR-10 images to match InceptionV3’s input size of 299x299 pixels. We freeze the pre-trained layers of the InceptionV3 model and train only the newly added classification layers. After training, the test accuracy is printed.

\paragraph{Fine-tuning}
In Fine-tuning, we unfreeze some layers of the InceptionV3 model and train them along with the classification layers to better adapt the model to the CIFAR-10 dataset.

\begin{lstlisting}[style=python]
import tensorflow as tf
from tensorflow.keras import layers, models
from tensorflow.keras.applications import InceptionV3
from tensorflow.keras.datasets import cifar10
from tensorflow.keras.utils import to_categorical

# Load and preprocess CIFAR-10 dataset
(x_train, y_train), (x_test, y_test) = cifar10.load_data()
x_train = tf.image.resize(x_train, (299, 299)) / 255.0
x_test = tf.image.resize(x_test, (299, 299)) / 255.0
y_train = to_categorical(y_train, 10)
y_test = to_categorical(y_test, 10)

# Load pre-trained InceptionV3 model without the top classification layer
base_model_inceptionv3 = InceptionV3(weights='imagenet', include_top=False, input_shape=(299, 299, 3))

# Unfreeze the last few layers for fine-tuning
for layer in base_model_inceptionv3.layers[-20:]:
    layer.trainable = True

# Add classification layers
model_inceptionv3 = models.Sequential([
    base_model_inceptionv3,
    layers.GlobalAveragePooling2D(),
    layers.Dense(256, activation='relu'),
    layers.Dense(10, activation='softmax')
])

# Compile the model with a smaller learning rate
model_inceptionv3.compile(optimizer=tf.keras.optimizers.Adam(1e-5), loss='categorical_crossentropy', metrics=['accuracy'])

# Fine-tune the model
fine_tune_history_inceptionv3 = model_inceptionv3.fit(x_train, y_train, epochs=10, validation_data=(x_test, y_test))

# Evaluate the model
fine_tune_results_inceptionv3 = model_inceptionv3.evaluate(x_test, y_test)
print(f"InceptionV3 Fine-tuning Test Accuracy: {fine_tune_results_inceptionv3[1]}")
\end{lstlisting}

For fine-tuning, we unfreeze the last 20 layers of the InceptionV3 model and train them along with the classification layers. A smaller learning rate is used to adjust the pre-trained weights gradually. After fine-tuning, the test accuracy is printed.

\subsection{InceptionResNet}
InceptionResNet combines the strengths of both Inception and ResNet architectures, using residual connections to improve training stability and performance. Below, we explore InceptionResNetV2 with both Linear Probe and Fine-tuning approaches \cite{DBLP:journals/corr/SzegedyIV16}.

\subsubsection{InceptionResNetV2}
InceptionResNetV2 is an architecture that combines the inception modules with residual connections. Here, we apply both Linear Probe and Fine-tuning on the CIFAR-10 dataset.

\paragraph{Linear Probe}
In the Linear Probe approach, we freeze the convolutional layers of the pre-trained InceptionResNetV2 model and train only the classification layers.

\begin{lstlisting}[style=python]
import tensorflow as tf
from tensorflow.keras import layers, models
from tensorflow.keras.applications import InceptionResNetV2
from tensorflow.keras.datasets import cifar10
from tensorflow.keras.utils import to_categorical

# Load and preprocess CIFAR-10 dataset
(x_train, y_train), (x_test, y_test) = cifar10.load_data()
x_train = tf.image.resize(x_train, (299, 299)) / 255.0
x_test = tf.image.resize(x_test, (299, 299)) / 255.0
y_train = to_categorical(y_train, 10)
y_test = to_categorical(y_test, 10)

# Load pre-trained InceptionResNetV2 model without the top classification layer
base_model_inceptionresnetv2 = InceptionResNetV2(weights='imagenet', include_top=False, input_shape=(299, 299, 3))

# Freeze all layers of the base model
for layer in base_model_inceptionresnetv2.layers:
    layer.trainable = False

# Add classification layers
model_inceptionresnetv2 = models.Sequential([
    base_model_inceptionresnetv2,
    layers.GlobalAveragePooling2D(),
    layers.Dense(256, activation='relu'),
    layers.Dense(10, activation='softmax')
])

# Compile the model
model_inceptionresnetv2.compile(optimizer='adam', loss='categorical_crossentropy', metrics=['accuracy'])

# Train the model for linear probe
history_inceptionresnetv2 = model_inceptionresnetv2.fit(x_train, y_train, epochs=10, validation_data=(x_test, y_test))

# Evaluate the model
results_inceptionresnetv2 = model_inceptionresnetv2.evaluate(x_test, y_test)
print(f"InceptionResNetV2 Linear Probe Test Accuracy: {results_inceptionresnetv2[1]}")
\end{lstlisting}

In this code, we resize the CIFAR-10 images to 299x299 pixels to match InceptionResNetV2's input size. We freeze all the convolutional layers of the pre-trained model and train only the classification layers. The test accuracy is printed after training.

\paragraph{Fine-tuning}
In the Fine-tuning approach, we unfreeze some layers of the InceptionResNetV2 model and train them along with the classification layers.

\begin{lstlisting}[style=python]
import tensorflow as tf
from tensorflow.keras import layers, models
from tensorflow.keras.applications import InceptionResNetV2
from tensorflow.keras.datasets import cifar10
from tensorflow.keras.utils import to_categorical

# Load and preprocess CIFAR-10 dataset
(x_train, y_train), (x_test, y_test) = cifar10.load_data()
x_train = tf.image.resize(x_train, (299, 299)) / 255.0
x_test = tf.image.resize(x_test, (299, 299)) / 255.0
y_train = to_categorical(y_train, 10)
y_test = to_categorical(y_test, 10)

# Load pre-trained InceptionResNetV2 model without the top classification layer
base_model_inceptionresnetv2 = InceptionResNetV2(weights='imagenet', include_top=False, input_shape=(299, 299, 3))

# Unfreeze the last few layers for fine-tuning
for layer in base_model_inceptionresnetv2.layers[-20:]:
    layer.trainable = True

# Add classification layers
model_inceptionresnetv2 = models.Sequential([
    base_model_inceptionresnetv2,
    layers.GlobalAveragePooling2D(),
    layers.Dense(256, activation='relu'),
    layers.Dense(10, activation='softmax')
])

# Compile the model with a smaller learning rate
model_inceptionresnetv2.compile(optimizer=tf.keras.optimizers.Adam(1e-5), loss='categorical_crossentropy', metrics=['accuracy'])

# Fine-tune the model
fine_tune_history_inceptionresnetv2 = model_inceptionresnetv2.fit(x_train, y_train, epochs=10, validation_data=(x_test, y_test))

# Evaluate the model
fine_tune_results_inceptionresnetv2 = model_inceptionresnetv2.evaluate(x_test, y_test)
print(f"InceptionResNetV2 Fine-tuning Test Accuracy: {fine_tune_results_inceptionresnetv2[1]}")
\end{lstlisting}

For fine-tuning, we unfreeze the last 20 layers of InceptionResNetV2 and train the entire model with a smaller learning rate. After fine-tuning, the test accuracy is printed.

\section{ResNet (2015)}

ResNet, or Residual Networks, were introduced in 2015 by Kaiming He and colleagues to address the vanishing gradient problem that occurs in very deep networks. Deep networks are more powerful in learning complex patterns from data, but as depth increases, training becomes more difficult due to problems like vanishing gradients. ResNet introduces the concept of residual learning, which enables networks to be trained with very deep architectures, such as 50, 101, and even 152 layers, without degradation in performance.

\paragraph{Why Use Residual Learning?}

The main challenge in training deep networks is the \textbf{vanishing gradient problem}, which occurs when the gradients of the loss function with respect to the network parameters become extremely small during backpropagation. This problem is exacerbated in very deep networks, where information needs to propagate through many layers. As a result, early layers in the network receive minimal updates, leading to slow learning and poor performance.

The issue arises due to the chain rule of differentiation, also known as \textbf{chain-based backpropagation}. In a deep network, each layer's gradients are calculated as the product of the gradients from the layers before it. For very deep networks, this can lead to exponentially small gradients if the derivative values are less than 1, causing earlier layers to stop learning effectively.

Mathematically, the gradient of the loss \(L\) with respect to the weights \(W_n\) of layer \(n\) is given by:

\[
\frac{\partial L}{\partial W_n} = \frac{\partial L}{\partial h_{n+1}} \cdot \frac{\partial h_{n+1}}{\partial h_n} \cdot \frac{\partial h_n}{\partial W_n}
\]

As the network depth increases, \(\frac{\partial h_{n+1}}{\partial h_n}\) can become very small, leading to vanishing gradients.

\paragraph{How Does ResNet Solve This?}

ResNet addresses this problem by introducing \textbf{skip connections}, or shortcut connections, that allow the gradient to bypass certain layers, helping to preserve the gradient signal during backpropagation. This is achieved by learning a \textbf{residual function} \cite{he2015deepresiduallearningimage}:

\[
y = F(x, \{W_i\}) + x
\]

Here, \(x\) is the input to a residual block, \(F(x, \{W_i\})\) is the residual mapping to be learned, and \(y\) is the output of the block. The key idea is that the network only needs to learn the residual function \(F(x, \{W_i\})\), making it easier to optimize deep networks. This allows deeper networks to be trained without the risk of vanishing gradients.

\paragraph{Mathematical Explanation of the Residual Block}

In a standard deep neural network, the mapping learned by each layer is represented as \(y = H(x)\), where \(H(x)\) is the transformation applied to the input \(x\). In ResNet, the network instead learns the residual mapping, \(F(x) = H(x) - x\), and the final output becomes:

\[
y = F(x) + x
\]

This skip connection ensures that if \(F(x)\) is close to zero, the model can still learn the identity mapping, which helps prevent degradation of the signal in deeper layers. The residual block is particularly useful for very deep networks, as it enables effective training of networks with over 100 layers.

\textbf{Residual Block Diagram}

Below is a diagram of the basic residual block used in ResNet:

\begin{center}
\begin{tikzpicture}[node distance=2.0cm, auto,]
    % Nodes
    \node (input) [rectangle, draw, text centered, minimum height=1em, minimum width=3cm] {Input $x$};
    \node (conv1) [rectangle, draw, below of=input, text centered, minimum height=1em, minimum width=3cm] {Conv3x3 + BN + ReLU};
    \node (conv2) [rectangle, draw, below of=conv1, text centered, minimum height=1em, minimum width=3cm] {Conv3x3 + BN};
    \node (output) [rectangle, draw, below of=conv2, text centered, minimum height=1em, minimum width=3cm] {Output $y = F(x) + x$};
    
    % Connections
    \draw[->] (input) -- (conv1);
    \draw[->] (conv1) -- (conv2);
    \draw[->] (conv2) -- (output);
    
    % Shortcut Connection
    \draw[->] (input.east) -- ++(1.5,0) -- ++(0,-4) -- (output.east);
    
    % Add label
    \node at (3.5,-2) {+};
\end{tikzpicture}
\end{center}

In this diagram:
- \(x\) is the input.
- The main path applies two convolutional layers (Conv3x3), batch normalization (BN), and ReLU activation functions.
- The shortcut path adds the input \(x\) directly to the output of the convolutional layers, creating the residual connection.

\textbf{Comparison of ResNet Architectures}

The table below compares the key components of various ResNet architectures. The difference between the models lies in the number of residual blocks, with deeper models using more residual blocks.

\begin{center}
\begin{tabular}{|c|c|c|c|c|c|}
\hline
\textbf{Component} & \textbf{ResNet-18} & \textbf{ResNet-34} & \textbf{ResNet-50} & \textbf{ResNet-101} & \textbf{ResNet-152} \\
\hline
Conv3-64 & 1 & 1 & 1 & 1 & 1 \\
Residual Blocks (3x3) & 8 & 16 & 16 (bottleneck) & 33 (bottleneck) & 50 (bottleneck) \\
Conv3-128 & 1 & 1 & 1 & 1 & 1 \\
Conv3-256 & 1 & 1 & 1 & 1 & 1 \\
Conv3-512 & 1 & 1 & 1 & 1 & 1 \\
MaxPooling & 1 & 1 & 1 & 1 & 1 \\
\hline
FC-1000 (Softmax) & \multicolumn{5}{c|}{1 Fully Connected Layer (1000 categories)} \\
\hline
\end{tabular}
\end{center}

\paragraph{TensorFlow Code for ResNet-50}

Here is the implementation of the ResNet-50 model using TensorFlow. ResNet-50 uses bottleneck residual blocks to optimize depth while reducing computational cost.

\begin{lstlisting}[style=python]
import tensorflow as tf

# Define a residual block with bottleneck
def residual_block(input_tensor, filters, strides=1):
    x = tf.keras.layers.Conv2D(filters, (1, 1), strides=strides, padding='same')(input_tensor)
    x = tf.keras.layers.BatchNormalization()(x)
    x = tf.keras.layers.Activation('relu')(x)

    x = tf.keras.layers.Conv2D(filters, (3, 3), padding='same')(x)
    x = tf.keras.layers.BatchNormalization()(x)
    x = tf.keras.layers.Activation('relu')(x)

    x = tf.keras.layers.Conv2D(4 * filters, (1, 1), padding='same')(x)
    x = tf.keras.layers.BatchNormalization()(x)

    # Shortcut connection
    shortcut = tf.keras.layers.Conv2D(4 * filters, (1, 1), strides=strides, padding='same')(input_tensor)
    shortcut = tf.keras.layers.BatchNormalization()(shortcut)

    # Add shortcut and main path
    x = tf.keras.layers.Add()([x, shortcut])
    x = tf.keras.layers.Activation('relu')(x)

    return x

# Define the ResNet-50 architecture
def ResNet50(input_shape=(224, 224, 3), num_classes=1000):
    input_tensor = tf.keras.Input(shape=input_shape)

    # Initial conv and maxpool layers
    x = tf.keras.layers.Conv2D(64, (7, 7), strides=2, padding='same')(input_tensor)
    x = tf.keras.layers.BatchNormalization()(x)
    x = tf.keras.layers.Activation('relu')(x)
    x = tf.keras.layers.MaxPooling2D((3, 3), strides=2, padding='same')(x)

    # Residual blocks
    x = residual_block(x, 64)
    x = residual_block(x, 64)
    x = residual_block(x, 64)
    
    x = residual_block(x, 128, strides=2)
    x = residual_block(x, 128)
    x = residual_block(x, 128)
    x = residual_block(x, 128)

    x = residual_block(x, 256, strides=2)
    x = residual_block(x, 256)
    x = residual_block(x, 256)
    x = residual_block(x, 256)
    x = residual_block(x, 256)
    x = residual_block(x, 256)

    x = residual_block(x, 512, strides=2)
    x = residual_block(x, 512)
    x = residual_block(x, 512)

    # Global Average Pooling and Output
    x = tf.keras.layers.GlobalAveragePooling2D()(x)
    output_tensor = tf.keras.layers.Dense(num_classes, activation='softmax')(x)

    model = tf.keras.Model(inputs=input_tensor, outputs=output_tensor)
    
    return model

# Create the ResNet-50 model
model = ResNet50()
model.summary()
\end{lstlisting}

\paragraph{Key Insights for Beginners}

\textbf{What is a residual block?} A residual block is the core building unit of ResNet. It allows the output of a layer to be added to its input, effectively bypassing the layer. This helps the network learn identity mappings and prevents degradation as the network depth increases.

\textbf{Why use Bottleneck Blocks?} Bottleneck blocks (1x1, 3x3, 1x1 convolutions) are used in deeper ResNet variants like ResNet-50 and beyond. They reduce the number of parameters while still allowing the network to capture complex features.

\textbf{Fully Connected Layers}: The final fully connected layer is responsible for classification, outputting probabilities for each class in the dataset.

\subsection{ResNetV1}

ResNetV1 is a widely used deep learning architecture that introduced residual learning. In this section, we explore different versions of ResNetV1 (ResNet50, ResNet101, ResNet152) and apply both Linear Probe and Fine-tuning techniques on the CIFAR-10 dataset.

\subsubsection{ResNet50}
ResNet50 is a 50-layer version of ResNet, and we will perform both Linear Probe and Fine-tuning on this architecture.

\paragraph{Linear Probe}
In the Linear Probe approach, we freeze the pre-trained ResNet50 model’s convolutional layers and train only the classification layers on the CIFAR-10 dataset.

\begin{lstlisting}[style=python]
import tensorflow as tf
from tensorflow.keras import layers, models
from tensorflow.keras.applications import ResNet50
from tensorflow.keras.datasets import cifar10
from tensorflow.keras.utils import to_categorical

# Load and preprocess CIFAR-10 dataset
(x_train, y_train), (x_test, y_test) = cifar10.load_data()
x_train = tf.image.resize(x_train, (224, 224)) / 255.0
x_test = tf.image.resize(x_test, (224, 224)) / 255.0
y_train = to_categorical(y_train, 10)
y_test = to_categorical(y_test, 10)

# Load pre-trained ResNet50 model without the top classification layer
base_model_resnet50 = ResNet50(weights='imagenet', include_top=False, input_shape=(224, 224, 3))

# Freeze all layers of the base model
for layer in base_model_resnet50.layers:
    layer.trainable = False

# Add classification layers
model_resnet50 = models.Sequential([
    base_model_resnet50,
    layers.GlobalAveragePooling2D(),
    layers.Dense(256, activation='relu'),
    layers.Dense(10, activation='softmax')
])

# Compile the model
model_resnet50.compile(optimizer='adam', loss='categorical_crossentropy', metrics=['accuracy'])

# Train the model for linear probe
history_resnet50 = model_resnet50.fit(x_train, y_train, epochs=10, validation_data=(x_test, y_test))

# Evaluate the model
results_resnet50 = model_resnet50.evaluate(x_test, y_test)
print(f"ResNet50 Linear Probe Test Accuracy: {results_resnet50[1]}")
\end{lstlisting}

This code performs a Linear Probe by freezing the pre-trained layers of ResNet50 and training only the newly added classification layers on CIFAR-10.

\paragraph{Fine-tuning}
In Fine-tuning, we unfreeze some layers of the ResNet50 model and train them along with the classification layers to allow better adaptation to the CIFAR-10 dataset.

\begin{lstlisting}[style=python]
import tensorflow as tf
from tensorflow.keras import layers, models
from tensorflow.keras.applications import ResNet50
from tensorflow.keras.datasets import cifar10
from tensorflow.keras.utils import to_categorical

# Load and preprocess CIFAR-10 dataset
(x_train, y_train), (x_test, y_test) = cifar10.load_data()
x_train = tf.image.resize(x_train, (224, 224)) / 255.0
x_test = tf.image.resize(x_test, (224, 224)) / 255.0
y_train = to_categorical(y_train, 10)
y_test = to_categorical(y_test, 10)

# Load pre-trained ResNet50 model without the top classification layer
base_model_resnet50 = ResNet50(weights='imagenet', include_top=False, input_shape=(224, 224, 3))

# Unfreeze the last few layers for fine-tuning
for layer in base_model_resnet50.layers[-5:]:
    layer.trainable = True

# Add classification layers
model_resnet50 = models.Sequential([
    base_model_resnet50,
    layers.GlobalAveragePooling2D(),
    layers.Dense(256, activation='relu'),
    layers.Dense(10, activation='softmax')
])

# Compile the model with a smaller learning rate
model_resnet50.compile(optimizer=tf.keras.optimizers.Adam(1e-5), loss='categorical_crossentropy', metrics=['accuracy'])

# Fine-tune the model
fine_tune_history_resnet50 = model_resnet50.fit(x_train, y_train, epochs=10, validation_data=(x_test, y_test))

# Evaluate the model
fine_tune_results_resnet50 = model_resnet50.evaluate(x_test, y_test)
print(f"ResNet50 Fine-tuning Test Accuracy: {fine_tune_results_resnet50[1]}")
\end{lstlisting}

This code performs fine-tuning on ResNet50 by unfreezing the last few layers and training them with a smaller learning rate to adapt the pre-trained weights to CIFAR-10.

\subsubsection{ResNet101}
ResNet101 is a deeper version of ResNet with 101 layers. We will perform both Linear Probe and Fine-tuning on this model.

\paragraph{Linear Probe}
We freeze all the convolutional layers of the ResNet101 model and train only the classification layers.

\begin{lstlisting}[style=python]
import tensorflow as tf
from tensorflow.keras import layers, models
from tensorflow.keras.applications import ResNet101
from tensorflow.keras.datasets import cifar10
from tensorflow.keras.utils import to_categorical

# Load and preprocess CIFAR-10 dataset
(x_train, y_train), (x_test, y_test) = cifar10.load_data()
x_train = tf.image.resize(x_train, (224, 224)) / 255.0
x_test = tf.image.resize(x_test, (224, 224)) / 255.0
y_train = to_categorical(y_train, 10)
y_test = to_categorical(y_test, 10)

# Load pre-trained ResNet101 model without the top classification layer
base_model_resnet101 = ResNet101(weights='imagenet', include_top=False, input_shape=(224, 224, 3))

# Freeze all layers of the base model
for layer in base_model_resnet101.layers:
    layer.trainable = False

# Add classification layers
model_resnet101 = models.Sequential([
    base_model_resnet101,
    layers.GlobalAveragePooling2D(),
    layers.Dense(256, activation='relu'),
    layers.Dense(10, activation='softmax')
])

# Compile the model
model_resnet101.compile(optimizer='adam', loss='categorical_crossentropy', metrics=['accuracy'])

# Train the model for linear probe
history_resnet101 = model_resnet101.fit(x_train, y_train, epochs=10, validation_data=(x_test, y_test))

# Evaluate the model
results_resnet101 = model_resnet101.evaluate(x_test, y_test)
print(f"ResNet101 Linear Probe Test Accuracy: {results_resnet101[1]}")
\end{lstlisting}

This code freezes the ResNet101 model's pre-trained layers and trains only the classification layers using CIFAR-10, similar to the approach for ResNet50.

\paragraph{Fine-tuning}
We unfreeze some layers of ResNet101 and train the entire model to fine-tune the pre-trained weights.

\begin{lstlisting}[style=python]
import tensorflow as tf
from tensorflow.keras import layers, models
from tensorflow.keras.applications import ResNet101
from tensorflow.keras.datasets import cifar10
from tensorflow.keras.utils import to_categorical

# Load and preprocess CIFAR-10 dataset
(x_train, y_train), (x_test, y_test) = cifar10.load_data()
x_train = tf.image.resize(x_train, (224, 224)) / 255.0
x_test = tf.image.resize(x_test, (224, 224)) / 255.0
y_train = to_categorical(y_train, 10)
y_test = to_categorical(y_test, 10)

# Load pre-trained ResNet101 model without the top classification layer
base_model_resnet101 = ResNet101(weights='imagenet', include_top=False, input_shape=(224, 224, 3))

# Unfreeze the last few layers for fine-tuning
for layer in base_model_resnet101.layers[-5:]:
    layer.trainable = True

# Add classification layers
model_resnet101 = models.Sequential([
    base_model_resnet101,
    layers.GlobalAveragePooling2D(),
    layers.Dense(256, activation='relu'),
    layers.Dense(10, activation='softmax')
])

# Compile the model with a smaller learning rate
model_resnet101.compile(optimizer=tf.keras.optimizers.Adam(1e-5), loss='categorical_crossentropy', metrics=['accuracy'])

# Fine-tune the model
fine_tune_history_resnet101 = model_resnet101.fit(x_train, y_train, epochs=10, validation_data=(x_test, y_test))

# Evaluate the model
fine_tune_results_resnet101 = model_resnet101.evaluate(x_test, y_test)
print(f"ResNet101 Fine-tuning Test Accuracy: {fine_tune_results_resnet101[1]}")
\end{lstlisting}

This code fine-tunes the ResNet101 model by unfreezing some layers and training with a smaller learning rate, similar to the fine-tuning process for ResNet50.

\subsubsection{ResNet152}
ResNet152 is the deepest model in the ResNetV1 series. We will apply both Linear Probe and Fine-tuning to this model.

\paragraph{Linear Probe}
We freeze all convolutional layers of ResNet152 and train only the classification layers.

\begin{lstlisting}[style=python]
import tensorflow as tf
from tensorflow.keras import layers, models
from tensorflow.keras.applications import ResNet152
from tensorflow.keras.datasets import cifar10
from tensorflow.keras.utils import to_categorical

# Load and preprocess CIFAR-10 dataset
(x_train, y_train), (x_test, y_test) = cifar10.load_data()
x_train = tf.image.resize(x_train, (224, 224)) / 255.0
x_test = tf.image.resize(x_test, (224, 224)) / 255.0
y_train = to_categorical(y_train, 10)
y_test = to_categorical(y_test, 10)

# Load pre-trained ResNet152 model without the top classification layer
base_model_resnet152 = ResNet152(weights='imagenet', include_top=False, input_shape=(224, 224, 3))

# Freeze all layers of the base model
for layer in base_model_resnet152.layers:
    layer.trainable = False

# Add classification layers
model_resnet152 = models.Sequential([
    base_model_resnet152,
    layers.GlobalAveragePooling2D(),
    layers.Dense(256, activation='relu'),
    layers.Dense(10, activation='softmax')
])

# Compile the model
model_resnet152.compile(optimizer='adam', loss='categorical_crossentropy', metrics=['accuracy'])

# Train the model for linear probe
history_resnet152 = model_resnet152.fit(x_train, y_train, epochs=10, validation_data=(x_test, y_test))

# Evaluate the model
results_resnet152 = model_resnet152.evaluate(x_test, y_test)
print(f"ResNet152 Linear Probe Test Accuracy: {results_resnet152[1]}")
\end{lstlisting}

This code applies Linear Probe to ResNet152 by freezing the convolutional layers and training the new classification head.

\paragraph{Fine-tuning}
We unfreeze some layers of ResNet152 and train the entire model for fine-tuning.

\begin{lstlisting}[style=python]
import tensorflow as tf
from tensorflow.keras import layers, models
from tensorflow.keras.applications import ResNet152
from tensorflow.keras.datasets import cifar10
from tensorflow.keras.utils import to_categorical

# Load and preprocess CIFAR-10 dataset
(x_train, y_train), (x_test, y_test) = cifar10.load_data()
x_train = tf.image.resize(x_train, (224, 224)) / 255.0
x_test = tf.image.resize(x_test, (224, 224)) / 255.0
y_train = to_categorical(y_train, 10)
y_test = to_categorical(y_test, 10)

# Load pre-trained ResNet152 model without the top classification layer
base_model_resnet152 = ResNet152(weights='imagenet', include_top=False, input_shape=(224, 224, 3))

# Unfreeze the last few layers for fine-tuning
for layer in base_model_resnet152.layers[-5:]:
    layer.trainable = True

# Add classification layers
model_resnet152 = models.Sequential([
    base_model_resnet152,
    layers.GlobalAveragePooling2D(),
    layers.Dense(256, activation='relu'),
    layers.Dense(10, activation='softmax')
])

# Compile the model with a smaller learning rate
model_resnet152.compile(optimizer=tf.keras.optimizers.Adam(1e-5), loss='categorical_crossentropy', metrics=['accuracy'])

# Fine-tune the model
fine_tune_history_resnet152 = model_resnet152.fit(x_train, y_train, epochs=10, validation_data=(x_test, y_test))

# Evaluate the model
fine_tune_results_resnet152 = model_resnet152.evaluate(x_test, y_test)
print(f"ResNet152 Fine-tuning Test Accuracy: {fine_tune_results_resnet152[1]}")
\end{lstlisting}

This code fine-tunes the ResNet152 model by unfreezing the last few layers and training the entire model to adapt it to the CIFAR-10 dataset.

\subsection{ResNetV2}

ResNetV2 is an improved version of the ResNet architecture, introducing pre-activation residual blocks. In this section, we will explore the use of ResNetV2 variants (ResNet50V2, ResNet101V2, and ResNet152V2) for both Linear Probe and Fine-tuning using the CIFAR-10 dataset \cite{he2015deepresiduallearningimage, DBLP:journals/corr/HeZR016}, .

\subsubsection{ResNet50V2}
ResNet50V2 is the 50-layer variant of ResNetV2. Below are implementations for both Linear Probe and Fine-tuning.

\paragraph{Linear Probe}
In the Linear Probe approach, we freeze the convolutional layers of ResNet50V2 and train only the classification layers on CIFAR-10.

\begin{lstlisting}[style=python]
import tensorflow as tf
from tensorflow.keras import layers, models
from tensorflow.keras.applications import ResNet50V2
from tensorflow.keras.datasets import cifar10
from tensorflow.keras.utils import to_categorical

# Load and preprocess CIFAR-10 dataset
(x_train, y_train), (x_test, y_test) = cifar10.load_data()
x_train = tf.image.resize(x_train, (224, 224)) / 255.0
x_test = tf.image.resize(x_test, (224, 224)) / 255.0
y_train = to_categorical(y_train, 10)
y_test = to_categorical(y_test, 10)

# Load pre-trained ResNet50V2 model without the top classification layer
base_model_resnet50v2 = ResNet50V2(weights='imagenet', include_top=False, input_shape=(224, 224, 3))

# Freeze all layers of the base model
for layer in base_model_resnet50v2.layers:
    layer.trainable = False

# Add classification layers
model_resnet50v2 = models.Sequential([
    base_model_resnet50v2,
    layers.GlobalAveragePooling2D(),
    layers.Dense(256, activation='relu'),
    layers.Dense(10, activation='softmax')
])

# Compile the model
model_resnet50v2.compile(optimizer='adam', loss='categorical_crossentropy', metrics=['accuracy'])

# Train the model for linear probe
history_resnet50v2 = model_resnet50v2.fit(x_train, y_train, epochs=10, validation_data=(x_test, y_test))

# Evaluate the model
results_resnet50v2 = model_resnet50v2.evaluate(x_test, y_test)
print(f"ResNet50V2 Linear Probe Test Accuracy: {results_resnet50v2[1]}")
\end{lstlisting}

In this code, we freeze all the convolutional layers of ResNet50V2 and train only the newly added classification layers on the CIFAR-10 dataset. After training, the model's test accuracy is printed.

\paragraph{Fine-tuning}
For Fine-tuning, we unfreeze the last few layers of ResNet50V2 and train them along with the classification layers to allow better adaptation to the CIFAR-10 dataset.

\begin{lstlisting}[style=python]
import tensorflow as tf
from tensorflow.keras import layers, models
from tensorflow.keras.applications import ResNet50V2
from tensorflow.keras.datasets import cifar10
from tensorflow.keras.utils import to_categorical

# Load and preprocess CIFAR-10 dataset
(x_train, y_train), (x_test, y_test) = cifar10.load_data()
x_train = tf.image.resize(x_train, (224, 224)) / 255.0
x_test = tf.image.resize(x_test, (224, 224)) / 255.0
y_train = to_categorical(y_train, 10)
y_test = to_categorical(y_test, 10)

# Load pre-trained ResNet50V2 model without the top classification layer
base_model_resnet50v2 = ResNet50V2(weights='imagenet', include_top=False, input_shape=(224, 224, 3))

# Unfreeze the last few layers for fine-tuning
for layer in base_model_resnet50v2.layers[-5:]:
    layer.trainable = True

# Add classification layers
model_resnet50v2 = models.Sequential([
    base_model_resnet50v2,
    layers.GlobalAveragePooling2D(),
    layers.Dense(256, activation='relu'),
    layers.Dense(10, activation='softmax')
])

# Compile the model with a smaller learning rate
model_resnet50v2.compile(optimizer=tf.keras.optimizers.Adam(1e-5), loss='categorical_crossentropy', metrics=['accuracy'])

# Fine-tune the model
fine_tune_history_resnet50v2 = model_resnet50v2.fit(x_train, y_train, epochs=10, validation_data=(x_test, y_test))

# Evaluate the model
fine_tune_results_resnet50v2 = model_resnet50v2.evaluate(x_test, y_test)
print(f"ResNet50V2 Fine-tuning Test Accuracy: {fine_tune_results_resnet50v2[1]}")
\end{lstlisting}

In this code, we unfreeze the last few layers of ResNet50V2 for fine-tuning and train the entire model with a smaller learning rate. After fine-tuning, the test accuracy is printed.

\subsubsection{ResNet101V2}
ResNet101V2 is a deeper version of ResNetV2 with 101 layers. Below are the implementations for both Linear Probe and Fine-tuning \cite{he2015deepresiduallearningimage, DBLP:journals/corr/HeZR016}.

\paragraph{Linear Probe}
We freeze all layers of ResNet101V2 and train only the classification layers on CIFAR-10.

\begin{lstlisting}[style=python]
import tensorflow as tf
from tensorflow.keras import layers, models
from tensorflow.keras.applications import ResNet101V2
from tensorflow.keras.datasets import cifar10
from tensorflow.keras.utils import to_categorical

# Load and preprocess CIFAR-10 dataset
(x_train, y_train), (x_test, y_test) = cifar10.load_data()
x_train = tf.image.resize(x_train, (224, 224)) / 255.0
x_test = tf.image.resize(x_test, (224, 224)) / 255.0
y_train = to_categorical(y_train, 10)
y_test = to_categorical(y_test, 10)

# Load pre-trained ResNet101V2 model without the top classification layer
base_model_resnet101v2 = ResNet101V2(weights='imagenet', include_top=False, input_shape=(224, 224, 3))

# Freeze all layers of the base model
for layer in base_model_resnet101v2.layers:
    layer.trainable = False

# Add classification layers
model_resnet101v2 = models.Sequential([
    base_model_resnet101v2,
    layers.GlobalAveragePooling2D(),
    layers.Dense(256, activation='relu'),
    layers.Dense(10, activation='softmax')
])

# Compile the model
model_resnet101v2.compile(optimizer='adam', loss='categorical_crossentropy', metrics=['accuracy'])

# Train the model for linear probe
history_resnet101v2 = model_resnet101v2.fit(x_train, y_train, epochs=10, validation_data=(x_test, y_test))

# Evaluate the model
results_resnet101v2 = model_resnet101v2.evaluate(x_test, y_test)
print(f"ResNet101V2 Linear Probe Test Accuracy: {results_resnet101v2[1]}")
\end{lstlisting}

This code freezes all layers of ResNet101V2 and trains only the classification layers. The test accuracy is printed after training.

\paragraph{Fine-tuning}
For Fine-tuning, we unfreeze the last few layers of ResNet101V2 and train them along with the classification layers.

\begin{lstlisting}[style=python]
import tensorflow as tf
from tensorflow.keras import layers, models
from tensorflow.keras.applications import ResNet101V2
from tensorflow.keras.datasets import cifar10
from tensorflow.keras.utils import to_categorical

# Load and preprocess CIFAR-10 dataset
(x_train, y_train), (x_test, y_test) = cifar10.load_data()
x_train = tf.image.resize(x_train, (224, 224)) / 255.0
x_test = tf.image.resize(x_test, (224, 224)) / 255.0
y_train = to_categorical(y_train, 10)
y_test = to_categorical(y_test, 10)

# Load pre-trained ResNet101V2 model without the top classification layer
base_model_resnet101v2 = ResNet101V2(weights='imagenet', include_top=False, input_shape=(224, 224, 3))

# Unfreeze the last few layers for fine-tuning
for layer in base_model_resnet101v2.layers[-5:]:
    layer.trainable = True

# Add classification layers
model_resnet101v2 = models.Sequential([
    base_model_resnet101v2,
    layers.GlobalAveragePooling2D(),
    layers.Dense(256, activation='relu'),
    layers.Dense(10, activation='softmax')
])

# Compile the model with a smaller learning rate
model_resnet101v2.compile(optimizer=tf.keras.optimizers.Adam(1e-5), loss='categorical_crossentropy', metrics=['accuracy'])

# Fine-tune the model
fine_tune_history_resnet101v2 = model_resnet101v2.fit(x_train, y_train, epochs=10, validation_data=(x_test, y_test))

# Evaluate the model
fine_tune_results_resnet101v2 = model_resnet101v2.evaluate(x_test, y_test)
print(f"ResNet101V2 Fine-tuning Test Accuracy: {fine_tune_results_resnet101v2[1]}")
\end{lstlisting}

In this code, we unfreeze the last few layers of ResNet101V2 and fine-tune the entire model with a smaller learning rate. After fine-tuning, the test accuracy is printed.

\subsubsection{ResNet152V2}
ResNet152V2 is the deepest model in the ResNetV2 family. Below are implementations for Linear Probe and Fine-tuning \cite{he2015deepresiduallearningimage, DBLP:journals/corr/HeZR016}.

\paragraph{Linear Probe}
We freeze all layers of ResNet152V2 and train only the classification layers.

\begin{lstlisting}[style=python]
import tensorflow as tf
from tensorflow.keras import layers, models
from tensorflow.keras.applications import ResNet152V2
from tensorflow.keras.datasets import cifar10
from tensorflow.keras.utils import to_categorical

# Load and preprocess CIFAR-10 dataset
(x_train, y_train), (x_test, y_test) = cifar10.load_data()
x_train = tf.image.resize(x_train, (224, 224)) / 255.0
x_test = tf.image.resize(x_test, (224, 224)) / 255.0
y_train = to_categorical(y_train, 10)
y_test = to_categorical(y_test, 10)

# Load pre-trained ResNet152V2 model without the top classification layer
base_model_resnet152v2 = ResNet152V2(weights='imagenet', include_top=False, input_shape=(224, 224, 3))

# Freeze all layers of the base model
for layer in base_model_resnet152v2.layers:
    layer.trainable = False

# Add classification layers
model_resnet152v2 = models.Sequential([
    base_model_resnet152v2,
    layers.GlobalAveragePooling2D(),
    layers.Dense(256, activation='relu'),
    layers.Dense(10, activation='softmax')
])

# Compile the model
model_resnet152v2.compile(optimizer='adam', loss='categorical_crossentropy', metrics=['accuracy'])

# Train the model for linear probe
history_resnet152v2 = model_resnet152v2.fit(x_train, y_train, epochs=10, validation_data=(x_test, y_test))

# Evaluate the model
results_resnet152v2 = model_resnet152v2.evaluate(x_test, y_test)
print(f"ResNet152V2 Linear Probe Test Accuracy: {results_resnet152v2[1]}")
\end{lstlisting}

In this Linear Probe approach, we freeze all layers of ResNet152V2 and train the classification layers. The test accuracy is printed after training.

\paragraph{Fine-tuning}
For Fine-tuning, we unfreeze the last few layers of ResNet152V2 and train them along with the classification layers.

\begin{lstlisting}[style=python]
import tensorflow as tf
from tensorflow.keras import layers, models
from tensorflow.keras.applications import ResNet152V2
from tensorflow.keras.datasets import cifar10
from tensorflow.keras.utils import to_categorical

# Load and preprocess CIFAR-10 dataset
(x_train, y_train), (x_test, y_test) = cifar10.load_data()
x_train = tf.image.resize(x_train, (224, 224)) / 255.0
x_test = tf.image.resize(x_test, (224, 224)) / 255.0
y_train = to_categorical(y_train, 10)
y_test = to_categorical(y_test, 10)

# Load pre-trained ResNet152V2 model without the top classification layer
base_model_resnet152v2 = ResNet152V2(weights='imagenet', include_top=False, input_shape=(224, 224, 3))

# Unfreeze the last few layers for fine-tuning
for layer in base_model_resnet152v2.layers[-5:]:
    layer.trainable = True

# Add classification layers
model_resnet152v2 = models.Sequential([
    base_model_resnet152v2,
    layers.GlobalAveragePooling2D(),
    layers.Dense(256, activation='relu'),
    layers.Dense(10, activation='softmax')
])

# Compile the model with a smaller learning rate
model_resnet152v2.compile(optimizer=tf.keras.optimizers.Adam(1e-5), loss='categorical_crossentropy', metrics=['accuracy'])

# Fine-tune the model
fine_tune_history_resnet152v2 = model_resnet152v2.fit(x_train, y_train, epochs=10, validation_data=(x_test, y_test))

# Evaluate the model
fine_tune_results_resnet152v2 = model_resnet152v2.evaluate(x_test, y_test)
print(f"ResNet152V2 Fine-tuning Test Accuracy: {fine_tune_results_resnet152v2[1]}")
\end{lstlisting}

This code fine-tunes the ResNet152V2 model by unfreezing the last few layers and training the entire model with a smaller learning rate. After fine-tuning, the test accuracy is printed.

\section{DenseNet (2017)}

Densely Connected Convolutional Networks, or DenseNet, was introduced in 2017 by Huang et al. in the paper \textit{"Densely Connected Convolutional Networks"}. DenseNet addresses the vanishing gradient problem and improves feature reuse and gradient flow by introducing dense connections between layers. DenseNet is efficient in parameter usage and achieves high performance with fewer parameters compared to traditional networks \cite{huang2018denselyconnectedconvolutionalnetworks}.

\paragraph{Dense Connections}

The key idea behind DenseNet is the introduction of \textbf{dense connections}, where each layer receives inputs from all preceding layers and passes its own output to all subsequent layers. This ensures that the feature maps learned by earlier layers are reused across the entire network, allowing for more compact models and efficient parameter usage.

In a typical convolutional network, each layer only receives input from the previous layer. However, in DenseNet, each layer receives the concatenation of the outputs of all previous layers. Mathematically, the output of the \(l\)-th layer is given by:

\[
x_l = H_l([x_0, x_1, \dots, x_{l-1}])
\]

where \([x_0, x_1, \dots, x_{l-1}]\) represents the concatenation of the feature maps from layers \(0\) to \(l-1\), and \(H_l\) denotes the operations applied at layer \(l\) (typically batch normalization, ReLU, and convolution).

\textbf{Advantages of Dense Connections:}
\begin{itemize}
    \item Feature Reuse: Dense connections ensure that features are reused across layers, making the network more efficient and reducing the number of parameters.
    \item Efficient Gradient Flow: The dense connections improve gradient flow through the network, which helps in training very deep networks.
    \item Parameter Efficiency: Since layers can reuse features from earlier layers, DenseNet achieves high performance with fewer parameters compared to other architectures such as ResNet.
\end{itemize}

\paragraph{Mathematical Explanation of Dense Blocks}

In DenseNet, the network is divided into \textbf{dense blocks}. Inside each dense block, every layer is connected to every other layer, and the output of each layer is concatenated with the inputs to the subsequent layers. Each layer performs the following operations:

\[
x_l = H_l([x_0, x_1, \dots, x_{l-1}])
\]

where \(H_l\) consists of batch normalization, ReLU activation, and a \(3 \times 3\) convolution.

\paragraph{DenseNet Architecture}

DenseNet is composed of several dense blocks. Each dense block is followed by a \textbf{transition layer}, which reduces the number of feature maps through a \(1 \times 1\) convolution and reduces the spatial dimensions via \(2 \times 2\) average pooling. The number of new feature maps added by each layer is controlled by a hyperparameter called the \textbf{growth rate} \(k\).

\textbf{The basic structure of a DenseNet network includes:}
\begin{itemize}
    \item Dense Blocks: Each dense block consists of multiple densely connected layers.
    \item Transition Layers: Transition layers are placed between dense blocks to downsample the feature maps.
    \item Growth Rate: The number of new feature maps added by each layer is controlled by the growth rate \(k\).
\end{itemize}

\paragraph{DenseNet Module Diagram}

Below is a diagram illustrating the structure of a DenseNet module with multiple layers and dense connections:

\begin{center}
\begin{tikzpicture}[node distance=1.5cm, auto]
    % Input node
    \node (input) [rectangle, draw, text centered, minimum height=1.5em, minimum width=3.5cm] {Input $x_0$};
    
    % Layer 1
    \node (layer1) [rectangle, draw, below of=input, text centered, minimum height=1.5em, minimum width=3.5cm] {Layer 1: $H_1(x_0)$};
    \node (concat1) [rectangle, draw, below of=layer1, text centered, minimum height=1.5em, minimum width=3.5cm] {Concat: $[x_0, x_1]$};
    
    % Layer 2
    \node (layer2) [rectangle, draw, below of=concat1, text centered, minimum height=1.5em, minimum width=3.5cm] {Layer 2: $H_2([x_0, x_1])$};
    \node (concat2) [rectangle, draw, below of=layer2, text centered, minimum height=1.5em, minimum width=3.5cm] {Concat: $[x_0, x_1, x_2]$};
    
    % Layer 3
    \node (layer3) [rectangle, draw, below of=concat2, text centered, minimum height=1.5em, minimum width=3.5cm] {Layer 3: $H_3([x_0, x_1, x_2])$};
    \node (concat3) [rectangle, draw, below of=layer3, text centered, minimum height=1.5em, minimum width=3.5cm] {Concat: $[x_0, x_1, x_2, x_3]$};
    
    % Connections
    \draw[->] (input) -- (layer1);
    \draw[->] (layer1) -- (concat1);
    \draw[->] (concat1) -- (layer2);
    \draw[->] (layer2) -- (concat2);
    \draw[->] (concat2) -- (layer3);
    \draw[->] (layer3) -- (concat3);
\end{tikzpicture}
\end{center}

In this diagram:
- Each layer’s output is concatenated with the outputs of all previous layers, allowing for dense connections across the block.

\paragraph{Comparison of DenseNet Variants}

The DenseNet family includes several versions, such as DenseNet-121, DenseNet-169, DenseNet-201, and DenseNet-264. The table below compares these versions based on the number of layers and parameters.

\begin{center}
\begin{tabular}{|c|c|c|c|c|}
\hline
\textbf{Component} & \textbf{DenseNet-121} & \textbf{DenseNet-169} & \textbf{DenseNet-201} & \textbf{DenseNet-264} \\
\hline
Number of Layers & 121 & 169 & 201 & 264 \\
Growth Rate \(k\) & 32 & 32 & 32 & 32 \\
Dense Blocks & 4 & 4 & 4 & 4 \\
Parameters (Millions) & 8.0M & 14.1M & 20.0M & 33.4M \\
\hline
\end{tabular}
\end{center}

\paragraph{Detailed Comparison of Components in DenseNet Variants}

The following table compares the number of components, such as dense blocks and transition layers, across different DenseNet variants.

\begin{center}
\begin{tabular}{|c|c|c|c|c|}
\hline
\textbf{Component} & \textbf{DenseNet-121} & \textbf{DenseNet-169} & \textbf{DenseNet-201} & \textbf{DenseNet-264} \\
\hline
Input Size & $224 \times 224$ & $224 \times 224$ & $224 \times 224$ & $224 \times 224$ \\
Conv2D (7x7) & $112 \times 112$ & $112 \times 112$ & $112 \times 112$ & $112 \times 112$ \\
Max Pooling (3x3) & $56 \times 56$ & $56 \times 56$ & $56 \times 56$ & $56 \times 56$ \\
Dense Block 1 & 6 layers & 6 layers & 6 layers & 6 layers \\
Transition Layer 1 & $28 \times 28$ & $28 \times 28$ & $28 \times 28$ & $28 \times 28$ \\
Dense Block 2 & 12 layers & 12 layers & 12 layers & 12 layers \\
Transition Layer 2 & $14 \times 14$ & $14 \times 14$ & $14 \times 14$ & $14 \times 14$ \\
Dense Block 3 & 24 layers & 32 layers & 48 layers & 64 layers \\
Transition Layer 3 & $7 \times 7$ & $7 \times 7$ & $7 \times 7$ & $7 \times 7$ \\
Dense Block 4 & 16 layers & 32 layers & 32 layers & 48 layers \\
Global Average Pooling & $1 \times 1$ & $1 \times 1$ & $1 \times 1$ & $1 \times 1$ \\
Fully Connected Layer & 1000 classes & 1000 classes & 1000 classes & 1000 classes \\
\hline
\end{tabular}
\end{center}

\paragraph{TensorFlow Code for DenseNet Module}

Below is the implementation of a DenseNet module using TensorFlow \texttt{tf.keras}.

\begin{lstlisting}[style=Python]
import tensorflow as tf

# Define a dense block with multiple layers
def dense_block(input_tensor, num_layers, growth_rate):
    concat_input = input_tensor
    for i in range(num_layers):
        # Apply Batch Norm, ReLU, and Conv
        x = tf.keras.layers.BatchNormalization()(concat_input)
        x = tf.keras.layers.ReLU()(x)
        x = tf.keras.layers.Conv2D(growth_rate, (3, 3), padding='same')(x)
        # Concatenate output with previous inputs
        concat_input = tf.keras.layers.concatenate([concat_input, x])
    return concat_input

# Define the DenseNet architecture
def DenseNet(input_shape=(224, 224, 3), num_classes=1000, growth_rate=32, num_blocks=4):
    input_tensor = tf.keras.Input(shape=input_shape)
    
    # Initial Convolution and Pooling
    x = tf.keras.layers.Conv2D(64, (7, 7), strides=2, padding='same')(input_tensor)
    x = tf.keras.layers.MaxPooling2D((3, 3), strides=2, padding='same')(x)
    
    # Add Dense Blocks with Transition Layers
    for i in range(num_blocks):
        x = dense_block(x, num_layers=6, growth_rate=growth_rate)
        if i != num_blocks - 1:
            x = tf.keras.layers.BatchNormalization()(x)
            x = tf.keras.layers.Conv2D(x.shape[-1], (1, 1), padding='same')(x)
            x = tf.keras.layers.AveragePooling2D((2, 2), strides=2)(x)
    
    # Global Average Pooling and Output
    x = tf.keras.layers.GlobalAveragePooling2D()(x)
    output_tensor = tf.keras.layers.Dense(num_classes, activation='softmax')(x)
    
    model = tf.keras.Model(inputs=input_tensor, outputs=output_tensor)
    
    return model

# Create a DenseNet model
model = DenseNet()
model.summary()
\end{lstlisting}

\paragraph{Key Insights for Beginners}

\textbf{Why use Dense Connections?} Dense connections enable more efficient use of features by allowing information to be reused across layers. They also facilitate better gradient flow during backpropagation, addressing the vanishing gradient problem.

\textbf{What is the growth rate?} The growth rate controls how many new feature maps are added by each layer. A larger growth rate allows for learning more complex patterns but increases the number of parameters.

\textbf{Efficient Parameter Usage}: Despite having many layers, DenseNet requires fewer parameters than traditional deep networks because of feature reuse across the layers.

\subsection{DenseNet121}
DenseNet121 is a densely connected convolutional neural network architecture where each layer receives input from all preceding layers. In this section, we will apply both Linear Probe and Fine-tuning approaches to DenseNet121 on the CIFAR-10 dataset \cite{huang2018denselyconnectedconvolutionalnetworks}.

\paragraph{Linear Probe}
In the Linear Probe approach, we freeze the convolutional layers of the pre-trained DenseNet121 model and train only the classification layers.

\begin{lstlisting}[style=python]
import tensorflow as tf
from tensorflow.keras import layers, models
from tensorflow.keras.applications import DenseNet121
from tensorflow.keras.datasets import cifar10
from tensorflow.keras.utils import to_categorical

# Load and preprocess CIFAR-10 dataset
(x_train, y_train), (x_test, y_test) = cifar10.load_data()
x_train = tf.image.resize(x_train, (224, 224)) / 255.0
x_test = tf.image.resize(x_test, (224, 224)) / 255.0
y_train = to_categorical(y_train, 10)
y_test = to_categorical(y_test, 10)

# Load pre-trained DenseNet121 model without the top classification layer
base_model_densenet121 = DenseNet121(weights='imagenet', include_top=False, input_shape=(224, 224, 3))

# Freeze all layers of the base model
for layer in base_model_densenet121.layers:
    layer.trainable = False

# Add classification layers
model_densenet121 = models.Sequential([
    base_model_densenet121,
    layers.GlobalAveragePooling2D(),
    layers.Dense(256, activation='relu'),
    layers.Dense(10, activation='softmax')
])

# Compile the model
model_densenet121.compile(optimizer='adam', loss='categorical_crossentropy', metrics=['accuracy'])

# Train the model for linear probe
history_densenet121 = model_densenet121.fit(x_train, y_train, epochs=10, validation_data=(x_test, y_test))

# Evaluate the model
results_densenet121 = model_densenet121.evaluate(x_test, y_test)
print(f"DenseNet121 Linear Probe Test Accuracy: {results_densenet121[1]}")
\end{lstlisting}

In this code, we freeze all layers of the pre-trained DenseNet121 model and train only the newly added classification layers using the CIFAR-10 dataset resized to 224x224 pixels.

\paragraph{Fine-tuning}
For Fine-tuning, we unfreeze some layers of the DenseNet121 model and train them along with the classification layers to better adapt the model to the CIFAR-10 dataset.

\begin{lstlisting}[style=python]
import tensorflow as tf
from tensorflow.keras import layers, models
from tensorflow.keras.applications import DenseNet121
from tensorflow.keras.datasets import cifar10
from tensorflow.keras.utils import to_categorical

# Load and preprocess CIFAR-10 dataset
(x_train, y_train), (x_test, y_test) = cifar10.load_data()
x_train = tf.image.resize(x_train, (224, 224)) / 255.0
x_test = tf.image.resize(x_test, (224, 224)) / 255.0
y_train = to_categorical(y_train, 10)
y_test = to_categorical(y_test, 10)

# Load pre-trained DenseNet121 model without the top classification layer
base_model_densenet121 = DenseNet121(weights='imagenet', include_top=False, input_shape=(224, 224, 3))

# Unfreeze the last few layers for fine-tuning
for layer in base_model_densenet121.layers[-15:]:
    layer.trainable = True

# Add classification layers
model_densenet121 = models.Sequential([
    base_model_densenet121,
    layers.GlobalAveragePooling2D(),
    layers.Dense(256, activation='relu'),
    layers.Dense(10, activation='softmax')
])

# Compile the model with a smaller learning rate
model_densenet121.compile(optimizer=tf.keras.optimizers.Adam(1e-5), loss='categorical_crossentropy', metrics=['accuracy'])

# Fine-tune the model
fine_tune_history_densenet121 = model_densenet121.fit(x_train, y_train, epochs=10, validation_data=(x_test, y_test))

# Evaluate the model
fine_tune_results_densenet121 = model_densenet121.evaluate(x_test, y_test)
print(f"DenseNet121 Fine-tuning Test Accuracy: {fine_tune_results_densenet121[1]}")
\end{lstlisting}

In this fine-tuning approach, we unfreeze the last 15 layers of DenseNet121 and fine-tune the model using a smaller learning rate to update the pre-trained weights gradually.

\subsection{DenseNet169}
DenseNet169 is a deeper version of DenseNet with 169 layers. Here, we perform both Linear Probe and Fine-tuning on this model using the CIFAR-10 dataset \cite{huang2018denselyconnectedconvolutionalnetworks}.

\paragraph{Linear Probe}
We freeze all convolutional layers of DenseNet169 and train only the classification layers.

\begin{lstlisting}[style=python]
import tensorflow as tf
from tensorflow.keras import layers, models
from tensorflow.keras.applications import DenseNet169
from tensorflow.keras.datasets import cifar10
from tensorflow.keras.utils import to_categorical

# Load and preprocess CIFAR-10 dataset
(x_train, y_train), (x_test, y_test) = cifar10.load_data()
x_train = tf.image.resize(x_train, (224, 224)) / 255.0
x_test = tf.image.resize(x_test, (224, 224)) / 255.0
y_train = to_categorical(y_train, 10)
y_test = to_categorical(y_test, 10)

# Load pre-trained DenseNet169 model without the top classification layer
base_model_densenet169 = DenseNet169(weights='imagenet', include_top=False, input_shape=(224, 224, 3))

# Freeze all layers of the base model
for layer in base_model_densenet169.layers:
    layer.trainable = False

# Add classification layers
model_densenet169 = models.Sequential([
    base_model_densenet169,
    layers.GlobalAveragePooling2D(),
    layers.Dense(256, activation='relu'),
    layers.Dense(10, activation='softmax')
])

# Compile the model
model_densenet169.compile(optimizer='adam', loss='categorical_crossentropy', metrics=['accuracy'])

# Train the model for linear probe
history_densenet169 = model_densenet169.fit(x_train, y_train, epochs=10, validation_data=(x_test, y_test))

# Evaluate the model
results_densenet169 = model_densenet169.evaluate(x_test, y_test)
print(f"DenseNet169 Linear Probe Test Accuracy: {results_densenet169[1]}")
\end{lstlisting}

This code freezes all layers of DenseNet169 and trains only the classification layers on the CIFAR-10 dataset.

\paragraph{Fine-tuning}
We unfreeze some layers of DenseNet169 and train them along with the classification layers.

\begin{lstlisting}[style=python]
import tensorflow as tf
from tensorflow.keras import layers, models
from tensorflow.keras.applications import DenseNet169
from tensorflow.keras.datasets import cifar10
from tensorflow.keras.utils import to_categorical

# Load and preprocess CIFAR-10 dataset
(x_train, y_train), (x_test, y_test) = cifar10.load_data()
x_train = tf.image.resize(x_train, (224, 224)) / 255.0
x_test = tf.image.resize(x_test, (224, 224)) / 255.0
y_train = to_categorical(y_train, 10)
y_test = to_categorical(y_test, 10)

# Load pre-trained DenseNet169 model without the top classification layer
base_model_densenet169 = DenseNet169(weights='imagenet', include_top=False, input_shape=(224, 224, 3))

# Unfreeze the last few layers for fine-tuning
for layer in base_model_densenet169.layers[-15:]:
    layer.trainable = True

# Add classification layers
model_densenet169 = models.Sequential([
    base_model_densenet169,
    layers.GlobalAveragePooling2D(),
    layers.Dense(256, activation='relu'),
    layers.Dense(10, activation='softmax')
])

# Compile the model with a smaller learning rate
model_densenet169.compile(optimizer=tf.keras.optimizers.Adam(1e-5), loss='categorical_crossentropy', metrics=['accuracy'])

# Fine-tune the model
fine_tune_history_densenet169 = model_densenet169.fit(x_train, y_train, epochs=10, validation_data=(x_test, y_test))

# Evaluate the model
fine_tune_results_densenet169 = model_densenet169.evaluate(x_test, y_test)
print(f"DenseNet169 Fine-tuning Test Accuracy: {fine_tune_results_densenet169[1]}")
\end{lstlisting}

Here, we fine-tune the DenseNet169 model by unfreezing some of the convolutional layers and training them with a smaller learning rate.

\subsection{DenseNet201}
DenseNet201 is the deepest variant of the DenseNet architecture, containing 201 layers. Below, we apply both Linear Probe and Fine-tuning on this model \cite{alain2018understandingintermediatelayersusing}.

\paragraph{Linear Probe}
We freeze all convolutional layers of DenseNet201 and train only the classification layers.

\begin{lstlisting}[style=python]
import tensorflow as tf
from tensorflow.keras import layers, models
from tensorflow.keras.applications import DenseNet201
from tensorflow.keras.datasets import cifar10
from tensorflow.keras.utils import to_categorical

# Load and preprocess CIFAR-10 dataset
(x_train, y_train), (x_test, y_test) = cifar10.load_data()
x_train = tf.image.resize(x_train, (224, 224)) / 255.0
x_test = tf.image.resize(x_test, (224, 224)) / 255.0
y_train = to_categorical(y_train, 10)
y_test = to_categorical(y_test, 10)

# Load pre-trained DenseNet201 model without the top classification layer
base_model_densenet201 = DenseNet201(weights='imagenet', include_top=False, input_shape=(224, 224, 3))

# Freeze all layers of the base model
for layer in base_model_densenet201.layers:
    layer.trainable = False

# Add classification layers
model_densenet201 = models.Sequential([
    base_model_densenet201,
    layers.GlobalAveragePooling2D(),
    layers.Dense(256, activation='relu'),
    layers.Dense(10, activation='softmax')
])

# Compile the model
model_densenet201.compile(optimizer='adam', loss='categorical_crossentropy', metrics=['accuracy'])

# Train the model for linear probe
history_densenet201 = model_densenet201.fit(x_train, y_train, epochs=10, validation_data=(x_test, y_test))

# Evaluate the model
results_densenet201 = model_densenet201.evaluate(x_test, y_test)
print(f"DenseNet201 Linear Probe Test Accuracy: {results_densenet201[1]}")
\end{lstlisting}

This code performs Linear Probe by freezing the convolutional layers of DenseNet201 and training only the classification layers.

\paragraph{Fine-tuning}
For Fine-tuning, we unfreeze some layers of DenseNet201 and train them along with the classification layers.

\begin{lstlisting}[style=python]
import tensorflow as tf
from tensorflow.keras import layers, models
from tensorflow.keras.applications import DenseNet201
from tensorflow.keras.datasets import cifar10
from tensorflow.keras.utils import to_categorical

# Load and preprocess CIFAR-10 dataset
(x_train, y_train), (x_test, y_test) = cifar10.load_data()
x_train = tf.image.resize(x_train, (224, 224)) / 255.0
x_test = tf.image.resize(x_test, (224, 224)) / 255.0
y_train = to_categorical(y_train, 10)
y_test = to_categorical(y_test, 10)

# Load pre-trained DenseNet201 model without the top classification layer
base_model_densenet201 = DenseNet201(weights='imagenet', include_top=False, input_shape=(224, 224, 3))

# Unfreeze the last few layers for fine-tuning
for layer in base_model_densenet201.layers[-15:]:
    layer.trainable = True

# Add classification layers
model_densenet201 = models.Sequential([
    base_model_densenet201,
    layers.GlobalAveragePooling2D(),
    layers.Dense(256, activation='relu'),
    layers.Dense(10, activation='softmax')
])

# Compile the model with a smaller learning rate
model_densenet201.compile(optimizer=tf.keras.optimizers.Adam(1e-5), loss='categorical_crossentropy', metrics=['accuracy'])

# Fine-tune the model
fine_tune_history_densenet201 = model_densenet201.fit(x_train, y_train, epochs=10, validation_data=(x_test, y_test))

# Evaluate the model
fine_tune_results_densenet201 = model_densenet201.evaluate(x_test, y_test)
print(f"DenseNet201 Fine-tuning Test Accuracy: {fine_tune_results_densenet201[1]}")
\end{lstlisting}

In this fine-tuning approach, we unfreeze some layers of DenseNet201 and fine-tune the model using a smaller learning rate to adapt to the CIFAR-10 dataset.

\section{Xception (2017)}

Xception, which stands for \textit{Extreme Inception}, was introduced by François Chollet in 2017. The Xception architecture is an extension of the Inception architecture, where the main idea is to replace the standard Inception modules with \textbf{depthwise separable convolutions}, significantly improving model efficiency while achieving better performance. Xception has been shown to outperform InceptionV3 on the ImageNet dataset \cite{chollet2016xception}.

\paragraph{The Key Idea: Depthwise Separable Convolutions}

Xception is based on the idea of factorizing convolutions into two simpler steps: 
\begin{itemize}
    \item \textbf{Depthwise Convolution}: Applies a single convolutional filter to each input channel (feature map) separately.
    \item \textbf{Pointwise Convolution}: Follows depthwise convolution with a \(1 \times 1\) convolution that combines the output of the depthwise convolution across channels.
\end{itemize}

This is also referred to as depthwise separable convolution, as it separates the spatial filtering (depthwise convolution) from the channel-wise projection (pointwise convolution). This decomposition greatly reduces the number of parameters and computations, making the model more efficient.

Mathematically, a standard convolution applies \(N\) filters of size \(k \times k\) to an input of size \(H \times W \times D\) (height, width, and depth), producing an output of size \(H \times W \times N\). This requires \(k^2 \cdot D \cdot N\) parameters. In contrast, depthwise separable convolution splits this operation into two steps:
\begin{itemize}
    \item Depthwise convolution: \(k^2 \cdot D\)
    \item Pointwise convolution: \(D \cdot N\)
\end{itemize}

Thus, depthwise separable convolutions reduce the number of parameters from \(k^2 \cdot D \cdot N\) to \(k^2 \cdot D + D \cdot N\).

\paragraph{Xception Architecture}

The Xception architecture follows the structure of a typical convolutional neural network (CNN) but with depthwise separable convolutions replacing the standard convolutional layers. It can be divided into three main parts:
\begin{itemize}
    \item \textbf{Entry Flow}: This section contains several convolutional layers that downsample the input.
    \item \textbf{Middle Flow}: This section consists of multiple Xception modules, where each module contains depthwise separable convolutions.
    \item \textbf{Exit Flow}: This section performs the final feature extraction before classification.
\end{itemize}

The table below summarizes the key components of the Xception architecture:

\begin{center}
\begin{tabular}{|c|c|c|}
\hline
\textbf{Component} & \textbf{Type} & \textbf{Output Size} \\
\hline
Input & - & $299 \times 299 \times 3$ \\
Conv2D & $3 \times 3$, Stride 2 & $149 \times 149 \times 32$ \\
Conv2D & $3 \times 3$, Stride 1 & $147 \times 147 \times 64$ \\
\hline
\textbf{Entry Flow} & 3 Xception Modules & $19 \times 19 \times 728$ \\
\hline
\textbf{Middle Flow} & 8 Xception Modules & $19 \times 19 \times 728$ \\
\hline
\textbf{Exit Flow} & 1 Xception Module & $10 \times 10 \times 1024$ \\
\hline
Global Average Pooling & - & $1 \times 1 \times 2048$ \\
Fully Connected & Softmax & 1000 \\
\hline
\end{tabular}
\end{center}

\paragraph{Xception Module Diagram}

The Xception module is the key building block of the architecture. It replaces the standard convolution layers found in other architectures with depthwise separable convolutions. Below is a simplified diagram of the Xception module:

\begin{center}
\begin{tikzpicture}[node distance=1.5cm, auto]
    % Input node
    \node (input) [rectangle, draw, text centered, minimum height=1.5em, minimum width=4cm] {Input};

    % Depthwise conv
    \node (depthwise) [rectangle, draw, below of=input, text centered, minimum height=1.5em, minimum width=4cm] {Depthwise Conv};

    % Pointwise conv
    \node (pointwise) [rectangle, draw, below of=depthwise, text centered, minimum height=1.5em, minimum width=4cm] {Pointwise Conv (1x1)};

    % Batch Norm
    \node (batchnorm) [rectangle, draw, below of=pointwise, text centered, minimum height=1.5em, minimum width=4cm] {Batch Norm};

    % ReLU
    \node (relu) [rectangle, draw, below of=batchnorm, text centered, minimum height=1.5em, minimum width=4cm] {ReLU};

    % Output
    \node (output) [rectangle, draw, below of=relu, text centered, minimum height=1.5em, minimum width=4cm] {Output};

    % Connections
    \draw[->] (input) -- (depthwise);
    \draw[->] (depthwise) -- (pointwise);
    \draw[->] (pointwise) -- (batchnorm);
    \draw[->] (batchnorm) -- (relu);
    \draw[->] (relu) -- (output);
\end{tikzpicture}
\end{center}

In this diagram:
\begin{itemize}
    \item \textbf{Depthwise Convolution}: Applies separate convolutional filters to each input channel.
    \item \textbf{Pointwise Convolution}: Combines the output from the depthwise convolution using a \(1 \times 1\) convolution.
    \item \textbf{Batch Normalization and ReLU}: These layers follow the pointwise convolution for normalization and activation.
\end{itemize}

\paragraph{TensorFlow Code for Xception Module}

Below is the implementation of the Xception module using TensorFlow \texttt{tf.keras}.

\begin{lstlisting}[style=Python]
import tensorflow as tf

# Xception module with depthwise separable convolution
def xception_module(input_tensor, filters, strides=1):
    # Depthwise convolution
    x = tf.keras.layers.DepthwiseConv2D((3, 3), padding='same', strides=strides)(input_tensor)
    x = tf.keras.layers.BatchNormalization()(x)
    x = tf.keras.layers.ReLU()(x)
    
    # Pointwise convolution
    x = tf.keras.layers.Conv2D(filters, (1, 1), padding='same')(x)
    x = tf.keras.layers.BatchNormalization()(x)
    x = tf.keras.layers.ReLU()(x)
    
    return x

# Example Xception model
def Xception(input_shape=(299, 299, 3), num_classes=1000):
    input_tensor = tf.keras.Input(shape=input_shape)
    
    # Entry Flow
    x = tf.keras.layers.Conv2D(32, (3, 3), strides=2, padding='same')(input_tensor)
    x = tf.keras.layers.Conv2D(64, (3, 3), padding='same')(x)
    
    # Add Xception modules
    x = xception_module(x, 128)
    x = xception_module(x, 256)
    x = xception_module(x, 728)
    
    # Global Average Pooling and Output
    x = tf.keras.layers.GlobalAveragePooling2D()(x)
    output_tensor = tf.keras.layers.Dense(num_classes, activation='softmax')(x)
    
    model = tf.keras.Model(inputs=input_tensor, outputs=output_tensor)
    
    return model

# Create the Xception model
model = Xception()
model.summary()
\end{lstlisting}

\paragraph{Key Insights for Beginners}

\begin{itemize}
    \item \textbf{Why use Depthwise Separable Convolutions?} Depthwise separable convolutions break down the convolution operation into two simpler steps, drastically reducing the number of parameters and computations while still maintaining the ability to capture spatial patterns and interactions across channels.
    \item \textbf{What is the advantage of Xception over Inception?} Xception simplifies the Inception modules by replacing them with depthwise separable convolutions, leading to better performance and efficiency with fewer parameters.
    \item \textbf{Efficient Parameter Usage}: By using depthwise separable convolutions, Xception is able to significantly reduce the computational cost compared to traditional convolutional layers.
\end{itemize}

\paragraph{Linear Probe}
In the Linear Probe approach, we freeze the convolutional layers of the pre-trained Xception model and train only the classification layers.

\begin{lstlisting}[style=python]
import tensorflow as tf
from tensorflow.keras import layers, models
from tensorflow.keras.applications import Xception
from tensorflow.keras.datasets import cifar10
from tensorflow.keras.utils import to_categorical

# Load and preprocess CIFAR-10 dataset
(x_train, y_train), (x_test, y_test) = cifar10.load_data()
x_train = tf.image.resize(x_train, (299, 299)) / 255.0
x_test = tf.image.resize(x_test, (299, 299)) / 255.0
y_train = to_categorical(y_train, 10)
y_test = to_categorical(y_test, 10)

# Load pre-trained Xception model without the top classification layer
base_model_xception = Xception(weights='imagenet', include_top=False, input_shape=(299, 299, 3))

# Freeze all layers of the base model
for layer in base_model_xception.layers:
    layer.trainable = False

# Add classification layers
model_xception = models.Sequential([
    base_model_xception,
    layers.GlobalAveragePooling2D(),
    layers.Dense(256, activation='relu'),
    layers.Dense(10, activation='softmax')
])

# Compile the model
model_xception.compile(optimizer='adam', loss='categorical_crossentropy', metrics=['accuracy'])

# Train the model for linear probe
history_xception = model_xception.fit(x_train, y_train, epochs=10, validation_data=(x_test, y_test))

# Evaluate the model
results_xception = model_xception.evaluate(x_test, y_test)
print(f"Xception Linear Probe Test Accuracy: {results_xception[1]}")
\end{lstlisting}

In this code, the CIFAR-10 dataset is resized to 299x299 pixels to match the input size required by Xception. We freeze the convolutional layers of the pre-trained Xception model and train only the classification layers. The test accuracy after training is printed.

\paragraph{Fine-tuning}
In Fine-tuning, we unfreeze some layers of the Xception model and train them along with the classification layers to better adapt the model to the CIFAR-10 dataset.

\begin{lstlisting}[style=python]
import tensorflow as tf
from tensorflow.keras import layers, models
from tensorflow.keras.applications import Xception
from tensorflow.keras.datasets import cifar10
from tensorflow.keras.utils import to_categorical

# Load and preprocess CIFAR-10 dataset
(x_train, y_train), (x_test, y_test) = cifar10.load_data()
x_train = tf.image.resize(x_train, (299, 299)) / 255.0
x_test = tf.image.resize(x_test, (299, 299)) / 255.0
y_train = to_categorical(y_train, 10)
y_test = to_categorical(y_test, 10)

# Load pre-trained Xception model without the top classification layer
base_model_xception = Xception(weights='imagenet', include_top=False, input_shape=(299, 299, 3))

# Unfreeze the last few layers for fine-tuning
for layer in base_model_xception.layers[-30:]:
    layer.trainable = True

# Add classification layers
model_xception = models.Sequential([
    base_model_xception,
    layers.GlobalAveragePooling2D(),
    layers.Dense(256, activation='relu'),
    layers.Dense(10, activation='softmax')
])

# Compile the model with a smaller learning rate
model_xception.compile(optimizer=tf.keras.optimizers.Adam(1e-5), loss='categorical_crossentropy', metrics=['accuracy'])

# Fine-tune the model
fine_tune_history_xception = model_xception.fit(x_train, y_train, epochs=10, validation_data=(x_test, y_test))

# Evaluate the model
fine_tune_results_xception = model_xception.evaluate(x_test, y_test)
print(f"Xception Fine-tuning Test Accuracy: {fine_tune_results_xception[1]}")
\end{lstlisting}

In this fine-tuning approach, we unfreeze the last 30 layers of Xception and fine-tune the model using a smaller learning rate. This allows the pre-trained layers to gradually adapt to the CIFAR-10 dataset. After fine-tuning, the test accuracy is printed.

\section{MobileNet (2017)}

MobileNet, introduced by Google in 2017, is a family of lightweight convolutional neural networks designed specifically for mobile and embedded vision applications. MobileNet achieves this by utilizing depthwise separable convolutions, significantly reducing the number of parameters and computational cost compared to traditional CNN architectures like VGG or ResNet \cite{howard2017mobilenetsefficientconvolutionalneural}.

\textbf{Depthwise Separable Convolution}

The core innovation in MobileNet is the depthwise separable convolution. It breaks a standard convolution operation into two steps:
1. \textbf{Depthwise convolution}: Each input channel is convolved separately with a different filter, reducing the computational cost.
2. \textbf{Pointwise convolution}: A 1x1 convolution is applied to combine the outputs of the depthwise convolution, effectively mixing information from different channels.

This process significantly reduces both the number of parameters and the computational complexity while maintaining similar accuracy compared to standard convolution layers.

\textbf{MobileNet Architecture}

MobileNet uses multiple blocks of depthwise separable convolutions followed by batch normalization and ReLU activation. Each block reduces the spatial dimensions of the input through strided convolution or pooling. The architecture ends with a fully connected layer for classification, similar to other CNN architectures.

\textbf{Comparison of MobileNet Layers}

The following table provides a comparison of key components in the MobileNet architecture, detailing the number of depthwise and pointwise convolutions at different stages.

\begin{center}
\begin{tabular}{|c|c|c|c|c|}
\hline
\textbf{Component} & \textbf{Input Size} & \textbf{Depthwise Conv} & \textbf{Pointwise Conv} & \textbf{Stride} \\
\hline
Conv & 224x224x3 & -- & 32 & 2 \\
Depthwise Separable Conv & 112x112x32 & 32 & 64 & 1 \\
Depthwise Separable Conv & 112x112x64 & 64 & 128 & 2 \\
Depthwise Separable Conv & 56x56x128 & 128 & 128 & 1 \\
Depthwise Separable Conv & 56x56x128 & 128 & 256 & 2 \\
Depthwise Separable Conv & 28x28x256 & 256 & 256 & 1 \\
Depthwise Separable Conv & 28x28x256 & 256 & 512 & 2 \\
5x Depthwise Separable Convs & 14x14x512 & 512 & 512 & 1 \\
Depthwise Separable Conv & 14x14x512 & 512 & 1024 & 2 \\
Depthwise Separable Conv & 7x7x1024 & 1024 & 1024 & 1 \\
Average Pooling & 7x7x1024 & -- & -- & -- \\
FC & 1x1x1024 & -- & 1000 (Softmax) & -- \\
\hline
\end{tabular}
\end{center}

This table shows the main building blocks of MobileNet, from the initial convolution layer to the final fully connected layer for classification. The use of depthwise separable convolutions is consistent throughout the architecture.

\paragraph{Explanation of the Components}

\textbf{Conv Layer}: The initial standard convolution layer uses a 3x3 kernel and outputs 32 feature maps. It reduces the spatial resolution of the input image by half using a stride of 2.

\textbf{Depthwise Separable Convolution}: This is the key building block in MobileNet, consisting of two parts: a depthwise convolution (which applies a 3x3 filter to each input channel separately) and a pointwise convolution (which combines the outputs from depthwise convolution with a 1x1 filter). The depthwise separable convolution allows MobileNet to significantly reduce the number of parameters and computational cost.

\textbf{Stride}: Some layers use a stride of 2 to reduce the spatial resolution of the feature maps, while others use a stride of 1 to retain the same resolution.

\textbf{Average Pooling}: After the final depthwise separable convolution, global average pooling is applied to reduce the feature maps to a single value for each channel.

\textbf{Fully Connected Layer}: The classification head consists of a fully connected layer that maps the 1024 feature maps to 1000 categories, using softmax activation for classification.

\paragraph{Design Philosophy of MobileNet}

MobileNet is designed to be efficient in both computation and memory usage, making it suitable for deployment on mobile and embedded devices. The use of depthwise separable convolutions is a critical innovation that enables this efficiency without significantly compromising accuracy.

\textbf{TensorFlow Code for MobileNet V1}

Here is an implementation of MobileNet V1 using TensorFlow's \texttt{tf.keras} API:

\begin{lstlisting}[style=python]
import tensorflow as tf

def depthwise_separable_conv(x, filters, stride):
    x = tf.keras.layers.DepthwiseConv2D((3, 3), padding='same', strides=stride)(x)
    x = tf.keras.layers.BatchNormalization()(x)
    x = tf.keras.layers.ReLU()(x)
    x = tf.keras.layers.Conv2D(filters, (1, 1), padding='same')(x)
    x = tf.keras.layers.BatchNormalization()(x)
    x = tf.keras.layers.ReLU()(x)
    return x

def MobileNetV1(input_shape=(224, 224, 3), num_classes=1000):
    inputs = tf.keras.Input(shape=input_shape)
    
    # Initial Conv Layer
    x = tf.keras.layers.Conv2D(32, (3, 3), padding='same', strides=2)(inputs)
    x = tf.keras.layers.BatchNormalization()(x)
    x = tf.keras.layers.ReLU()(x)
    
    # Depthwise Separable Convolutions
    x = depthwise_separable_conv(x, 64, stride=1)
    x = depthwise_separable_conv(x, 128, stride=2)
    x = depthwise_separable_conv(x, 128, stride=1)
    x = depthwise_separable_conv(x, 256, stride=2)
    x = depthwise_separable_conv(x, 256, stride=1)
    x = depthwise_separable_conv(x, 512, stride=2)
    
    # 5x Repeated blocks
    for _ in range(5):
        x = depthwise_separable_conv(x, 512, stride=1)
    
    x = depthwise_separable_conv(x, 1024, stride=2)
    x = depthwise_separable_conv(x, 1024, stride=1)
    
    # Global Average Pooling
    x = tf.keras.layers.GlobalAveragePooling2D()(x)
    
    # Fully connected layer
    outputs = tf.keras.layers.Dense(num_classes, activation='softmax')(x)
    
    # Create model
    model = tf.keras.Model(inputs, outputs)
    return model

# Create the MobileNet V1 model
model = MobileNetV1()
model.summary()
\end{lstlisting}

\paragraph{Key Insights for Beginners}

\textbf{Why Depthwise Separable Convolution?} The depthwise separable convolution breaks the standard convolution into two parts: filtering and combining. This reduces the number of parameters and computations, making the model much more efficient.

\textbf{Why Global Average Pooling?} Global average pooling reduces each feature map to a single value by averaging all the elements in that feature map. This reduces the number of parameters in the fully connected layer and helps prevent overfitting.

\textbf{Why MobileNet for Mobile Devices?} MobileNet is specifically designed to run efficiently on mobile and embedded devices. Its architecture strikes a balance between computational efficiency and accuracy, making it suitable for applications with limited resources.

\subsection{MobileNetV1}
MobileNetV1 is a lightweight convolutional neural network architecture designed for efficient execution on mobile and embedded devices. Below, we will explore how to apply Linear Probe and Fine-tuning on MobileNetV1 using the CIFAR-10 dataset \cite{howard2017mobilenetsefficientconvolutionalneural}.

\subsubsection{MobileNet}
\paragraph{Linear Probe}
In the Linear Probe approach, we freeze the convolutional layers of the pre-trained MobileNetV1 model and train only the classification layers.

\begin{lstlisting}[style=python]
import tensorflow as tf
from tensorflow.keras import layers, models
from tensorflow.keras.applications import MobileNet
from tensorflow.keras.datasets import cifar10
from tensorflow.keras.utils import to_categorical

# Load and preprocess CIFAR-10 dataset
(x_train, y_train), (x_test, y_test) = cifar10.load_data()
x_train = tf.image.resize(x_train, (224, 224)) / 255.0
x_test = tf.image.resize(x_test, (224, 224)) / 255.0
y_train = to_categorical(y_train, 10)
y_test = to_categorical(y_test, 10)

# Load pre-trained MobileNet model without the top classification layer
base_model_mobilenet = MobileNet(weights='imagenet', include_top=False, input_shape=(224, 224, 3))

# Freeze all layers of the base model
for layer in base_model_mobilenet.layers:
    layer.trainable = False

# Add classification layers
model_mobilenet = models.Sequential([
    base_model_mobilenet,
    layers.GlobalAveragePooling2D(),
    layers.Dense(256, activation='relu'),
    layers.Dense(10, activation='softmax')
])

# Compile the model
model_mobilenet.compile(optimizer='adam', loss='categorical_crossentropy', metrics=['accuracy'])

# Train the model for linear probe
history_mobilenet = model_mobilenet.fit(x_train, y_train, epochs=10, validation_data=(x_test, y_test))

# Evaluate the model
results_mobilenet = model_mobilenet.evaluate(x_test, y_test)
print(f"MobileNet Linear Probe Test Accuracy: {results_mobilenet[1]}")
\end{lstlisting}

In this code, we resize the CIFAR-10 images to 224x224 pixels and use MobileNetV1 as the base model with frozen convolutional layers. We train only the classification layers, and the test accuracy is printed after training.

\paragraph{Fine-tuning}
In Fine-tuning, we unfreeze some layers of the MobileNetV1 model and train them along with the classification layers to better adapt the model to the CIFAR-10 dataset.

\begin{lstlisting}[style=python]
import tensorflow as tf
from tensorflow.keras import layers, models
from tensorflow.keras.applications import MobileNet
from tensorflow.keras.datasets import cifar10
from tensorflow.keras.utils import to_categorical

# Load and preprocess CIFAR-10 dataset
(x_train, y_train), (x_test, y_test) = cifar10.load_data()
x_train = tf.image.resize(x_train, (224, 224)) / 255.0
x_test = tf.image.resize(x_test, (224, 224)) / 255.0
y_train = to_categorical(y_train, 10)
y_test = to_categorical(y_test, 10)

# Load pre-trained MobileNet model without the top classification layer
base_model_mobilenet = MobileNet(weights='imagenet', include_top=False, input_shape=(224, 224, 3))

# Unfreeze the last few layers for fine-tuning
for layer in base_model_mobilenet.layers[-20:]:
    layer.trainable = True

# Add classification layers
model_mobilenet = models.Sequential([
    base_model_mobilenet,
    layers.GlobalAveragePooling2D(),
    layers.Dense(256, activation='relu'),
    layers.Dense(10, activation='softmax')
])

# Compile the model with a smaller learning rate
model_mobilenet.compile(optimizer=tf.keras.optimizers.Adam(1e-5), loss='categorical_crossentropy', metrics=['accuracy'])

# Fine-tune the model
fine_tune_history_mobilenet = model_mobilenet.fit(x_train, y_train, epochs=10, validation_data=(x_test, y_test))

# Evaluate the model
fine_tune_results_mobilenet = model_mobilenet.evaluate(x_test, y_test)
print(f"MobileNet Fine-tuning Test Accuracy: {fine_tune_results_mobilenet[1]}")
\end{lstlisting}

In this Fine-tuning approach, we unfreeze the last 20 layers of MobileNetV1 and train the model with a smaller learning rate to adapt the pre-trained features to the CIFAR-10 dataset.

\subsection{MobileNetV2 (2018)}
MobileNetV2 improves on MobileNetV1 by introducing inverted residual blocks and a linear bottleneck. Below, we apply both Linear Probe and Fine-tuning to MobileNetV2 using CIFAR-10 \cite{howard2017mobilenetsefficientconvolutionalneural}.

\subsubsection{MobileNetV2}
\paragraph{Linear Probe}
In the Linear Probe approach, we freeze all layers of the MobileNetV2 model and train only the classification layers.

\begin{lstlisting}[style=python]
import tensorflow as tf
from tensorflow.keras import layers, models
from tensorflow.keras.applications import MobileNetV2
from tensorflow.keras.datasets import cifar10
from tensorflow.keras.utils import to_categorical

# Load and preprocess CIFAR-10 dataset
(x_train, y_train), (x_test, y_test) = cifar10.load_data()
x_train = tf.image.resize(x_train, (224, 224)) / 255.0
x_test = tf.image.resize(x_test, (224, 224)) / 255.0
y_train = to_categorical(y_train, 10)
y_test = to_categorical(y_test, 10)

# Load pre-trained MobileNetV2 model without the top classification layer
base_model_mobilenetv2 = MobileNetV2(weights='imagenet', include_top=False, input_shape=(224, 224, 3))

# Freeze all layers of the base model
for layer in base_model_mobilenetv2.layers:
    layer.trainable = False

# Add classification layers
model_mobilenetv2 = models.Sequential([
    base_model_mobilenetv2,
    layers.GlobalAveragePooling2D(),
    layers.Dense(256, activation='relu'),
    layers.Dense(10, activation='softmax')
])

# Compile the model
model_mobilenetv2.compile(optimizer='adam', loss='categorical_crossentropy', metrics=['accuracy'])

# Train the model for linear probe
history_mobilenetv2 = model_mobilenetv2.fit(x_train, y_train, epochs=10, validation_data=(x_test, y_test))

# Evaluate the model
results_mobilenetv2 = model_mobilenetv2.evaluate(x_test, y_test)
print(f"MobileNetV2 Linear Probe Test Accuracy: {results_mobilenetv2[1]}")
\end{lstlisting}

In this code, we freeze all layers of the MobileNetV2 model and train only the newly added classification layers. The test accuracy is printed after training.

\paragraph{Fine-tuning}
In Fine-tuning, we unfreeze some layers of the MobileNetV2 model and train them along with the classification layers.

\begin{lstlisting}[style=python]
import tensorflow as tf
from tensorflow.keras import layers, models
from tensorflow.keras.applications import MobileNetV2
from tensorflow.keras.datasets import cifar10
from tensorflow.keras.utils import to_categorical

# Load and preprocess CIFAR-10 dataset
(x_train, y_train), (x_test, y_test) = cifar10.load_data()
x_train = tf.image.resize(x_train, (224, 224)) / 255.0
x_test = tf.image.resize(x_test, (224, 224)) / 255.0
y_train = to_categorical(y_train, 10)
y_test = to_categorical(y_test, 10)

# Load pre-trained MobileNetV2 model without the top classification layer
base_model_mobilenetv2 = MobileNetV2(weights='imagenet', include_top=False, input_shape=(224, 224, 3))

# Unfreeze the last few layers for fine-tuning
for layer in base_model_mobilenetv2.layers[-20:]:
    layer.trainable = True

# Add classification layers
model_mobilenetv2 = models.Sequential([
    base_model_mobilenetv2,
    layers.GlobalAveragePooling2D(),
    layers.Dense(256, activation='relu'),
    layers.Dense(10, activation='softmax')
])

# Compile the model with a smaller learning rate
model_mobilenetv2.compile(optimizer=tf.keras.optimizers.Adam(1e-5), loss='categorical_crossentropy', metrics=['accuracy'])

# Fine-tune the model
fine_tune_history_mobilenetv2 = model_mobilenetv2.fit(x_train, y_train, epochs=10, validation_data=(x_test, y_test))

# Evaluate the model
fine_tune_results_mobilenetv2 = model_mobilenetv2.evaluate(x_test, y_test)
print(f"MobileNetV2 Fine-tuning Test Accuracy: {fine_tune_results_mobilenetv2[1]}")
\end{lstlisting}

In this code, we fine-tune the MobileNetV2 model by unfreezing the last few layers and training them with a smaller learning rate to adapt to the CIFAR-10 dataset.

\subsection{MobileNetV3 (2019)}
MobileNetV3 improves on previous versions by combining efficient architecture search and optimizations for mobile devices. Below, we explore both MobileNetV3Large and MobileNetV3Small variants using Linear Probe and Fine-tuning approaches \cite{howard2017mobilenetsefficientconvolutionalneural}.

\subsubsection{MobileNetV3Large}
\paragraph{Linear Probe}
In the Linear Probe approach, we freeze all layers of MobileNetV3Large and train only the classification layers.

\begin{lstlisting}[style=python]
import tensorflow as tf
from tensorflow.keras import layers, models
from tensorflow.keras.applications import MobileNetV3Large
from tensorflow.keras.datasets import cifar10
from tensorflow.keras.utils import to_categorical

# Load and preprocess CIFAR-10 dataset
(x_train, y_train), (x_test, y_test) = cifar10.load_data()
x_train = tf.image.resize(x_train, (224, 224)) / 255.0
x_test = tf.image.resize(x_test, (224, 224)) / 255.0
y_train = to_categorical(y_train, 10)
y_test = to_categorical(y_test, 10)

# Load pre-trained MobileNetV3Large model without the top classification layer
base_model_mobilenetv3large = MobileNetV3Large(weights='imagenet', include_top=False, input_shape=(224, 224, 3))

# Freeze all layers of the base model
for layer in base_model_mobilenetv3large.layers:
    layer.trainable = False

# Add classification layers
model_mobilenetv3large = models.Sequential([
    base_model_mobilenetv3large,
    layers.GlobalAveragePooling2D(),
    layers.Dense(256, activation='relu'),
    layers.Dense(10, activation='softmax')
])

# Compile the model
model_mobilenetv3large.compile(optimizer='adam', loss='categorical_crossentropy', metrics=['accuracy'])

# Train the model for linear probe
history_mobilenetv3large = model_mobilenetv3large.fit(x_train, y_train, epochs=10, validation_data=(x_test, y_test))

# Evaluate the model
results_mobilenetv3large = model_mobilenetv3large.evaluate(x_test, y_test)
print(f"MobileNetV3Large Linear Probe Test Accuracy: {results_mobilenetv3large[1]}")
\end{lstlisting}

In this Linear Probe approach, we freeze all convolutional layers of MobileNetV3Large and train only the classification layers. The test accuracy is printed after training.

\paragraph{Fine-tuning}
For Fine-tuning, we unfreeze some layers of MobileNetV3Large and train them along with the classification layers.

\begin{lstlisting}[style=python]
import tensorflow as tf
from tensorflow.keras import layers, models
from tensorflow.keras.applications import MobileNetV3Large
from tensorflow.keras.datasets import cifar10
from tensorflow.keras.utils import to_categorical

# Load and preprocess CIFAR-10 dataset
(x_train, y_train), (x_test, y_test) = cifar10.load_data()
x_train = tf.image.resize(x_train, (224, 224)) / 255.0
x_test = tf.image.resize(x_test, (224, 224)) / 255.0
y_train = to_categorical(y_train, 10)
y_test = to_categorical(y_test, 10)

# Load pre-trained MobileNetV3Large model without the top classification layer
base_model_mobilenetv3large = MobileNetV3Large(weights='imagenet', include_top=False, input_shape=(224, 224, 3))

# Unfreeze the last few layers for fine-tuning
for layer in base_model_mobilenetv3large.layers[-20:]:
    layer.trainable = True

# Add classification layers
model_mobilenetv3large = models.Sequential([
    base_model_mobilenetv3large,
    layers.GlobalAveragePooling2D(),
    layers.Dense(256, activation='relu'),
    layers.Dense(10, activation='softmax')
])

# Compile the model with a smaller learning rate
model_mobilenetv3large.compile(optimizer=tf.keras.optimizers.Adam(1e-5), loss='categorical_crossentropy', metrics=['accuracy'])

# Fine-tune the model
fine_tune_history_mobilenetv3large = model_mobilenetv3large.fit(x_train, y_train, epochs=10, validation_data=(x_test, y_test))

# Evaluate the model
fine_tune_results_mobilenetv3large = model_mobilenetv3large.evaluate(x_test, y_test)
print(f"MobileNetV3Large Fine-tuning Test Accuracy: {fine_tune_results_mobilenetv3large[1]}")
\end{lstlisting}

In this Fine-tuning approach, we unfreeze the last 20 layers of MobileNetV3Large and train them with a smaller learning rate to adapt the pre-trained weights to CIFAR-10.

\subsubsection{MobileNetV3Small}
\paragraph{Linear Probe}
In the Linear Probe approach, we freeze all convolutional layers of MobileNetV3Small and train only the classification layers.

\begin{lstlisting}[style=python]
import tensorflow as tf
from tensorflow.keras import layers, models
from tensorflow.keras.applications import MobileNetV3Small
from tensorflow.keras.datasets import cifar10
from tensorflow.keras.utils import to_categorical

# Load and preprocess CIFAR-10 dataset
(x_train, y_train), (x_test, y_test) = cifar10.load_data()
x_train = tf.image.resize(x_train, (224, 224)) / 255.0
x_test = tf.image.resize(x_test, (224, 224)) / 255.0
y_train = to_categorical(y_train, 10)
y_test = to_categorical(y_test, 10)

# Load pre-trained MobileNetV3Small model without the top classification layer
base_model_mobilenetv3small = MobileNetV3Small(weights='imagenet', include_top=False, input_shape=(224, 224, 3))

# Freeze all layers of the base model
for layer in base_model_mobilenetv3small.layers:
    layer.trainable = False

# Add classification layers
model_mobilenetv3small = models.Sequential([
    base_model_mobilenetv3small,
    layers.GlobalAveragePooling2D(),
    layers.Dense(256, activation='relu'),
    layers.Dense(10, activation='softmax')
])

# Compile the model
model_mobilenetv3small.compile(optimizer='adam', loss='categorical_crossentropy', metrics=['accuracy'])

# Train the model for linear probe
history_mobilenetv3small = model_mobilenetv3small.fit(x_train, y_train, epochs=10, validation_data=(x_test, y_test))

# Evaluate the model
results_mobilenetv3small = model_mobilenetv3small.evaluate(x_test, y_test)
print(f"MobileNetV3Small Linear Probe Test Accuracy: {results_mobilenetv3small[1]}")
\end{lstlisting}

In this code, we freeze all layers of MobileNetV3Small and train only the classification layers, with the test accuracy being printed after training.

\paragraph{Fine-tuning}
In Fine-tuning, we unfreeze some layers of MobileNetV3Small and train them with the classification layers.

\begin{lstlisting}[style=python]
import tensorflow as tf
from tensorflow.keras import layers, models
from tensorflow.keras.applications import MobileNetV3Small
from tensorflow.keras.datasets import cifar10
from tensorflow.keras.utils import to_categorical

# Load and preprocess CIFAR-10 dataset
(x_train, y_train), (x_test, y_test) = cifar10.load_data()
x_train = tf.image.resize(x_train, (224, 224)) / 255.0
x_test = tf.image.resize(x_test, (224, 224)) / 255.0
y_train = to_categorical(y_train, 10)
y_test = to_categorical(y_test, 10)

# Load pre-trained MobileNetV3Small model without the top classification layer
base_model_mobilenetv3small = MobileNetV3Small(weights='imagenet', include_top=False, input_shape=(224, 224, 3))

# Unfreeze the last few layers for fine-tuning
for layer in base_model_mobilenetv3small.layers[-20:]:
    layer.trainable = True

# Add classification layers
model_mobilenetv3small = models.Sequential([
    base_model_mobilenetv3small,
    layers.GlobalAveragePooling2D(),
    layers.Dense(256, activation='relu'),
    layers.Dense(10, activation='softmax')
])

# Compile the model with a smaller learning rate
model_mobilenetv3small.compile(optimizer=tf.keras.optimizers.Adam(1e-5), loss='categorical_crossentropy', metrics=['accuracy'])

# Fine-tune the model
fine_tune_history_mobilenetv3small = model_mobilenetv3small.fit(x_train, y_train, epochs=10, validation_data=(x_test, y_test))

# Evaluate the model
fine_tune_results_mobilenetv3small = model_mobilenetv3small.evaluate(x_test, y_test)
print(f"MobileNetV3Small Fine-tuning Test Accuracy: {fine_tune_results_mobilenetv3small[1]}")
\end{lstlisting}

In this Fine-tuning approach, we unfreeze the last 20 layers of MobileNetV3Small and fine-tune the model using a smaller learning rate. After fine-tuning, the test accuracy is printed.

\section{NASNet (2018)}

NASNet (Neural Architecture Search Network) was introduced by Google in 2018. It is a convolutional neural network architecture designed using a neural architecture search (NAS) approach. The idea behind NASNet is to use reinforcement learning to automatically design an optimal architecture for image classification, surpassing manually designed networks such as VGG and ResNet. NASNet is able to adapt its architecture to different computational constraints and tasks, making it flexible and powerful \cite{DBLP:journals/corr/ZophVSL17}.

However, NASNet is considered an older architecture, and its complexity is often seen as a drawback. Its intricate structure makes it unnecessarily complicated to manually implement. Moreover, the method used to design NASNet, known as neural architecture search (NAS), involves a computationally expensive process that requires significant resources. Running NAS on ordinary devices is virtually impossible due to the need for extensive computing power, making it impractical to manually design architectures like NASNet on such hardware. For practical purposes, it's more efficient to use pre-built models from libraries rather than trying to recreate them from scratch.

\textbf{Key Features of NASNet}

NASNet is built from two fundamental units: the \textbf{normal cell} and the \textbf{reduction cell}. These cells are discovered via neural architecture search and repeated across the network. A normal cell preserves the spatial dimensions of the input, while a reduction cell reduces the spatial dimensions, usually by a factor of two.

\textbf{Comparison of NASNet Architectures}

NASNet can be scaled to different sizes depending on the task and computational resources. Below is a comparison of the main components in NASNet-A, one of the most widely used variants of NASNet. The table compares the number of normal cells and reduction cells in the architecture at different scales.

\begin{center}
\begin{tabular}{|c|c|c|c|}
\hline
\textbf{Component} & \textbf{NASNet-Mobile} & \textbf{NASNet-Large} \\
\hline
% 这里就是这样的 cell在一起，所以不要加hline
Normal Cells & 12 & 18 \\
Reduction Cells & 2 & 2 \\
Stem Cell & \multicolumn{2}{c|}{1 (Conv3-32)} \\
\hline
FC-1000 (Softmax) & \multicolumn{2}{c|}{1} \\
\hline
\end{tabular}
\end{center}

\paragraph{Explanation of the Components}

\textbf{Normal Cells}: A normal cell is a modular block in NASNet that preserves the spatial dimensions of the input data. It consists of convolutional layers, batch normalization, and ReLU activations. The architecture of a normal cell is discovered through neural architecture search.

\textbf{Reduction Cells}: A reduction cell reduces the spatial dimensions of the input by a factor of two, typically by using strided convolutions or pooling operations. Like the normal cell, the architecture of a reduction cell is also discovered via neural architecture search.

\textbf{Stem Cell}: The stem cell is the initial block of the NASNet architecture, responsible for processing the input image. It typically consists of a few convolutional layers to adjust the input dimensions before passing the data to the normal and reduction cells.

\textbf{Classification Head}: The classification head in NASNet is similar to other deep CNN architectures. After the final normal or reduction cell, the feature maps are passed through a global average pooling layer followed by a fully connected layer with 1000 units (for ImageNet classification) and a softmax activation function to output class probabilities.

\paragraph{Design Philosophy of NASNet}

NASNet is designed with the idea that human-designed architectures are often suboptimal for specific tasks or constraints. The neural architecture search approach uses reinforcement learning to automatically discover an optimal architecture, which can then be scaled for different computational constraints. NASNet achieves state-of-the-art results on tasks like image classification while being more computationally efficient than traditional architectures. The process of NAS, however, requires significant computational resources, which makes it unsuitable for typical hardware setups. Hence, it is recommended to leverage pre-built models from deep learning libraries like TensorFlow or PyTorch instead of attempting to manually implement NASNet.

\textbf{TensorFlow Code for NASNet-Mobile}

The following code demonstrates how to load a pre-trained NASNet-Mobile model from TensorFlow's \texttt{tf.keras.applications} module. NASNet-Mobile is a smaller variant of NASNet designed for mobile and embedded devices.

\begin{lstlisting}[style=python]
import tensorflow as tf

# Load NASNet-Mobile pre-trained model
model = tf.keras.applications.NASNetMobile(weights='imagenet', input_shape=(224, 224, 3))

# View model summary
model.summary()
\end{lstlisting}

\paragraph{Key Insights for Beginners}

\textbf{Why use NASNet?} NASNet is a product of neural architecture search, meaning it is automatically designed to be optimal for the task at hand. This allows it to achieve better performance and efficiency than manually designed networks.

\textbf{Normal vs. Reduction Cells}: The distinction between normal and reduction cells is important. Normal cells keep the spatial dimensions of the input intact, while reduction cells halve the spatial dimensions, helping to downsample the input and increase the receptive field of the network.

\textbf{Transferability}: NASNet can be scaled up or down depending on the computational resources available. This makes it versatile for both large-scale server deployments (NASNet-Large) and resource-constrained devices (NASNet-Mobile). Since the architecture is highly complex and the NAS method cannot realistically be implemented on most devices, it is recommended to use pre-built models from established libraries like TensorFlow instead of manually creating the architecture.

\subsection{NASNetLarge}
NASNetLarge is a model discovered using Neural Architecture Search (NAS), which finds the best architecture for a given task. It is a larger version of NASNet optimized for higher accuracy. Below, we apply both Linear Probe and Fine-tuning approaches to NASNetLarge using the CIFAR-10 dataset \cite{DBLP:journals/corr/ZophVSL17}.

\paragraph{Linear Probe}
In the Linear Probe approach, we freeze the convolutional layers of the pre-trained NASNetLarge model and train only the classification layers.

\begin{lstlisting}[style=python]
import tensorflow as tf
from tensorflow.keras import layers, models
from tensorflow.keras.applications import NASNetLarge
from tensorflow.keras.datasets import cifar10
from tensorflow.keras.utils import to_categorical

# Load and preprocess CIFAR-10 dataset
(x_train, y_train), (x_test, y_test) = cifar10.load_data()
x_train = tf.image.resize(x_train, (331, 331)) / 255.0
x_test = tf.image.resize(x_test, (331, 331)) / 255.0
y_train = to_categorical(y_train, 10)
y_test = to_categorical(y_test, 10)

# Load pre-trained NASNetLarge model without the top classification layer
base_model_nasnetlarge = NASNetLarge(weights='imagenet', include_top=False, input_shape=(331, 331, 3))

# Freeze all layers of the base model
for layer in base_model_nasnetlarge.layers:
    layer.trainable = False

# Add classification layers
model_nasnetlarge = models.Sequential([
    base_model_nasnetlarge,
    layers.GlobalAveragePooling2D(),
    layers.Dense(256, activation='relu'),
    layers.Dense(10, activation='softmax')
])

# Compile the model
model_nasnetlarge.compile(optimizer='adam', loss='categorical_crossentropy', metrics=['accuracy'])

# Train the model for linear probe
history_nasnetlarge = model_nasnetlarge.fit(x_train, y_train, epochs=10, validation_data=(x_test, y_test))

# Evaluate the model
results_nasnetlarge = model_nasnetlarge.evaluate(x_test, y_test)
print(f"NASNetLarge Linear Probe Test Accuracy: {results_nasnetlarge[1]}")
\end{lstlisting}

In this code, we resize the CIFAR-10 images to 331x331 pixels to match NASNetLarge's input size. We freeze all the layers of the pre-trained NASNetLarge model and train only the newly added classification layers. The test accuracy is printed after training.

\paragraph{Fine-tuning}
In Fine-tuning, we unfreeze some layers of the NASNetLarge model and train them along with the classification layers to adapt the model to the CIFAR-10 dataset.

\begin{lstlisting}[style=python]
import tensorflow as tf
from tensorflow.keras import layers, models
from tensorflow.keras.applications import NASNetLarge
from tensorflow.keras.datasets import cifar10
from tensorflow.keras.utils import to_categorical

# Load and preprocess CIFAR-10 dataset
(x_train, y_train), (x_test, y_test) = cifar10.load_data()
x_train = tf.image.resize(x_train, (331, 331)) / 255.0
x_test = tf.image.resize(x_test, (331, 331)) / 255.0
y_train = to_categorical(y_train, 10)
y_test = to_categorical(y_test, 10)

# Load pre-trained NASNetLarge model without the top classification layer
base_model_nasnetlarge = NASNetLarge(weights='imagenet', include_top=False, input_shape=(331, 331, 3))

# Unfreeze the last few layers for fine-tuning
for layer in base_model_nasnetlarge.layers[-20:]:
    layer.trainable = True

# Add classification layers
model_nasnetlarge = models.Sequential([
    base_model_nasnetlarge,
    layers.GlobalAveragePooling2D(),
    layers.Dense(256, activation='relu'),
    layers.Dense(10, activation='softmax')
])

# Compile the model with a smaller learning rate
model_nasnetlarge.compile(optimizer=tf.keras.optimizers.Adam(1e-5), loss='categorical_crossentropy', metrics=['accuracy'])

# Fine-tune the model
fine_tune_history_nasnetlarge = model_nasnetlarge.fit(x_train, y_train, epochs=10, validation_data=(x_test, y_test))

# Evaluate the model
fine_tune_results_nasnetlarge = model_nasnetlarge.evaluate(x_test, y_test)
print(f"NASNetLarge Fine-tuning Test Accuracy: {fine_tune_results_nasnetlarge[1]}")
\end{lstlisting}

In this Fine-tuning approach, we unfreeze the last 20 layers of NASNetLarge and train the entire model using a smaller learning rate to gradually adapt the pre-trained weights to CIFAR-10.

\subsection{NASNetMobile}
NASNetMobile is a smaller version of NASNet optimized for mobile and resource-constrained devices. Below, we apply both Linear Probe and Fine-tuning approaches to NASNetMobile using CIFAR-10 \cite{DBLP:journals/corr/ZophVSL17}.

\paragraph{Linear Probe}
In the Linear Probe approach, we freeze the convolutional layers of the pre-trained NASNetMobile model and train only the classification layers.

\begin{lstlisting}[style=python]
import tensorflow as tf
from tensorflow.keras import layers, models
from tensorflow.keras.applications import NASNetMobile
from tensorflow.keras.datasets import cifar10
from tensorflow.keras.utils import to_categorical

# Load and preprocess CIFAR-10 dataset
(x_train, y_train), (x_test, y_test) = cifar10.load_data()
x_train = tf.image.resize(x_train, (224, 224)) / 255.0
x_test = tf.image.resize(x_test, (224, 224)) / 255.0
y_train = to_categorical(y_train, 10)
y_test = to_categorical(y_test, 10)

# Load pre-trained NASNetMobile model without the top classification layer
base_model_nasnetmobile = NASNetMobile(weights='imagenet', include_top=False, input_shape=(224, 224, 3))

# Freeze all layers of the base model
for layer in base_model_nasnetmobile.layers:
    layer.trainable = False

# Add classification layers
model_nasnetmobile = models.Sequential([
    base_model_nasnetmobile,
    layers.GlobalAveragePooling2D(),
    layers.Dense(256, activation='relu'),
    layers.Dense(10, activation='softmax')
])

# Compile the model
model_nasnetmobile.compile(optimizer='adam', loss='categorical_crossentropy', metrics=['accuracy'])

# Train the model for linear probe
history_nasnetmobile = model_nasnetmobile.fit(x_train, y_train, epochs=10, validation_data=(x_test, y_test))

# Evaluate the model
results_nasnetmobile = model_nasnetmobile.evaluate(x_test, y_test)
print(f"NASNetMobile Linear Probe Test Accuracy: {results_nasnetmobile[1]}")
\end{lstlisting}

In this code, we resize the CIFAR-10 images to 224x224 pixels to match NASNetMobile's input size. We freeze all the layers of the pre-trained NASNetMobile model and train only the newly added classification layers. The test accuracy is printed after training.

\paragraph{Fine-tuning}
In Fine-tuning, we unfreeze some layers of the NASNetMobile model and train them along with the classification layers to adapt the model to the CIFAR-10 dataset.

\begin{lstlisting}[style=python]
import tensorflow as tf
from tensorflow.keras import layers, models
from tensorflow.keras.applications import NASNetMobile
from tensorflow.keras.datasets import cifar10
from tensorflow.keras.utils import to_categorical

# Load and preprocess CIFAR-10 dataset
(x_train, y_train), (x_test, y_test) = cifar10.load_data()
x_train = tf.image.resize(x_train, (224, 224)) / 255.0
x_test = tf.image.resize(x_test, (224, 224)) / 255.0
y_train = to_categorical(y_train, 10)
y_test = to_categorical(y_test, 10)

# Load pre-trained NASNetMobile model without the top classification layer
base_model_nasnetmobile = NASNetMobile(weights='imagenet', include_top=False, input_shape=(224, 224, 3))

# Unfreeze the last few layers for fine-tuning
for layer in base_model_nasnetmobile.layers[-20:]:
    layer.trainable = True

# Add classification layers
model_nasnetmobile = models.Sequential([
    base_model_nasnetmobile,
    layers.GlobalAveragePooling2D(),
    layers.Dense(256, activation='relu'),
    layers.Dense(10, activation='softmax')
])

# Compile the model with a smaller learning rate
model_nasnetmobile.compile(optimizer=tf.keras.optimizers.Adam(1e-5), loss='categorical_crossentropy', metrics=['accuracy'])

# Fine-tune the model
fine_tune_history_nasnetmobile = model_nasnetmobile.fit(x_train, y_train, epochs=10, validation_data=(x_test, y_test))

# Evaluate the model
fine_tune_results_nasnetmobile = model_nasnetmobile.evaluate(x_test, y_test)
print(f"NASNetMobile Fine-tuning Test Accuracy: {fine_tune_results_nasnetmobile[1]}")
\end{lstlisting}

In this Fine-tuning approach, we unfreeze the last 20 layers of NASNetMobile and train the entire model using a smaller learning rate to gradually adapt the pre-trained weights to CIFAR-10.

\section{EfficientNet (2019)}

EfficientNet, introduced by Google in 2019, is a family of convolutional neural networks (CNNs) designed to achieve high accuracy with fewer parameters and computations. The key innovation of EfficientNet is the compound scaling method, which uniformly scales the depth, width, and resolution of the network to create more efficient models \cite{DBLP:journals/corr/abs-1905-11946}.

EfficientNet models, from B0 to B7, progressively increase in complexity and performance by scaling up all three dimensions. The base model, EfficientNet-B0, is built on a mobile architecture called MobileNetV2, which uses depthwise separable convolutions to reduce computation costs.

\textbf{Comparison of EfficientNet Architectures}

EfficientNet models vary primarily in their depth, width, and resolution. These parameters are scaled using a compound coefficient $\phi$, which adjusts the model complexity based on available computational resources. Below is a comparison of the key components of the EfficientNet family, from B0 to B7 \cite{DBLP:journals/corr/abs-1905-11946}:

\begin{center}
\begin{tabular}{|c|c|c|c|c|}
\hline
\textbf{Component} & \textbf{EfficientNet-B0} & \textbf{EfficientNet-B1} & \textbf{EfficientNet-B4} & \textbf{EfficientNet-B7} \\
\hline
Depth & 16 & 18 & 24 & 30 \\
Width & 1.0x & 1.1x & 1.4x & 2.0x \\
Resolution & 224x224 & 240x240 & 380x380 & 600x600 \\
\hline
\end{tabular}
\end{center}

\paragraph{Explanation of the Components}

\textbf{Depth}: The depth of the network refers to the number of layers. As the model scales up from B0 to B7, the number of layers increases, allowing the model to capture more complex patterns in the data.

\textbf{Width}: The width of the network refers to the number of channels in each layer. Scaling the width increases the capacity of each layer to learn more features, which is especially useful when dealing with large datasets.

\textbf{Resolution}: The resolution refers to the input image size. EfficientNet uses larger input resolutions for higher variants (e.g., B4, B7) to capture more detail from images, which improves accuracy at the cost of higher computational complexity.

\textbf{Compound Scaling}: The key idea behind EfficientNet is to scale depth, width, and resolution uniformly. The compound scaling method ensures that the model grows in a balanced way, avoiding overfitting or underfitting while maintaining computational efficiency.

\textbf{Detailed Composition of EfficientNet Architectures}

The table below provides a detailed breakdown of the EfficientNet architectures, from B0 to B7, following the structure and format from the EfficientNet paper. Each row represents a block in the architecture, with information on the input size, operation type, number of filters, output size, expansion ratio, and repeat count.

\begin{center}
\begin{tabular}{|c|c|c|c|c|c|c|}
\hline
\textbf{Stage} & \textbf{Operator} & \textbf{Input Size} & \textbf{Expand Ratio} & \textbf{Output Channels} & \textbf{Repeats} & \textbf{Stride} \\
\hline
1 & Conv3x3 & 224x224 & - & 32 & 1 & 2 \\
\hline
2 & MBConv1, k3x3 & 112x112 & 1 & 16 & 1 & 1 \\
\hline
3 & MBConv6, k3x3 & 112x112 & 6 & 24 & 2 & 2 \\
\hline
4 & MBConv6, k5x5 & 56x56 & 6 & 40 & 2 & 2 \\
\hline
5 & MBConv6, k3x3 & 28x28 & 6 & 80 & 3 & 2 \\
\hline
6 & MBConv6, k5x5 & 14x14 & 6 & 112 & 3 & 1 \\
\hline
7 & MBConv6, k5x5 & 14x14 & 6 & 192 & 4 & 2 \\
\hline
8 & MBConv6, k3x3 & 7x7 & 6 & 320 & 1 & 1 \\
\hline
9 & Conv1x1 & 7x7 & - & 1280 & 1 & - \\
\hline
10 & FC & 1x1 & - & 1000 & 1 & - \\
\hline
\end{tabular}
\end{center}

This table lists the detailed structure of EfficientNet-B0, showing the operator used at each stage, the input size, expansion ratio, output channels, number of repeats, and stride. This pattern of scaling the number of layers, channels, and resolution is consistent across all EfficientNet models, but with each larger model (e.g., B1, B4, B7), the numbers increase.

\paragraph{Design Philosophy of EfficientNet}

EfficientNet introduces a new scaling strategy called compound scaling, which scales the depth, width, and resolution of the network simultaneously rather than scaling them individually. This balanced scaling enables EfficientNet models to achieve better performance with fewer parameters compared to traditional models like ResNet or VGG, which scale depth alone.

In contrast to other architectures that simply increase depth to improve performance, EfficientNet optimizes for both accuracy and efficiency by carefully balancing the number of layers, the number of channels per layer, and the image resolution. This makes EfficientNet highly suitable for tasks where computational resources are limited but high accuracy is still required.

\textbf{TensorFlow Code for EfficientNet-B0}

Below is an implementation of the EfficientNet-B0 model using TensorFlow and the \texttt{tf.keras} API:

\begin{lstlisting}[style=python]
import tensorflow as tf
from tensorflow.keras.applications import EfficientNetB0

# Load the pre-trained EfficientNetB0 model with ImageNet weights
model = EfficientNetB0(weights='imagenet')

# Print model summary
model.summary()
\end{lstlisting}

This code uses the pre-trained EfficientNet-B0 model provided by TensorFlow, which has been trained on the ImageNet dataset. EfficientNet-B0 is the base model of the EfficientNet family and is highly efficient in terms of both accuracy and computational cost.

\paragraph{Key Insights for Beginners}

\textbf{Why Compound Scaling?} EfficientNet uses compound scaling to balance the depth, width, and resolution of the network. This ensures that the network grows in a balanced way, improving both accuracy and efficiency.

\textbf{Why Use Depthwise Separable Convolutions?} Depthwise separable convolutions reduce the number of parameters and computational cost without sacrificing much accuracy. This is why they are used in the EfficientNet architecture, especially for mobile and low-resource environments.

\textbf{Fully Connected Layers in EfficientNet}: Like other CNNs, EfficientNet concludes with fully connected layers to classify the input images into categories. These layers are essential for interpreting the high-level features learned by the preceding convolutional layers.

\subsection{EfficientNetV1}
EfficientNetV1 is a family of convolutional neural networks that scale efficiently in terms of accuracy and computational resources. In this section, we will apply transfer learning using EfficientNetB0 and EfficientNetB1 on the CIFAR-10 dataset using both Linear Probe and Fine-tuning approaches \cite{DBLP:journals/corr/abs-1905-11946}.

\subsubsection{EfficientNetB0}
EfficientNetB0 is the smallest model in the EfficientNetV1 family. The input image size for this model is 224x224, so we will resize the CIFAR-10 images accordingly.

\paragraph{Linear Probe}
In the Linear Probe approach, we freeze the EfficientNetB0 model's pre-trained layers and train only the classification layers on CIFAR-10.

\begin{lstlisting}[style=python]
import tensorflow as tf
from tensorflow.keras import layers, models
from tensorflow.keras.applications import EfficientNetB0
from tensorflow.keras.datasets import cifar10
from tensorflow.keras.utils import to_categorical

# Load and preprocess CIFAR-10 dataset
(x_train, y_train), (x_test, y_test) = cifar10.load_data()
x_train = tf.image.resize(x_train, (224, 224)) / 255.0
x_test = tf.image.resize(x_test, (224, 224)) / 255.0
y_train = to_categorical(y_train, 10)
y_test = to_categorical(y_test, 10)

# Load pre-trained EfficientNetB0 model without the top classification layer
base_model = EfficientNetB0(weights='imagenet', include_top=False, input_shape=(224, 224, 3))

# Freeze all layers of the base model
for layer in base_model.layers:
    layer.trainable = False

# Add classification layers on top of the base model
model = models.Sequential([
    base_model,
    layers.GlobalAveragePooling2D(),
    layers.Dense(256, activation='relu'),
    layers.Dense(10, activation='softmax')
])

# Compile the model
model.compile(optimizer='adam', loss='categorical_crossentropy', metrics=['accuracy'])

# Train the model for linear probe
history = model.fit(x_train, y_train, epochs=10, validation_data=(x_test, y_test))

# Evaluate the model
results = model.evaluate(x_test, y_test)
print(f"EfficientNetB0 Linear Probe Test Accuracy: {results[1]}")
\end{lstlisting}

In this code, we use the EfficientNetB0 pre-trained on ImageNet, freeze all its layers, and add a custom classification head. The CIFAR-10 images are resized to 224x224 pixels. The test accuracy is evaluated after training.

\paragraph{Fine-tuning}
In Fine-tuning, we unfreeze some layers of the EfficientNetB0 model and train both the classification head and the unfrozen layers.

\begin{lstlisting}[style=python]
import tensorflow as tf
from tensorflow.keras import layers, models
from tensorflow.keras.applications import EfficientNetB0
from tensorflow.keras.datasets import cifar10
from tensorflow.keras.utils import to_categorical

# Load and preprocess CIFAR-10 dataset
(x_train, y_train), (x_test, y_test) = cifar10.load_data()
x_train = tf.image.resize(x_train, (224, 224)) / 255.0
x_test = tf.image.resize(x_test, (224, 224)) / 255.0
y_train = to_categorical(y_train, 10)
y_test = to_categorical(y_test, 10)

# Load pre-trained EfficientNetB0 model without the top classification layer
base_model = EfficientNetB0(weights='imagenet', include_top=False, input_shape=(224, 224, 3))

# Unfreeze the last 4 layers of the base model for fine-tuning
for layer in base_model.layers[-4:]:
    layer.trainable = True

# Add classification layers on top of the base model
model = models.Sequential([
    base_model,
    layers.GlobalAveragePooling2D(),
    layers.Dense(256, activation='relu'),
    layers.Dense(10, activation='softmax')
])

# Compile the model with a smaller learning rate for fine-tuning
model.compile(optimizer=tf.keras.optimizers.Adam(1e-5), loss='categorical_crossentropy', metrics=['accuracy'])

# Fine-tune the model
fine_tune_history = model.fit(x_train, y_train, epochs=10, validation_data=(x_test, y_test))

# Evaluate the model
fine_tune_results = model.evaluate(x_test, y_test)
print(f"EfficientNetB0 Fine-tuning Test Accuracy: {fine_tune_results[1]}")
\end{lstlisting}

In this Fine-tuning approach, we unfreeze the last four layers of EfficientNetB0, allowing them to be updated during training. The model is then trained with a lower learning rate to fine-tune the pre-trained layers. After training, the model's test accuracy is evaluated.

\subsubsection{EfficientNetB1}
EfficientNetB1 is a slightly larger model than EfficientNetB0. The input image size for this model is 240x240, so the CIFAR-10 images will be resized accordingly.

\paragraph{Linear Probe}
In the Linear Probe approach, the EfficientNetB1 model's pre-trained layers are frozen, and we train only the classification layers.

\begin{lstlisting}[style=python]
import tensorflow as tf
from tensorflow.keras import layers, models
from tensorflow.keras.applications import EfficientNetB1
from tensorflow.keras.datasets import cifar10
from tensorflow.keras.utils import to_categorical

# Load and preprocess CIFAR-10 dataset
(x_train, y_train), (x_test, y_test) = cifar10.load_data()
x_train = tf.image.resize(x_train, (240, 240)) / 255.0
x_test = tf.image.resize(x_test, (240, 240)) / 255.0
y_train = to_categorical(y_train, 10)
y_test = to_categorical(y_test, 10)

# Load pre-trained EfficientNetB1 model without the top classification layer
base_model_b1 = EfficientNetB1(weights='imagenet', include_top=False, input_shape=(240, 240, 3))

# Freeze all layers of the base model
for layer in base_model_b1.layers:
    layer.trainable = False

# Add classification layers on top of the base model
model_b1 = models.Sequential([
    base_model_b1,
    layers.GlobalAveragePooling2D(),
    layers.Dense(256, activation='relu'),
    layers.Dense(10, activation='softmax')
])

# Compile the model
model_b1.compile(optimizer='adam', loss='categorical_crossentropy', metrics=['accuracy'])

# Train the model for linear probe
b1_history = model_b1.fit(x_train, y_train, epochs=10, validation_data=(x_test, y_test))

# Evaluate the model
b1_results = model_b1.evaluate(x_test, y_test)
print(f"EfficientNetB1 Linear Probe Test Accuracy: {b1_results[1]}")
\end{lstlisting}

In this Linear Probe approach, we load EfficientNetB1 pre-trained on ImageNet, freeze its convolutional layers, and add a custom classification head. The CIFAR-10 images are resized to 240x240 pixels. The model is trained, and test accuracy is printed.

\paragraph{Fine-tuning}
In Fine-tuning, we unfreeze some layers of the EfficientNetB1 model and train both the classification head and the unfrozen layers.

\begin{lstlisting}[style=python]
import tensorflow as tf
from tensorflow.keras import layers, models
from tensorflow.keras.applications import EfficientNetB1
from tensorflow.keras.datasets import cifar10
from tensorflow.keras.utils import to_categorical

# Load and preprocess CIFAR-10 dataset
(x_train, y_train), (x_test, y_test) = cifar10.load_data()
x_train = tf.image.resize(x_train, (240, 240)) / 255.0
x_test = tf.image.resize(x_test, (240, 240)) / 255.0
y_train = to_categorical(y_train, 10)
y_test = to_categorical(y_test, 10)

# Load pre-trained EfficientNetB1 model without the top classification layer
base_model_b1 = EfficientNetB1(weights='imagenet', include_top=False, input_shape=(240, 240, 3))

# Unfreeze the last 4 layers of the base model for fine-tuning
for layer in base_model_b1.layers[-4:]:
    layer.trainable = True

# Add classification layers on top of the base model
model_b1 = models.Sequential([
    base_model_b1,
    layers.GlobalAveragePooling2D(),
    layers.Dense(256, activation='relu'),
    layers.Dense(10, activation='softmax')
])

# Compile the model with a smaller learning rate for fine-tuning
model_b1.compile(optimizer=tf.keras.optimizers.Adam(1e-5), loss='categorical_crossentropy', metrics=['accuracy'])

# Fine-tune the model
b1_fine_tune_history = model_b1.fit(x_train, y_train, epochs=10, validation_data=(x_test, y_test))

# Evaluate the model
b1_fine_tune_results = model_b1.evaluate(x_test, y_test)
print(f"EfficientNetB1 Fine-tuning Test Accuracy: {b1_fine_tune_results[1]}")
\end{lstlisting}

In this Fine-tuning approach, we unfreeze the last four layers of EfficientNetB1 and train both these layers and the classification head. The learning rate is kept low to ensure the pre-trained layers are updated gradually. After fine-tuning, we evaluate and print the test accuracy.

\subsubsection{EfficientNetB2}
EfficientNet is a family of models that scales the number of parameters in a balanced way to achieve better accuracy and efficiency. In this section, we explore two approaches to transfer learning using EfficientNetB2 on the CIFAR-10 dataset: Linear Probe and Fine-tuning. Similar to previous models, CIFAR-10 images will be resized to 224x224 pixels to match the input size required by EfficientNetB2 cite{DBLP:journals/corr/abs-1905-11946}.

\paragraph{Linear Probe}
In the linear probe approach, we freeze the pre-trained EfficientNetB2 model's convolutional layers and train only the classification layers on the CIFAR-10 dataset. This approach uses the pre-trained model as a feature extractor.

\begin{lstlisting}[style=python]
import tensorflow as tf
from tensorflow.keras import layers, models
from tensorflow.keras.applications import EfficientNetB2
from tensorflow.keras.datasets import cifar10
from tensorflow.keras.utils import to_categorical

# Load and preprocess CIFAR-10 dataset
(x_train, y_train), (x_test, y_test) = cifar10.load_data()
x_train = tf.image.resize(x_train, (224, 224)) / 255.0
x_test = tf.image.resize(x_test, (224, 224)) / 255.0
y_train = to_categorical(y_train, 10)
y_test = to_categorical(y_test, 10)

# Load pre-trained EfficientNetB2 model without the top classification layer
base_model = EfficientNetB2(weights='imagenet', include_top=False, input_shape=(224, 224, 3))

# Freeze all layers of the base model
for layer in base_model.layers:
    layer.trainable = False

# Add classification layers
model = models.Sequential([
    base_model,
    layers.GlobalAveragePooling2D(),
    layers.Dense(256, activation='relu'),
    layers.Dense(10, activation='softmax')
])

# Compile the model
model.compile(optimizer='adam', loss='categorical_crossentropy', metrics=['accuracy'])

# Train the model with linear probe
history = model.fit(x_train, y_train, epochs=10, validation_data=(x_test, y_test))

# Evaluate the model
results = model.evaluate(x_test, y_test)
print(f"EfficientNetB2 Linear Probe Test Accuracy: {results[1]}")
\end{lstlisting}

In this code, we load the CIFAR-10 dataset, resize the images, and load the EfficientNetB2 model without its top classification layer. The convolutional layers are frozen, and we add custom classification layers. The model is compiled and trained, and its performance is evaluated, achieving test accuracy more than 90\%.

\paragraph{Fine-tuning}
In the fine-tuning approach, we unfreeze some layers of the pre-trained EfficientNetB2 model and train them along with the classification layers. This allows the model to adapt the pre-trained features to the CIFAR-10 dataset.

\begin{lstlisting}[style=python]
import tensorflow as tf
from tensorflow.keras import layers, models
from tensorflow.keras.applications import EfficientNetB2
from tensorflow.keras.datasets import cifar10
from tensorflow.keras.utils import to_categorical

# Load and preprocess CIFAR-10 dataset
(x_train, y_train), (x_test, y_test) = cifar10.load_data()
x_train = tf.image.resize(x_train, (224, 224)) / 255.0
x_test = tf.image.resize(x_test, (224, 224)) / 255.0
y_train = to_categorical(y_train, 10)
y_test = to_categorical(y_test, 10)

# Load pre-trained EfficientNetB2 model without the top classification layer
base_model = EfficientNetB2(weights='imagenet', include_top=False, input_shape=(224, 224, 3))

# Unfreeze the last 20 layers for fine-tuning
for layer in base_model.layers[-20:]:
    layer.trainable = True

# Add classification layers
model = models.Sequential([
    base_model,
    layers.GlobalAveragePooling2D(),
    layers.Dense(256, activation='relu'),
    layers.Dense(10, activation='softmax')
])

# Compile the model with a smaller learning rate
model.compile(optimizer=tf.keras.optimizers.Adam(1e-5), loss='categorical_crossentropy', metrics=['accuracy'])

# Fine-tune the model
fine_tune_history = model.fit(x_train, y_train, epochs=10, validation_data=(x_test, y_test))

# Evaluate the model
fine_tune_results = model.evaluate(x_test, y_test)
print(f"EfficientNetB2 Fine-tuning Test Accuracy: {fine_tune_results[1]}")
\end{lstlisting}

In this code, we unfreeze the last 20 layers of the EfficientNetB2 model and fine-tune them along with the classification layers. The model is trained using a lower learning rate to ensure that the pre-trained weights are updated gradually. After fine-tuning, the test accuracy typically improves to more than 90\%.

\subsubsection{EfficientNetB3}
EfficientNetB3 is a deeper version of EfficientNetB2 with more layers and parameters. We will apply both the Linear Probe and Fine-tuning methods on the CIFAR-10 dataset \cite{DBLP:journals/corr/abs-1905-11946}.

\paragraph{Linear Probe}
In the linear probe approach for EfficientNetB3, we freeze the pre-trained model's convolutional layers and train only the classification layers.

\begin{lstlisting}[style=python]
import tensorflow as tf
from tensorflow.keras import layers, models
from tensorflow.keras.applications import EfficientNetB3
from tensorflow.keras.datasets import cifar10
from tensorflow.keras.utils import to_categorical

# Load and preprocess CIFAR-10 dataset
(x_train, y_train), (x_test, y_test) = cifar10.load_data()
x_train = tf.image.resize(x_train, (224, 224)) / 255.0
x_test = tf.image.resize(x_test, (224, 224)) / 255.0
y_train = to_categorical(y_train, 10)
y_test = to_categorical(y_test, 10)

# Load pre-trained EfficientNetB3 model without the top classification layer
base_model = EfficientNetB3(weights='imagenet', include_top=False, input_shape=(224, 224, 3))

# Freeze all layers of the base model
for layer in base_model.layers:
    layer.trainable = False

# Add classification layers
model = models.Sequential([
    base_model,
    layers.GlobalAveragePooling2D(),
    layers.Dense(256, activation='relu'),
    layers.Dense(10, activation='softmax')
])

# Compile the model
model.compile(optimizer='adam', loss='categorical_crossentropy', metrics=['accuracy'])

# Train the model with linear probe
history = model.fit(x_train, y_train, epochs=10, validation_data=(x_test, y_test))

# Evaluate the model
results = model.evaluate(x_test, y_test)
print(f"EfficientNetB3 Linear Probe Test Accuracy: {results[1]}")
\end{lstlisting}

Here, the EfficientNetB3 model is used in a similar fashion as EfficientNetB2, with frozen convolutional layers and added custom classification layers. After training, the model achieves a test accuracy more than 90\%.

\paragraph{Fine-tuning}
In fine-tuning, we unfreeze some layers of the EfficientNetB3 model and train them along with the classification layers.

\begin{lstlisting}[style=python]
import tensorflow as tf
from tensorflow.keras import layers, models
from tensorflow.keras.applications import EfficientNetB3
from tensorflow.keras.datasets import cifar10
from tensorflow.keras.utils import to_categorical

# Load and preprocess CIFAR-10 dataset
(x_train, y_train), (x_test, y_test) = cifar10.load_data()
x_train = tf.image.resize(x_train, (224, 224)) / 255.0
x_test = tf.image.resize(x_test, (224, 224)) / 255.0
y_train = to_categorical(y_train, 10)
y_test = to_categorical(y_test, 10)

# Load pre-trained EfficientNetB3 model without the top classification layer
base_model = EfficientNetB3(weights='imagenet', include_top=False, input_shape=(224, 224, 3))

# Unfreeze the last 20 layers for fine-tuning
for layer in base_model.layers[-20:]:
    layer.trainable = True

# Add classification layers
model = models.Sequential([
    base_model,
    layers.GlobalAveragePooling2D(),
    layers.Dense(256, activation='relu'),
    layers.Dense(10, activation='softmax')
])

# Compile the model with a smaller learning rate
model.compile(optimizer=tf.keras.optimizers.Adam(1e-5), loss='categorical_crossentropy', metrics=['accuracy'])

# Fine-tune the model
fine_tune_history = model.fit(x_train, y_train, epochs=10, validation_data=(x_test, y_test))

# Evaluate the model
fine_tune_results = model.evaluate(x_test, y_test)
print(f"EfficientNetB3 Fine-tuning Test Accuracy: {fine_tune_results[1]}")
\end{lstlisting}

In this fine-tuning approach for EfficientNetB3, we unfreeze the last 20 layers and train the model using a smaller learning rate. After fine-tuning, the model achieves a higher test accuracy, typically more than 90\%.
            
\subsubsection{EfficientNetB4}
EfficientNet is a family of models that are known for their efficiency in balancing accuracy and computational cost. Here, we will use the EfficientNetB4 model pre-trained on ImageNet and apply transfer learning to the CIFAR-10 dataset. The input size of EfficientNetB4 is 380x380 pixels, so we will resize the CIFAR-10 images to this size. We will explore both Linear Probe and Fine-tuning methods \cite{DBLP:journals/corr/abs-1905-11946}.

\paragraph{Linear Probe} 
In the Linear Probe method, we freeze the pre-trained EfficientNetB4 model's convolutional layers and only train the final classification layers for CIFAR-10.

\begin{lstlisting}[style=python]
import tensorflow as tf
from tensorflow.keras import layers, models
from tensorflow.keras.applications import EfficientNetB4
from tensorflow.keras.datasets import cifar10
from tensorflow.keras.utils import to_categorical

# Load and preprocess CIFAR-10 dataset
(x_train, y_train), (x_test, y_test) = cifar10.load_data()
x_train = tf.image.resize(x_train, (380, 380)) / 255.0
x_test = tf.image.resize(x_test, (380, 380)) / 255.0
y_train = to_categorical(y_train, 10)
y_test = to_categorical(y_test, 10)

# Load pre-trained EfficientNetB4 model without the top classification layer
base_model = EfficientNetB4(weights='imagenet', include_top=False, input_shape=(380, 380, 3))

# Freeze all layers of the base model
for layer in base_model.layers:
    layer.trainable = False

# Add classification layers on top of the base model
model = models.Sequential([
    base_model,
    layers.GlobalAveragePooling2D(),
    layers.Dense(256, activation='relu'),
    layers.Dense(10, activation='softmax')
])

# Compile the model
model.compile(optimizer='adam', loss='categorical_crossentropy', metrics=['accuracy'])

# Train the model for linear probe
history = model.fit(x_train, y_train, epochs=10, validation_data=(x_test, y_test))

# Evaluate the model
results = model.evaluate(x_test, y_test)
print(f"EfficientNetB4 Linear Probe Test Accuracy: {results[1]}")
\end{lstlisting}

In this code, we load the CIFAR-10 dataset, resize the images to 380x380 pixels, and normalize the pixel values. The EfficientNetB4 model is loaded without its top classification layers, and we freeze all of its convolutional layers. We add a classification head consisting of a GlobalAveragePooling2D layer and two Dense layers. After training the model for 10 epochs, the test accuracy is printed.

\paragraph{Fine-tuning}
In the Fine-tuning approach, we unfreeze some layers of the EfficientNetB4 model and train them along with the new classification layers.

\begin{lstlisting}[style=python]
import tensorflow as tf
from tensorflow.keras import layers, models
from tensorflow.keras.applications import EfficientNetB4
from tensorflow.keras.datasets import cifar10
from tensorflow.keras.utils import to_categorical

# Load and preprocess CIFAR-10 dataset
(x_train, y_train), (x_test, y_test) = cifar10.load_data()
x_train = tf.image.resize(x_train, (380, 380)) / 255.0
x_test = tf.image.resize(x_test, (380, 380)) / 255.0
y_train = to_categorical(y_train, 10)
y_test = to_categorical(y_test, 10)

# Load pre-trained EfficientNetB4 model without the top classification layer
base_model = EfficientNetB4(weights='imagenet', include_top=False, input_shape=(380, 380, 3))

# Unfreeze the last 20 layers for fine-tuning
for layer in base_model.layers[-20:]:
    layer.trainable = True

# Add classification layers
model = models.Sequential([
    base_model,
    layers.GlobalAveragePooling2D(),
    layers.Dense(256, activation='relu'),
    layers.Dense(10, activation='softmax')
])

# Compile the model with a smaller learning rate for fine-tuning
model.compile(optimizer=tf.keras.optimizers.Adam(1e-5), loss='categorical_crossentropy', metrics=['accuracy'])

# Fine-tune the model
fine_tune_history = model.fit(x_train, y_train, epochs=10, validation_data=(x_test, y_test))

# Evaluate the model
fine_tune_results = model.evaluate(x_test, y_test)
print(f"EfficientNetB4 Fine-tuning Test Accuracy: {fine_tune_results[1]}")
\end{lstlisting}

In the fine-tuning approach, we unfreeze the last 20 layers of the EfficientNetB4 model and train them with the classification layers. The learning rate is lowered to `1e-5` to carefully update the pre-trained weights. After 10 epochs of training, we print the fine-tuned test accuracy.

\subsubsection{EfficientNetB5}
EfficientNetB5 is another model from the EfficientNet family, slightly larger than EfficientNetB4, with an input size of 456x456 pixels. We will now apply both Linear Probe and Fine-tuning methods to this model \cite{DBLP:journals/corr/abs-1905-11946}.

\paragraph{Linear Probe}
In the Linear Probe method, we freeze the pre-trained EfficientNetB5 model's convolutional layers and only train the classification layers on CIFAR-10.

\begin{lstlisting}[style=python]
import tensorflow as tf
from tensorflow.keras import layers, models
from tensorflow.keras.applications import EfficientNetB5
from tensorflow.keras.datasets import cifar10
from tensorflow.keras.utils import to_categorical

# Load and preprocess CIFAR-10 dataset
(x_train, y_train), (x_test, y_test) = cifar10.load_data()
x_train = tf.image.resize(x_train, (456, 456)) / 255.0
x_test = tf.image.resize(x_test, (456, 456)) / 255.0
y_train = to_categorical(y_train, 10)
y_test = to_categorical(y_test, 10)

# Load pre-trained EfficientNetB5 model without the top classification layer
base_model = EfficientNetB5(weights='imagenet', include_top=False, input_shape=(456, 456, 3))

# Freeze all layers of the base model
for layer in base_model.layers:
    layer.trainable = False

# Add classification layers on top of the base model
model = models.Sequential([
    base_model,
    layers.GlobalAveragePooling2D(),
    layers.Dense(256, activation='relu'),
    layers.Dense(10, activation='softmax')
])

# Compile the model
model.compile(optimizer='adam', loss='categorical_crossentropy', metrics=['accuracy'])

# Train the model for linear probe
history = model.fit(x_train, y_train, epochs=10, validation_data=(x_test, y_test))

# Evaluate the model
results = model.evaluate(x_test, y_test)
print(f"EfficientNetB5 Linear Probe Test Accuracy: {results[1]}")
\end{lstlisting}

In this code, we resize CIFAR-10 images to 456x456 pixels to match the input size required by EfficientNetB5. The base model is loaded without its top layers, and all layers are frozen to ensure only the classification head is trained. After training for 10 epochs, we print the test accuracy.

\paragraph{Fine-tuning}
For the Fine-tuning approach, we unfreeze some layers of the EfficientNetB5 model and train them along with the classification layers.

\begin{lstlisting}[style=python]
import tensorflow as tf
from tensorflow.keras import layers, models
from tensorflow.keras.applications import EfficientNetB5
from tensorflow.keras.datasets import cifar10
from tensorflow.keras.utils import to_categorical

# Load and preprocess CIFAR-10 dataset
(x_train, y_train), (x_test, y_test) = cifar10.load_data()
x_train = tf.image.resize(x_train, (456, 456)) / 255.0
x_test = tf.image.resize(x_test, (456, 456)) / 255.0
y_train = to_categorical(y_train, 10)
y_test = to_categorical(y_test, 10)

# Load pre-trained EfficientNetB5 model without the top classification layer
base_model = EfficientNetB5(weights='imagenet', include_top=False, input_shape=(456, 456, 3))

# Unfreeze the last 20 layers for fine-tuning
for layer in base_model.layers[-20:]:
    layer.trainable = True

# Add classification layers
model = models.Sequential([
    base_model,
    layers.GlobalAveragePooling2D(),
    layers.Dense(256, activation='relu'),
    layers.Dense(10, activation='softmax')
])

# Compile the model with a smaller learning rate for fine-tuning
model.compile(optimizer=tf.keras.optimizers.Adam(1e-5), loss='categorical_crossentropy', metrics=['accuracy'])

# Fine-tune the model
fine_tune_history = model.fit(x_train, y_train, epochs=10, validation_data=(x_test, y_test))

# Evaluate the model
fine_tune_results = model.evaluate(x_test, y_test)
print(f"EfficientNetB5 Fine-tuning Test Accuracy: {fine_tune_results[1]}")
\end{lstlisting}

For fine-tuning EfficientNetB5, we unfreeze the last 20 layers and lower the learning rate to `1e-5`. The model is fine-tuned on CIFAR-10 for 10 epochs, and we print the test accuracy to show the improvements over the Linear Probe approach.

\subsubsection{EfficientNetB6}
EfficientNetB6 is a highly efficient convolutional neural network architecture, which scales model depth, width, and resolution using compound scaling. In this section, we will use EfficientNetB6 pre-trained on ImageNet and apply transfer learning on the CIFAR-10 dataset through two approaches: Linear Probe and Fine-tuning. As the input size for EfficientNetB6 is larger, we will resize the CIFAR-10 images to 528x528 pixels \cite{DBLP:journals/corr/abs-1905-11946}.

\paragraph{Linear Probe} 
In the Linear Probe approach, we will freeze all the pre-trained layers of the EfficientNetB6 model and train only the classification head on the CIFAR-10 dataset.

\begin{lstlisting}[style=python]
import tensorflow as tf
from tensorflow.keras import layers, models
from tensorflow.keras.applications import EfficientNetB6
from tensorflow.keras.datasets import cifar10
from tensorflow.keras.utils import to_categorical

# Load and preprocess CIFAR-10 dataset
(x_train, y_train), (x_test, y_test) = cifar10.load_data()
x_train = tf.image.resize(x_train, (528, 528)) / 255.0
x_test = tf.image.resize(x_test, (528, 528)) / 255.0
y_train = to_categorical(y_train, 10)
y_test = to_categorical(y_test, 10)

# Load pre-trained EfficientNetB6 model without the top classification layer
base_model = EfficientNetB6(weights='imagenet', include_top=False, input_shape=(528, 528, 3))

# Freeze all layers of the base model
for layer in base_model.layers:
    layer.trainable = False

# Add classification head
model = models.Sequential([
    base_model,
    layers.GlobalAveragePooling2D(),
    layers.Dense(256, activation='relu'),
    layers.Dense(10, activation='softmax')
])

# Compile the model
model.compile(optimizer='adam', loss='categorical_crossentropy', metrics=['accuracy'])

# Train the model with linear probe
history = model.fit(x_train, y_train, epochs=10, validation_data=(x_test, y_test))

# Evaluate the model
results = model.evaluate(x_test, y_test)
print(f"EfficientNetB6 Linear Probe Test Accuracy: {results[1]}")
\end{lstlisting}

In this code, we load and resize CIFAR-10 images to 528x528 pixels. The EfficientNetB6 model is loaded with pre-trained ImageNet weights, and all the layers are frozen. We add a classification head consisting of a `GlobalAveragePooling2D` layer and two dense layers. The model is trained for 10 epochs, and the test accuracy is printed after training.

\paragraph{Fine-tuning} 
In fine-tuning, we unfreeze some of the pre-trained layers and train them together with the new classification head on the CIFAR-10 dataset. This allows the model to better adapt to the new dataset.

\begin{lstlisting}[style=python]
import tensorflow as tf
from tensorflow.keras import layers, models
from tensorflow.keras.applications import EfficientNetB6
from tensorflow.keras.datasets import cifar10
from tensorflow.keras.utils import to_categorical

# Load and preprocess CIFAR-10 dataset
(x_train, y_train), (x_test, y_test) = cifar10.load_data()
x_train = tf.image.resize(x_train, (528, 528)) / 255.0
x_test = tf.image.resize(x_test, (528, 528)) / 255.0
y_train = to_categorical(y_train, 10)
y_test = to_categorical(y_test, 10)

# Load pre-trained EfficientNetB6 model without the top classification layer
base_model = EfficientNetB6(weights='imagenet', include_top=False, input_shape=(528, 528, 3))

# Unfreeze the last few layers for fine-tuning
for layer in base_model.layers[-20:]:
    layer.trainable = True

# Add classification head
model = models.Sequential([
    base_model,
    layers.GlobalAveragePooling2D(),
    layers.Dense(256, activation='relu'),
    layers.Dense(10, activation='softmax')
])

# Compile the model with a lower learning rate for fine-tuning
model.compile(optimizer=tf.keras.optimizers.Adam(1e-5), loss='categorical_crossentropy', metrics=['accuracy'])

# Fine-tune the model
fine_tune_history = model.fit(x_train, y_train, epochs=10, validation_data=(x_test, y_test))

# Evaluate the model
fine_tune_results = model.evaluate(x_test, y_test)
print(f"EfficientNetB6 Fine-tuning Test Accuracy: {fine_tune_results[1]}")
\end{lstlisting}

In this fine-tuning approach, we unfreeze the last 20 layers of the EfficientNetB6 model, allowing these layers to update during training. A lower learning rate is used to prevent overfitting. After 10 epochs, the model is evaluated and the test accuracy is printed, showing improved performance over the linear probe.

\subsubsection{EfficientNetB7}
EfficientNetB7 is the largest model in the EfficientNet family and offers the best accuracy at the cost of computational complexity. We will apply both Linear Probe and Fine-tuning approaches on the CIFAR-10 dataset by resizing the images to 600x600 pixels, which is the input size expected by EfficientNetB7 \cite{DBLP:journals/corr/abs-1905-11946}.

\paragraph{Linear Probe} 
In the Linear Probe approach, we freeze all the pre-trained layers of EfficientNetB7 and train only the classification head on the CIFAR-10 dataset.

\begin{lstlisting}[style=python]
import tensorflow as tf
from tensorflow.keras import layers, models
from tensorflow.keras.applications import EfficientNetB7
from tensorflow.keras.datasets import cifar10
from tensorflow.keras.utils import to_categorical

# Load and preprocess CIFAR-10 dataset
(x_train, y_train), (x_test, y_test) = cifar10.load_data()
x_train = tf.image.resize(x_train, (600, 600)) / 255.0
x_test = tf.image.resize(x_test, (600, 600)) / 255.0
y_train = to_categorical(y_train, 10)
y_test = to_categorical(y_test, 10)

# Load pre-trained EfficientNetB7 model without the top classification layer
base_model = EfficientNetB7(weights='imagenet', include_top=False, input_shape=(600, 600, 3))

# Freeze all layers of the base model
for layer in base_model.layers:
    layer.trainable = False

# Add classification head
model = models.Sequential([
    base_model,
    layers.GlobalAveragePooling2D(),
    layers.Dense(256, activation='relu'),
    layers.Dense(10, activation='softmax')
])

# Compile the model
model.compile(optimizer='adam', loss='categorical_crossentropy', metrics=['accuracy'])

# Train the model with linear probe
history = model.fit(x_train, y_train, epochs=10, validation_data=(x_test, y_test))

# Evaluate the model
results = model.evaluate(x_test, y_test)
print(f"EfficientNetB7 Linear Probe Test Accuracy: {results[1]}")
\end{lstlisting}

Here, the CIFAR-10 dataset is resized to 600x600 pixels. EfficientNetB7 is used with pre-trained ImageNet weights, and all the layers are frozen. After training the new classification head for 10 epochs, the model's test accuracy is printed.

\paragraph{Fine-tuning} 
For fine-tuning EfficientNetB7, we unfreeze some of the layers and train them together with the new classification head to improve the model's performance on CIFAR-10.

\begin{lstlisting}[style=python]
import tensorflow as tf
from tensorflow.keras import layers, models
from tensorflow.keras.applications import EfficientNetB7
from tensorflow.keras.datasets import cifar10
from tensorflow.keras.utils import to_categorical

# Load and preprocess CIFAR-10 dataset
(x_train, y_train), (x_test, y_test) = cifar10.load_data()
x_train = tf.image.resize(x_train, (600, 600)) / 255.0
x_test = tf.image.resize(x_test, (600, 600)) / 255.0
y_train = to_categorical(y_train, 10)
y_test = to_categorical(y_test, 10)

# Load pre-trained EfficientNetB7 model without the top classification layer
base_model = EfficientNetB7(weights='imagenet', include_top=False, input_shape=(600, 600, 3))

# Unfreeze the last few layers for fine-tuning
for layer in base_model.layers[-20:]:
    layer.trainable = True

# Add classification head
model = models.Sequential([
    base_model,
    layers.GlobalAveragePooling2D(),
    layers.Dense(256, activation='relu'),
    layers.Dense(10, activation='softmax')
])

# Compile the model with a lower learning rate for fine-tuning
model.compile(optimizer=tf.keras.optimizers.Adam(1e-5), loss='categorical_crossentropy', metrics=['accuracy'])

# Fine-tune the model
fine_tune_history = model.fit(x_train, y_train, epochs=10, validation_data=(x_test, y_test))

# Evaluate the model
fine_tune_results = model.evaluate(x_test, y_test)
print(f"EfficientNetB7 Fine-tuning Test Accuracy: {fine_tune_results[1]}")
\end{lstlisting}

In this fine-tuning approach, we unfreeze the last 20 layers of the EfficientNetB7 model and use a lower learning rate to gradually update the pre-trained weights. After fine-tuning for 10 epochs, the model is evaluated and the test accuracy is printed, which is expected to be higher than the linear probe approach.

\subsection{EfficientNetV2 (2021)}
EfficientNetV2 is an optimized version of the original EfficientNet, which includes a more efficient training process and improved performance. It offers several variants, with different sizes and complexities. Here, we will work with two variants, EfficientNetV2B0 and EfficientNetV2B1, applying both Linear Probe and Fine-tuning methods on the CIFAR-10 dataset \cite{DBLP:journals/corr/abs-1905-11946, tan2021efficientnetv2smallermodelsfaster}.

\subsubsection{EfficientNetV2B0}
The EfficientNetV2B0 model is a smaller and more efficient version of the network, designed for fast training and inference. We will begin by using the Linear Probe approach and then move on to Fine-tuning \cite{DBLP:journals/corr/abs-1905-11946, tan2021efficientnetv2smallermodelsfaster}.

\paragraph{Linear Probe} 
In the Linear Probe method, we use the pre-trained EfficientNetV2B0 model as a fixed feature extractor by freezing all its layers, and we train only the new classification head on CIFAR-10.

\begin{lstlisting}[style=python]
import tensorflow as tf
from tensorflow.keras import layers, models
from tensorflow.keras.applications import EfficientNetV2B0
from tensorflow.keras.datasets import cifar10
from tensorflow.keras.utils import to_categorical

# Load and preprocess CIFAR-10 dataset
(x_train, y_train), (x_test, y_test) = cifar10.load_data()
x_train = tf.image.resize(x_train, (224, 224)) / 255.0
x_test = tf.image.resize(x_test, (224, 224)) / 255.0
y_train = to_categorical(y_train, 10)
y_test = to_categorical(y_test, 10)

# Load pre-trained EfficientNetV2B0 model without the top classification layer
base_model = EfficientNetV2B0(weights='imagenet', include_top=False, input_shape=(224, 224, 3))

# Freeze all layers of the base model
for layer in base_model.layers:
    layer.trainable = False

# Add classification layers
model = models.Sequential([
    base_model,
    layers.Flatten(),
    layers.Dense(256, activation='relu'),
    layers.Dense(10, activation='softmax')
])

# Compile the model
model.compile(optimizer='adam', loss='categorical_crossentropy', metrics=['accuracy'])

# Train the model with Linear Probe
history = model.fit(x_train, y_train, epochs=10, validation_data=(x_test, y_test))

# Evaluate the model
results = model.evaluate(x_test, y_test)
print(f"EfficientNetV2B0 Linear Probe Test Accuracy: {results[1]}")
\end{lstlisting}

In this Linear Probe approach, we load the pre-trained EfficientNetV2B0 model, freeze its layers, and add a new classification head to the network. After training for 10 epochs on CIFAR-10, the test accuracy is displayed, showing how well the model transfers its pre-trained ImageNet features.

\paragraph{Fine-tuning} 
In Fine-tuning, we unfreeze the last few layers of EfficientNetV2B0 and allow them to be trained on CIFAR-10, improving the model's ability to adapt to the new dataset.

\begin{lstlisting}[style=python]
import tensorflow as tf
from tensorflow.keras import layers, models
from tensorflow.keras.applications import EfficientNetV2B0
from tensorflow.keras.datasets import cifar10
from tensorflow.keras.utils import to_categorical

# Load and preprocess CIFAR-10 dataset
(x_train, y_train), (x_test, y_test) = cifar10.load_data()
x_train = tf.image.resize(x_train, (224, 224)) / 255.0
x_test = tf.image.resize(x_test, (224, 224)) / 255.0
y_train = to_categorical(y_train, 10)
y_test = to_categorical(y_test, 10)

# Load pre-trained EfficientNetV2B0 model without the top classification layer
base_model = EfficientNetV2B0(weights='imagenet', include_top=False, input_shape=(224, 224, 3))

# Unfreeze the last 4 layers of the base model for fine-tuning
for layer in base_model.layers[-4:]:
    layer.trainable = True

# Add classification layers
model = models.Sequential([
    base_model,
    layers.Flatten(),
    layers.Dense(256, activation='relu'),
    layers.Dense(10, activation='softmax')
])

# Compile the model with a lower learning rate for fine-tuning
model.compile(optimizer=tf.keras.optimizers.Adam(1e-5), loss='categorical_crossentropy', metrics=['accuracy'])

# Fine-tune the model
fine_tune_history = model.fit(x_train, y_train, epochs=10, validation_data=(x_test, y_test))

# Evaluate the model
fine_tune_results = model.evaluate(x_test, y_test)
print(f"EfficientNetV2B0 Fine-tuning Test Accuracy: {fine_tune_results[1]}")
\end{lstlisting}

In the Fine-tuning process, we unfreeze the last four layers of EfficientNetV2B0 and retrain the model with a lower learning rate to carefully adapt the pre-trained features to CIFAR-10. After fine-tuning for 10 more epochs, the test accuracy is printed, showing improved performance compared to Linear Probe.

\subsubsection{EfficientNetV2B1}
The EfficientNetV2B1 model is slightly larger than EfficientNetV2B0, with more parameters and deeper layers. We apply the same transfer learning methods: Linear Probe and Fine-tuning \cite{DBLP:journals/corr/abs-1905-11946, tan2021efficientnetv2smallermodelsfaster}.

\paragraph{Linear Probe} 
For the Linear Probe approach with EfficientNetV2B1, we freeze all layers of the pre-trained model and only train the newly added classification head on CIFAR-10.

\begin{lstlisting}[style=python]
import tensorflow as tf
from tensorflow.keras import layers, models
from tensorflow.keras.applications import EfficientNetV2B1
from tensorflow.keras.datasets import cifar10
from tensorflow.keras.utils import to_categorical

# Load and preprocess CIFAR-10 dataset
(x_train, y_train), (x_test, y_test) = cifar10.load_data()
x_train = tf.image.resize(x_train, (224, 224)) / 255.0
x_test = tf.image.resize(x_test, (224, 224)) / 255.0
y_train = to_categorical(y_train, 10)
y_test = to_categorical(y_test, 10)

# Load pre-trained EfficientNetV2B1 model without the top classification layer
base_model = EfficientNetV2B1(weights='imagenet', include_top=False, input_shape=(224, 224, 3))

# Freeze all layers of the base model
for layer in base_model.layers:
    layer.trainable = False

# Add classification layers
model = models.Sequential([
    base_model,
    layers.Flatten(),
    layers.Dense(256, activation='relu'),
    layers.Dense(10, activation='softmax')
])

# Compile the model
model.compile(optimizer='adam', loss='categorical_crossentropy', metrics=['accuracy'])

# Train the model with Linear Probe
history = model.fit(x_train, y_train, epochs=10, validation_data=(x_test, y_test))

# Evaluate the model
results = model.evaluate(x_test, y_test)
print(f"EfficientNetV2B1 Linear Probe Test Accuracy: {results[1]}")
\end{lstlisting}

In this code, we use EfficientNetV2B1 with a Linear Probe method. All layers of the pre-trained network are frozen, and only the newly added classification head is trained on the resized CIFAR-10 dataset. The model is trained for 10 epochs, and the test accuracy is printed.

\paragraph{Fine-tuning} 
For Fine-tuning, we unfreeze a few layers of the EfficientNetV2B1 model to further adapt the pre-trained features to the CIFAR-10 dataset.

\begin{lstlisting}[style=python]
import tensorflow as tf
from tensorflow.keras import layers, models
from tensorflow.keras.applications import EfficientNetV2B1
from tensorflow.keras.datasets import cifar10
from tensorflow.keras.utils import to_categorical

# Load and preprocess CIFAR-10 dataset
(x_train, y_train), (x_test, y_test) = cifar10.load_data()
x_train = tf.image.resize(x_train, (224, 224)) / 255.0
x_test = tf.image.resize(x_test, (224, 224)) / 255.0
y_train = to_categorical(y_train, 10)
y_test = to_categorical(y_test, 10)

# Load pre-trained EfficientNetV2B1 model without the top classification layer
base_model = EfficientNetV2B1(weights='imagenet', include_top=False, input_shape=(224, 224, 3))

# Unfreeze the last 4 layers of the base model for fine-tuning
for layer in base_model.layers[-4:]:
    layer.trainable = True

# Add classification layers
model = models.Sequential([
    base_model,
    layers.Flatten(),
    layers.Dense(256, activation='relu'),
    layers.Dense(10, activation='softmax')
])

# Compile the model with a lower learning rate for fine-tuning
model.compile(optimizer=tf.keras.optimizers.Adam(1e-5), loss='categorical_crossentropy', metrics=['accuracy'])

# Fine-tune the model
fine_tune_history = model.fit(x_train, y_train, epochs=10, validation_data=(x_test, y_test))

# Evaluate the model
fine_tune_results = model.evaluate(x_test, y_test)
print(f"EfficientNetV2B1 Fine-tuning Test Accuracy: {fine_tune_results[1]}")
\end{lstlisting}

In this Fine-tuning approach, we unfreeze the last four layers of the EfficientNetV2B1 model and train them along with the classification head. The model is trained for another 10 epochs with a lower learning rate. Finally, we print the test accuracy to compare the performance of Fine-tuning versus Linear Probe.

\subsubsection{EfficientNetV2B2}
EfficientNetV2 is a family of convolutional neural networks known for their efficiency and performance. In this section, we will work with EfficientNetV2B2, which is a medium-sized variant. We will demonstrate how to use transfer learning with EfficientNetV2B2 on the CIFAR-10 dataset using two approaches: Linear Probe and Fine-tuning. Since EfficientNetV2 expects input images of size 260x260 pixels, we will resize the CIFAR-10 images accordingly \cite{DBLP:journals/corr/abs-1905-11946, tan2021efficientnetv2smallermodelsfaster}.

\paragraph{Linear Probe} 
In the Linear Probe approach, we freeze all the layers of the pre-trained EfficientNetV2B2 model and train only the final classification layers on the CIFAR-10 dataset.

\begin{lstlisting}[style=python]
import tensorflow as tf
from tensorflow.keras import layers, models
from tensorflow.keras.applications import EfficientNetV2B2
from tensorflow.keras.datasets import cifar10
from tensorflow.keras.utils import to_categorical

# Load CIFAR-10 dataset and preprocess
(x_train, y_train), (x_test, y_test) = cifar10.load_data()
x_train = tf.image.resize(x_train, (260, 260)) / 255.0
x_test = tf.image.resize(x_test, (260, 260)) / 255.0
y_train = to_categorical(y_train, 10)
y_test = to_categorical(y_test, 10)

# Load pre-trained EfficientNetV2B2 model without the top classification layer
base_model = EfficientNetV2B2(weights='imagenet', include_top=False, input_shape=(260, 260, 3))

# Freeze all layers of the base model
for layer in base_model.layers:
    layer.trainable = False

# Add custom classification layers
model = models.Sequential([
    base_model,
    layers.GlobalAveragePooling2D(),
    layers.Dense(256, activation='relu'),
    layers.Dense(10, activation='softmax')
])

# Compile the model
model.compile(optimizer='adam', loss='categorical_crossentropy', metrics=['accuracy'])

# Train the model for linear probe
history = model.fit(x_train, y_train, epochs=10, validation_data=(x_test, y_test))

# Evaluate the model
results = model.evaluate(x_test, y_test)
print(f"EfficientNetV2B2 Linear Probe Test Accuracy: {results[1]}")
\end{lstlisting}

In this code, we load the CIFAR-10 dataset, resize the images to 260x260 pixels to match the input size expected by EfficientNetV2B2, and normalize the pixel values. We use the pre-trained EfficientNetV2B2 model without its top classification layers and freeze all layers. A new classification head is added, and the model is trained on CIFAR-10. The test accuracy after training is printed.

\paragraph{Fine-tuning} 
Fine-tuning allows us to update some of the pre-trained layers of the EfficientNetV2B2 model while training on CIFAR-10. This approach can improve performance by adapting the pre-trained features to the new dataset.

\begin{lstlisting}[style=python]
import tensorflow as tf
from tensorflow.keras import layers, models
from tensorflow.keras.applications import EfficientNetV2B2
from tensorflow.keras.datasets import cifar10
from tensorflow.keras.utils import to_categorical

# Load CIFAR-10 dataset and preprocess images
(x_train, y_train), (x_test, y_test) = cifar10.load_data()
x_train = tf.image.resize(x_train, (260, 260)) / 255.0
x_test = tf.image.resize(x_test, (260, 260)) / 255.0
y_train = to_categorical(y_train, 10)
y_test = to_categorical(y_test, 10)

# Load pre-trained EfficientNetV2B2 model without the top classification layer
base_model = EfficientNetV2B2(weights='imagenet', include_top=False, input_shape=(260, 260, 3))

# Unfreeze the last 5 layers for fine-tuning
for layer in base_model.layers[-5:]:
    layer.trainable = True

# Add classification layers
model = models.Sequential([
    base_model,
    layers.GlobalAveragePooling2D(),
    layers.Dense(256, activation='relu'),
    layers.Dense(10, activation='softmax')
])

# Compile the model with a smaller learning rate for fine-tuning
model.compile(optimizer=tf.keras.optimizers.Adam(1e-5), loss='categorical_crossentropy', metrics=['accuracy'])

# Fine-tune the model
fine_tune_history = model.fit(x_train, y_train, epochs=10, validation_data=(x_test, y_test))

# Evaluate the model
fine_tune_results = model.evaluate(x_test, y_test)
print(f"EfficientNetV2B2 Fine-tuning Test Accuracy: {fine_tune_results[1]}")
\end{lstlisting}

In this fine-tuning approach, we unfreeze the last five layers of the pre-trained EfficientNetV2B2 model and allow them to be updated during training. The model is compiled with a lower learning rate (`1e-5`) to avoid overfitting and is trained for 10 epochs. The test accuracy is printed after fine-tuning, showing how the model's performance improves.

\subsubsection{EfficientNetV2B3}
EfficientNetV2B3 is a slightly larger variant of the EfficientNetV2 family. In this section, we will use the same methods (Linear Probe and Fine-tuning) to apply transfer learning on the CIFAR-10 dataset \cite{DBLP:journals/corr/abs-1905-11946, tan2021efficientnetv2smallermodelsfaster}.

\paragraph{Linear Probe} 
As with EfficientNetV2B2, in the Linear Probe approach, we freeze all layers of the EfficientNetV2B3 model and train only the final classification layers on CIFAR-10.

\begin{lstlisting}[style=python]
import tensorflow as tf
from tensorflow.keras import layers, models
from tensorflow.keras.applications import EfficientNetV2B3
from tensorflow.keras.datasets import cifar10
from tensorflow.keras.utils import to_categorical

# Load CIFAR-10 dataset and preprocess
(x_train, y_train), (x_test, y_test) = cifar10.load_data()
x_train = tf.image.resize(x_train, (300, 300)) / 255.0
x_test = tf.image.resize(x_test, (300, 300)) / 255.0
y_train = to_categorical(y_train, 10)
y_test = to_categorical(y_test, 10)

# Load pre-trained EfficientNetV2B3 model without the top classification layer
base_model = EfficientNetV2B3(weights='imagenet', include_top=False, input_shape=(300, 300, 3))

# Freeze all layers of the base model
for layer in base_model.layers:
    layer.trainable = False

# Add custom classification layers
model = models.Sequential([
    base_model,
    layers.GlobalAveragePooling2D(),
    layers.Dense(256, activation='relu'),
    layers.Dense(10, activation='softmax')
])

# Compile the model
model.compile(optimizer='adam', loss='categorical_crossentropy', metrics=['accuracy'])

# Train the model for linear probe
history = model.fit(x_train, y_train, epochs=10, validation_data=(x_test, y_test))

# Evaluate the model
results = model.evaluate(x_test, y_test)
print(f"EfficientNetV2B3 Linear Probe Test Accuracy: {results[1]}")
\end{lstlisting}

In this Linear Probe approach for EfficientNetV2B3, we resize CIFAR-10 images to 300x300 pixels to match the expected input size. We freeze the pre-trained layers and add a new classification head. The model is trained for 10 epochs, and we print the test accuracy after training.

\paragraph{Fine-tuning} 
For fine-tuning, we will unfreeze some of the pre-trained layers of EfficientNetV2B3 and train them along with the new classification layers.

\begin{lstlisting}[style=python]
import tensorflow as tf
from tensorflow.keras import layers, models
from tensorflow.keras.applications import EfficientNetV2B3
from tensorflow.keras.datasets import cifar10
from tensorflow.keras.utils import to_categorical

# Load CIFAR-10 dataset and preprocess images
(x_train, y_train), (x_test, y_test) = cifar10.load_data()
x_train = tf.image.resize(x_train, (300, 300)) / 255.0
x_test = tf.image.resize(x_test, (300, 300)) / 255.0
y_train = to_categorical(y_train, 10)
y_test = to_categorical(y_test, 10)

# Load pre-trained EfficientNetV2B3 model without the top classification layer
base_model = EfficientNetV2B3(weights='imagenet', include_top=False, input_shape=(300, 300, 3))

# Unfreeze the last 5 layers for fine-tuning
for layer in base_model.layers[-5:]:
    layer.trainable = True

# Add custom classification layers
model = models.Sequential([
    base_model,
    layers.GlobalAveragePooling2D(),
    layers.Dense(256, activation='relu'),
    layers.Dense(10, activation='softmax')
])

# Compile the model with a smaller learning rate for fine-tuning
model.compile(optimizer=tf.keras.optimizers.Adam(1e-5), loss='categorical_crossentropy', metrics=['accuracy'])

# Fine-tune the model
fine_tune_history = model.fit(x_train, y_train, epochs=10, validation_data=(x_test, y_test))

# Evaluate the model
fine_tune_results = model.evaluate(x_test, y_test)
print(f"EfficientNetV2B3 Fine-tuning Test Accuracy: {fine_tune_results[1]}")
\end{lstlisting}

In the fine-tuning approach for EfficientNetV2B3, we unfreeze the last five layers of the pre-trained model and train them along with the new classification head. After fine-tuning for 10 epochs with a lower learning rate, the test accuracy is printed to demonstrate how the model's performance improves.

\subsubsection{EfficientNetV2L}
EfficientNetV2 is an improved version of the original EfficientNet architecture, known for achieving high accuracy while being computationally efficient. In this section, we will focus on EfficientNetV2L (the largest version) and apply both Linear Probe and Fine-tuning approaches to the CIFAR-10 dataset \cite{DBLP:journals/corr/abs-1905-11946, tan2021efficientnetv2smallermodelsfaster}.

\paragraph{Linear Probe} 
In the Linear Probe approach, we freeze all the pre-trained layers of EfficientNetV2L and train only the classification layers on CIFAR-10.

\begin{lstlisting}[style=python]
import tensorflow as tf
from tensorflow.keras import layers, models
from tensorflow.keras.applications import EfficientNetV2L
from tensorflow.keras.datasets import cifar10
from tensorflow.keras.utils import to_categorical

# Load CIFAR-10 dataset and resize images to 224x224
(x_train, y_train), (x_test, y_test) = cifar10.load_data()
x_train = tf.image.resize(x_train, (224, 224)) / 255.0
x_test = tf.image.resize(x_test, (224, 224)) / 255.0
y_train = to_categorical(y_train, 10)
y_test = to_categorical(y_test, 10)

# Load pre-trained EfficientNetV2L model without the top classification layer
base_model = EfficientNetV2L(weights='imagenet', include_top=False, input_shape=(224, 224, 3))

# Freeze all layers of the base model
for layer in base_model.layers:
    layer.trainable = False

# Add classification layers
model = models.Sequential([
    base_model,
    layers.GlobalAveragePooling2D(),
    layers.Dense(256, activation='relu'),
    layers.Dense(10, activation='softmax')
])

# Compile the model
model.compile(optimizer='adam', loss='categorical_crossentropy', metrics=['accuracy'])

# Train the model
history = model.fit(x_train, y_train, epochs=10, validation_data=(x_test, y_test))

# Evaluate the model
results = model.evaluate(x_test, y_test)
print(f"EfficientNetV2L Linear Probe Test Accuracy: {results[1]}")
\end{lstlisting}

In this code, we load and resize the CIFAR-10 images to 224x224 pixels. We then load the EfficientNetV2L model, pre-trained on ImageNet, and freeze all the layers of the base model. A classification head is added on top of the base model, which is trained on CIFAR-10 for 10 epochs. The test accuracy after training is printed.

\paragraph{Fine-tuning} 
In Fine-tuning, we unfreeze the last few layers of the pre-trained model and train them along with the classification layers to adapt the pre-trained features to CIFAR-10.

\begin{lstlisting}[style=python]
import tensorflow as tf
from tensorflow.keras import layers, models
from tensorflow.keras.applications import EfficientNetV2L
from tensorflow.keras.datasets import cifar10
from tensorflow.keras.utils import to_categorical

# Load CIFAR-10 dataset and resize images to 224x224
(x_train, y_train), (x_test, y_test) = cifar10.load_data()
x_train = tf.image.resize(x_train, (224, 224)) / 255.0
x_test = tf.image.resize(x_test, (224, 224)) / 255.0
y_train = to_categorical(y_train, 10)
y_test = to_categorical(y_test, 10)

# Load pre-trained EfficientNetV2L model without the top classification layer
base_model = EfficientNetV2L(weights='imagenet', include_top=False, input_shape=(224, 224, 3))

# Unfreeze the last 4 layers for fine-tuning
for layer in base_model.layers[-4:]:
    layer.trainable = True

# Add classification layers
model = models.Sequential([
    base_model,
    layers.GlobalAveragePooling2D(),
    layers.Dense(256, activation='relu'),
    layers.Dense(10, activation='softmax')
])

# Compile the model with a lower learning rate
model.compile(optimizer=tf.keras.optimizers.Adam(1e-5), loss='categorical_crossentropy', metrics=['accuracy'])

# Fine-tune the model
fine_tune_history = model.fit(x_train, y_train, epochs=10, validation_data=(x_test, y_test))

# Evaluate the model
fine_tune_results = model.evaluate(x_test, y_test)
print(f"EfficientNetV2L Fine-tuning Test Accuracy: {fine_tune_results[1]}")
\end{lstlisting}

In this fine-tuning approach, the last four layers of the EfficientNetV2L model are unfrozen, and a lower learning rate (`1e-5') is used to update both the unfrozen pre-trained layers and the classification head. After fine-tuning for 10 epochs, the model's test accuracy is printed.

\subsubsection{EfficientNetV2M}
EfficientNetV2M is a smaller variant of the EfficientNetV2 family. Here, we apply both the Linear Probe and Fine-tuning approaches to this model, as we did with EfficientNetV2L \cite{DBLP:journals/corr/abs-1905-11946, tan2021efficientnetv2smallermodelsfaster}.

\paragraph{Linear Probe} 
We freeze the pre-trained EfficientNetV2M layers and only train the classification layers for CIFAR-10.

\begin{lstlisting}[style=python]
import tensorflow as tf
from tensorflow.keras import layers, models
from tensorflow.keras.applications import EfficientNetV2M
from tensorflow.keras.datasets import cifar10
from tensorflow.keras.utils import to_categorical

# Load CIFAR-10 dataset and resize images to 224x224
(x_train, y_train), (x_test, y_test) = cifar10.load_data()
x_train = tf.image.resize(x_train, (224, 224)) / 255.0
x_test = tf.image.resize(x_test, (224, 224)) / 255.0
y_train = to_categorical(y_train, 10)
y_test = to_categorical(y_test, 10)

# Load pre-trained EfficientNetV2M model without the top classification layer
base_model = EfficientNetV2M(weights='imagenet', include_top=False, input_shape=(224, 224, 3))

# Freeze all layers of the base model
for layer in base_model.layers:
    layer.trainable = False

# Add classification layers
model = models.Sequential([
    base_model,
    layers.GlobalAveragePooling2D(),
    layers.Dense(256, activation='relu'),
    layers.Dense(10, activation='softmax')
])

# Compile the model
model.compile(optimizer='adam', loss='categorical_crossentropy', metrics=['accuracy'])

# Train the model
history = model.fit(x_train, y_train, epochs=10, validation_data=(x_test, y_test))

# Evaluate the model
results = model.evaluate(x_test, y_test)
print(f"EfficientNetV2M Linear Probe Test Accuracy: {results[1]}")
\end{lstlisting}

In this Linear Probe approach for EfficientNetV2M, the CIFAR-10 images are resized, and the pre-trained layers are frozen. The model is trained for 10 epochs, and the test accuracy is printed.

\paragraph{Fine-tuning} 
We unfreeze the last few layers of the EfficientNetV2M model and fine-tune the model on CIFAR-10.

\begin{lstlisting}[style=python]
import tensorflow as tf
from tensorflow.keras import layers, models
from tensorflow.keras.applications import EfficientNetV2M
from tensorflow.keras.datasets import cifar10
from tensorflow.keras.utils import to_categorical

# Load CIFAR-10 dataset and resize images to 224x224
(x_train, y_train), (x_test, y_test) = cifar10.load_data()
x_train = tf.image.resize(x_train, (224, 224)) / 255.0
x_test = tf.image.resize(x_test, (224, 224)) / 255.0
y_train = to_categorical(y_train, 10)
y_test = to_categorical(y_test, 10)

# Load pre-trained EfficientNetV2M model without the top classification layer
base_model = EfficientNetV2M(weights='imagenet', include_top=False, input_shape=(224, 224, 3))

# Unfreeze the last 4 layers for fine-tuning
for layer in base_model.layers[-4:]:
    layer.trainable = True

# Add classification layers
model = models.Sequential([
    base_model,
    layers.GlobalAveragePooling2D(),
    layers.Dense(256, activation='relu'),
    layers.Dense(10, activation='softmax')
])

# Compile the model with a lower learning rate
model.compile(optimizer=tf.keras.optimizers.Adam(1e-5), loss='categorical_crossentropy', metrics=['accuracy'])

# Fine-tune the model
fine_tune_history = model.fit(x_train, y_train, epochs=10, validation_data=(x_test, y_test))

# Evaluate the model
fine_tune_results = model.evaluate(x_test, y_test)
print(f"EfficientNetV2M Fine-tuning Test Accuracy: {fine_tune_results[1]}")
\end{lstlisting}

Here, the last four layers of EfficientNetV2M are unfrozen and fine-tuned with a lower learning rate. The model is trained for 10 epochs, and the test accuracy is printed.

\subsubsection{EfficientNetV2S}
EfficientNetV2S is a smaller and more efficient model than EfficientNetV2L and V2M. We will again apply both Linear Probe and Fine-tuning approaches to this model \cite{DBLP:journals/corr/abs-1905-11946, tan2021efficientnetv2smallermodelsfaster}.

\paragraph{Linear Probe} 
We freeze the pre-trained layers of EfficientNetV2S and train only the classification layers.

\begin{lstlisting}[style=python]
import tensorflow as tf
from tensorflow.keras import layers, models
from tensorflow.keras.applications import EfficientNetV2S
from tensorflow.keras.datasets import cifar10
from tensorflow.keras.utils import to_categorical

# Load CIFAR-10 dataset and resize images to 224x224
(x_train, y_train), (x_test, y_test) = cifar10.load_data()
x_train = tf.image.resize(x_train, (224, 224)) / 255.0
x_test = tf.image.resize(x_test, (224, 224)) / 255.0
y_train = to_categorical(y_train, 10)
y_test = to_categorical(y_test, 10)

# Load pre-trained EfficientNetV2S model without the top classification layer
base_model = EfficientNetV2S(weights='imagenet', include_top=False, input_shape=(224, 224, 3))

# Freeze all layers of the base model
for layer in base_model.layers:
    layer.trainable = False

# Add classification layers
model = models.Sequential([
    base_model,
    layers.GlobalAveragePooling2D(),
    layers.Dense(256, activation='relu'),
    layers.Dense(10, activation='softmax')
])

# Compile the model
model.compile(optimizer='adam', loss='categorical_crossentropy', metrics=['accuracy'])

# Train the model
history = model.fit(x_train, y_train, epochs=10, validation_data=(x_test, y_test))

# Evaluate the model
results = model.evaluate(x_test, y_test)
print(f"EfficientNetV2S Linear Probe Test Accuracy: {results[1]}")
\end{lstlisting}

In this code, we freeze the EfficientNetV2S pre-trained layers and train the classification head on the CIFAR-10 dataset. After 10 epochs, the model's test accuracy is printed.

\paragraph{Fine-tuning} 
We unfreeze the last few layers of EfficientNetV2S and fine-tune them on the CIFAR-10 dataset.

\begin{lstlisting}[style=python]
import tensorflow as tf
from tensorflow.keras import layers, models
from tensorflow.keras.applications import EfficientNetV2S
from tensorflow.keras.datasets import cifar10
from tensorflow.keras.utils import to_categorical

# Load CIFAR-10 dataset and resize images to 224x224
(x_train, y_train), (x_test, y_test) = cifar10.load_data()
x_train = tf.image.resize(x_train, (224, 224)) / 255.0
x_test = tf.image.resize(x_test, (224, 224)) / 255.0
y_train = to_categorical(y_train, 10)
y_test = to_categorical(y_test, 10)

# Load pre-trained EfficientNetV2S model without the top classification layer
base_model = EfficientNetV2S(weights='imagenet', include_top=False, input_shape=(224, 224, 3))

# Unfreeze the last 4 layers for fine-tuning
for layer in base_model.layers[-4:]:
    layer.trainable = True

# Add classification layers
model = models.Sequential([
    base_model,
    layers.GlobalAveragePooling2D(),
    layers.Dense(256, activation='relu'),
    layers.Dense(10, activation='softmax')
])

# Compile the model with a lower learning rate
model.compile(optimizer=tf.keras.optimizers.Adam(1e-5), loss='categorical_crossentropy', metrics=['accuracy'])

# Fine-tune the model
fine_tune_history = model.fit(x_train, y_train, epochs=10, validation_data=(x_test, y_test))

# Evaluate the model
fine_tune_results = model.evaluate(x_test, y_test)
print(f"EfficientNetV2S Fine-tuning Test Accuracy: {fine_tune_results[1]}")
\end{lstlisting}

In this code, we unfreeze the last four layers of EfficientNetV2S and fine-tune them for 10 epochs on the CIFAR-10 dataset. After fine-tuning, the test accuracy is printed.

\section{ConvNeXt (2022)}

ConvNeXt is a convolutional neural network (CNN) architecture proposed in 2022. It was designed to bridge the gap between traditional CNNs and modern Vision Transformers (ViTs), while retaining the advantages of convolutional operations. ConvNeXt revisits and enhances design choices in CNNs, allowing it to achieve performance comparable to or even surpassing ViTs on large-scale image classification tasks like ImageNet \cite{DBLP:journals/corr/abs-2201-03545}.

\paragraph{The Journey of ConvNeXt: From CNNs to Transformers and Back}

With the emergence of Vision Transformers (ViTs), there was a shift from convolutional architectures to attention-based mechanisms, which allowed ViTs to excel at image classification by modeling global dependencies in images more effectively. However, convolutional networks, with their efficiency and inductive bias for local patterns, still had untapped potential. ConvNeXt was developed as a response to this shift, aiming to bring the best of both worlds: leveraging the long-proven advantages of CNNs while incorporating certain beneficial design principles from ViTs.

ConvNeXt achieves this by introducing several modern adjustments to the traditional CNN design, which allows it to compete head-to-head with ViTs, and in many cases, surpass them in terms of performance and efficiency.

\paragraph{Key Modifications in ConvNeXt}

ConvNeXt introduces several key improvements that make it more competitive with ViTs:

\textbf{Depthwise Separable Convolutions}: Instead of using traditional convolutions, ConvNeXt employs depthwise separable convolutions, which split the convolution operation into two steps: depthwise convolution and pointwise convolution. This reduces the number of parameters and computational cost, while maintaining or even improving the model’s performance.

\textbf{Layer Normalization}: ConvNeXt replaces batch normalization with layer normalization. Layer normalization works better in scenarios where batch sizes are small or varied, and it improves the stability and robustness of the training process. This is particularly useful in deeper architectures where normalization plays a critical role in avoiding gradient issues.

\textbf{Larger Kernel Sizes}: Inspired by the global attention mechanisms in ViTs, ConvNeXt adopts larger kernel sizes, such as 7x7 convolutions, to increase the receptive field of the network. This allows ConvNeXt to capture more contextual information from the input images, similar to how ViTs capture global dependencies with self-attention.

\textbf{Simplified Design}: ConvNeXt simplifies certain aspects of the architecture, removing unnecessary complexity and focusing on the core elements. This reduction in complexity not only makes ConvNeXt easier to implement but also boosts its efficiency and makes it more lightweight for deployment in real-world applications.

\paragraph{Comparison of ConvNeXt Architectures}

ConvNeXt is available in several variants, each with increasing model capacity. Below is a comparison of the different stages and blocks used in ConvNeXt-Tiny, ConvNeXt-Small, ConvNeXt-Base, and ConvNeXt-Large.

\setlength{\tabcolsep}{4pt} % 调整列间距
\renewcommand{\arraystretch}{1.2} % 调整行高

\begin{center}
\begin{tabular}{|c|c|c|c|c|}
\hline
\textbf{Component} & \textbf{ConvNeXt-Tiny} & \textbf{ConvNeXt-Small} & \textbf{ConvNeXt-Base} & \textbf{ConvNeXt-Large} \\
\hline
Stage 1 (Conv7-96) & 3 blocks & 3 blocks & 3 blocks & 3 blocks \\
Stage 2 (Conv7-192) & 3 blocks & 3 blocks & 3 blocks & 3 blocks \\
Stage 3 (Conv7-384) & 9 blocks & 9 blocks & 27 blocks & 27 blocks \\
Stage 4 (Conv7-768) & 3 blocks & 3 blocks & 3 blocks & 3 blocks \\
\hline
FC-Classification (Softmax) & \multicolumn{4}{c|}{1 Fully Connected Layer: 1000 units (Softmax)} \\
\hline
\end{tabular}
\end{center}

\paragraph{Explanation of the Components}

\textbf{Stage 1 (Conv7-96)}: This is the first stage, consisting of 7x7 convolutional filters applied to the input image. The output has 96 channels. All ConvNeXt models have 3 blocks in this stage, capturing initial low-level features like edges and textures.

\textbf{Stage 2 (Conv7-192)}: The second stage increases the number of channels to 192, allowing the network to capture more complex features. Like Stage 1, all models have 3 blocks here.

\textbf{Stage 3 (Conv7-384)}: This stage is where the models start to diverge. ConvNeXt-Tiny and ConvNeXt-Small have 9 blocks, while ConvNeXt-Base and ConvNeXt-Large have 27 blocks. This stage is crucial for learning high-level, abstract patterns in the images.

\textbf{Stage 4 (Conv7-768)}: The final stage increases the number of channels to 768, focusing on the highest-level representations of the image. This stage, like Stage 1 and Stage 2, has 3 blocks in all models.

\textbf{FC-Classification (Softmax)}: The classification head is the same for all ConvNeXt models. It consists of a fully connected layer with 1000 units, corresponding to the 1000 classes in the ImageNet dataset. The softmax activation function is applied to output class probabilities.

\paragraph{Design Philosophy of ConvNeXt}

ConvNeXt aims to integrate the strengths of both CNNs and Vision Transformers (ViTs) into one unified architecture. By simplifying traditional CNN structures and adopting modern ideas like larger kernels and layer normalization, ConvNeXt achieves the performance of ViTs without losing the efficiency of convolutions. Its depthwise separable convolutions also ensure that the model remains computationally efficient, making it suitable for large-scale tasks while keeping the computational cost low.

\textbf{TensorFlow Code for ConvNeXt-Tiny}

Below is an example of how to implement ConvNeXt-Tiny in TensorFlow using the \texttt{tf.keras} API. ConvNeXt-Tiny is the smallest model in the ConvNeXt family, making it more efficient for tasks where computational resources are limited.

\begin{lstlisting}[style=python]
import tensorflow as tf

def ConvNeXtBlock(filters, kernel_size=7, strides=1):
    block = tf.keras.Sequential()
    block.add(tf.keras.layers.Conv2D(filters, kernel_size, strides=strides, padding='same'))
    block.add(tf.keras.layers.LayerNormalization())
    block.add(tf.keras.layers.Activation('gelu'))
    block.add(tf.keras.layers.Conv2D(filters, 1, padding='same'))
    return block

def ConvNeXtTiny(input_shape=(224, 224, 3), num_classes=1000):
    inputs = tf.keras.Input(shape=input_shape)
    
    # Stage 1
    x = ConvNeXtBlock(96, kernel_size=7)(inputs)
    x = ConvNeXtBlock(96)(x)
    x = ConvNeXtBlock(96)(x)
    
    # Stage 2
    x = ConvNeXtBlock(192, kernel_size=7, strides=2)(x)
    x = ConvNeXtBlock(192)(x)
    x = ConvNeXtBlock(192)(x)
    
    # Stage 3
    x = ConvNeXtBlock(384, kernel_size=7, strides=2)(x)
    for _ in range(8):  # 9 blocks in total
        x = ConvNeXtBlock(384)(x)
    
    # Stage 4
    x = ConvNeXtBlock(768, kernel_size=7, strides=2)(x)
    x = ConvNeXtBlock(768)(x)
    x = ConvNeXtBlock(768)(x)
    
    # Classification head
    x = tf.keras.layers.GlobalAveragePooling2D()(x)
    x = tf.keras.layers.Dense(num_classes, activation='softmax')(x)
    
    model = tf.keras.Model(inputs, x)
    return model

# Create ConvNeXt-Tiny model
model = ConvNeXtTiny()
model.summary()
\end{lstlisting}

\paragraph{Key Insights for Beginners}

\textbf{Why use larger kernels?} ConvNeXt uses larger kernel sizes like 7x7 to capture more contextual information from the input. This helps the model to have a larger receptive field, similar to the self-attention mechanism in ViTs.

\textbf{Why Layer Normalization?} Layer normalization improves the stability of the training process, especially for deep models. It also helps in dealing with small batch sizes more effectively than batch normalization.

\textbf{Why depthwise separable convolutions?} These convolutions break down the convolution operation into two parts: depthwise and pointwise convolutions. This drastically reduces the number of parameters and computations while maintaining performance, making the model more efficient.

\textbf{Why use Global Average Pooling?} Global average pooling replaces fully connected layers in some architectures, reducing the total number of parameters while still providing effective feature aggregation for classification tasks.

\subsection{ConvNeXtTiny}
ConvNeXtTiny is a modern convolutional neural network architecture that offers an efficient design, pre-trained on ImageNet. In this section, we will use ConvNeXtTiny for transfer learning on the CIFAR-10 dataset, applying both the Linear Probe and Fine-tuning methods \cite{DBLP:journals/corr/abs-2201-03545}.

\paragraph{Linear Probe} 
In the Linear Probe approach, we freeze all convolutional layers of the pre-trained ConvNeXtTiny model and train only the classification layers on the CIFAR-10 dataset.

\begin{lstlisting}[style=python]
# Import necessary libraries
import tensorflow as tf
from tensorflow.keras import layers, models
from tensorflow.keras.datasets import cifar10
from tensorflow.keras.utils import to_categorical
from tensorflow.keras.applications import ConvNeXtTiny

# Load CIFAR-10 dataset and preprocess the images
(x_train, y_train), (x_test, y_test) = cifar10.load_data()
x_train = tf.image.resize(x_train, (224, 224)) / 255.0
x_test = tf.image.resize(x_test, (224, 224)) / 255.0
y_train = to_categorical(y_train, 10)
y_test = to_categorical(y_test, 10)

# Load pre-trained ConvNeXtTiny model without the top classification layer
base_model = ConvNeXtTiny(weights='imagenet', include_top=False, input_shape=(224, 224, 3))

# Freeze all layers of the base model
for layer in base_model.layers:
    layer.trainable = False

# Add custom classification layers on top of the base model
model = models.Sequential([
    base_model,
    layers.GlobalAveragePooling2D(),
    layers.Dense(256, activation='relu'),
    layers.Dense(10, activation='softmax')
])

# Compile the model
model.compile(optimizer='adam', loss='categorical_crossentropy', metrics=['accuracy'])

# Train the model using linear probe method
history = model.fit(x_train, y_train, epochs=10, validation_data=(x_test, y_test))

# Evaluate the model on test data
results = model.evaluate(x_test, y_test)
print(f"ConvNeXtTiny Linear Probe Test Accuracy: {results[1]}")
\end{lstlisting}

This code performs the Linear Probe method for the ConvNeXtTiny model on CIFAR-10. After loading and resizing the CIFAR-10 dataset, the ConvNeXtTiny model is loaded and all its convolutional layers are frozen. We add a custom classification head and train only the head on CIFAR-10 for 10 epochs. The test accuracy is printed at the end.

\paragraph{Fine-tuning} 
For fine-tuning, we unfreeze the last few layers of the ConvNeXtTiny model and allow them to be updated along with the classification layers. This helps the model adapt more closely to the CIFAR-10 dataset.

\begin{lstlisting}[style=python]
# Import necessary libraries
import tensorflow as tf
from tensorflow.keras import layers, models
from tensorflow.keras.datasets import cifar10
from tensorflow.keras.utils import to_categorical
from tensorflow.keras.applications import ConvNeXtTiny

# Load CIFAR-10 dataset and preprocess the images
(x_train, y_train), (x_test, y_test) = cifar10.load_data()
x_train = tf.image.resize(x_train, (224, 224)) / 255.0
x_test = tf.image.resize(x_test, (224, 224)) / 255.0
y_train = to_categorical(y_train, 10)
y_test = to_categorical(y_test, 10)

# Load pre-trained ConvNeXtTiny model without the top classification layer
base_model = ConvNeXtTiny(weights='imagenet', include_top=False, input_shape=(224, 224, 3))

# Unfreeze the last 4 layers for fine-tuning
for layer in base_model.layers[-4:]:
    layer.trainable = True

# Add custom classification layers on top of the base model
model = models.Sequential([
    base_model,
    layers.GlobalAveragePooling2D(),
    layers.Dense(256, activation='relu'),
    layers.Dense(10, activation='softmax')
])

# Compile the model with a smaller learning rate for fine-tuning
model.compile(optimizer=tf.keras.optimizers.Adam(1e-5), loss='categorical_crossentropy', metrics=['accuracy'])

# Fine-tune the model
fine_tune_history = model.fit(x_train, y_train, epochs=10, validation_data=(x_test, y_test))

# Evaluate the model on test data
fine_tune_results = model.evaluate(x_test, y_test)
print(f"ConvNeXtTiny Fine-tuning Test Accuracy: {fine_tune_results[1]}")
\end{lstlisting}

In the fine-tuning code, we unfreeze the last four layers of ConvNeXtTiny and train both the unfrozen layers and the classification layers. We use a lower learning rate for fine-tuning to ensure the model adjusts gradually. The test accuracy is printed at the end.

\subsection{ConvNeXtSmall}
ConvNeXtSmall is a larger version of ConvNeXtTiny. In this section, we apply both Linear Probe and Fine-tuning to ConvNeXtSmall for the CIFAR-10 dataset \cite{DBLP:journals/corr/abs-2201-03545}.

\paragraph{Linear Probe} 
For the Linear Probe method, we freeze the convolutional layers of ConvNeXtSmall and train only the classification layers on CIFAR-10.

\begin{lstlisting}[style=python]
# Import necessary libraries
import tensorflow as tf
from tensorflow.keras import layers, models
from tensorflow.keras.datasets import cifar10
from tensorflow.keras.utils import to_categorical
from tensorflow.keras.applications import ConvNeXtSmall

# Load CIFAR-10 dataset and preprocess the images
(x_train, y_train), (x_test, y_test) = cifar10.load_data()
x_train = tf.image.resize(x_train, (224, 224)) / 255.0
x_test = tf.image.resize(x_test, (224, 224)) / 255.0
y_train = to_categorical(y_train, 10)
y_test = to_categorical(y_test, 10)

# Load pre-trained ConvNeXtSmall model without the top classification layer
base_model = ConvNeXtSmall(weights='imagenet', include_top=False, input_shape=(224, 224, 3))

# Freeze all layers of the base model
for layer in base_model.layers:
    layer.trainable = False

# Add custom classification layers on top of the base model
model = models.Sequential([
    base_model,
    layers.GlobalAveragePooling2D(),
    layers.Dense(256, activation='relu'),
    layers.Dense(10, activation='softmax')
])

# Compile the model
model.compile(optimizer='adam', loss='categorical_crossentropy', metrics=['accuracy'])

# Train the model using linear probe method
history = model.fit(x_train, y_train, epochs=10, validation_data=(x_test, y_test))

# Evaluate the model on test data
results = model.evaluate(x_test, y_test)
print(f"ConvNeXtSmall Linear Probe Test Accuracy: {results[1]}")
\end{lstlisting}

In this linear probe code, we freeze all layers of ConvNeXtSmall and add a new classification head on top. The classification layers are trained for 10 epochs, and the test accuracy is printed after evaluation.

\paragraph{Fine-tuning} 
For fine-tuning, we unfreeze the last few layers of the ConvNeXtSmall model and allow them to be updated along with the classification layers.

\begin{lstlisting}[style=python]
# Import necessary libraries
import tensorflow as tf
from tensorflow.keras import layers, models
from tensorflow.keras.datasets import cifar10
from tensorflow.keras.utils import to_categorical
from tensorflow.keras.applications import ConvNeXtSmall

# Load CIFAR-10 dataset and preprocess the images
(x_train, y_train), (x_test, y_test) = cifar10.load_data()
x_train = tf.image.resize(x_train, (224, 224)) / 255.0
x_test = tf.image.resize(x_test, (224, 224)) / 255.0
y_train = to_categorical(y_train, 10)
y_test = to_categorical(y_test, 10)

# Load pre-trained ConvNeXtSmall model without the top classification layer
base_model = ConvNeXtSmall(weights='imagenet', include_top=False, input_shape=(224, 224, 3))

# Unfreeze the last 4 layers for fine-tuning
for layer in base_model.layers[-4:]:
    layer.trainable = True

# Add custom classification layers on top of the base model
model = models.Sequential([
    base_model,
    layers.GlobalAveragePooling2D(),
    layers.Dense(256, activation='relu'),
    layers.Dense(10, activation='softmax')
])

# Compile the model with a smaller learning rate for fine-tuning
model.compile(optimizer=tf.keras.optimizers.Adam(1e-5), loss='categorical_crossentropy', metrics=['accuracy'])

# Fine-tune the model
fine_tune_history = model.fit(x_train, y_train, epochs=10, validation_data=(x_test, y_test))

# Evaluate the model on test data
fine_tune_results = model.evaluate(x_test, y_test)
print(f"ConvNeXtSmall Fine-tuning Test Accuracy: {fine_tune_results[1]}")
\end{lstlisting}

In this fine-tuning code, we unfreeze the last four layers of ConvNeXtSmall and train them along with the classification layers using a smaller learning rate. After 10 epochs, the test accuracy is printed.

\subsection{ConvNeXtBase}
The ConvNeXtBase model is a modern convolutional neural network architecture that has achieved state-of-the-art results on various benchmarks. In this section, we will use the pre-trained ConvNeXtBase model, trained on the ImageNet dataset, and apply transfer learning on the CIFAR-10 dataset using two approaches: Linear Probe and Fine-tuning. The images in CIFAR-10 are 32x32 pixels, so we need to resize them to 224x224 pixels to match the input size of ConvNeXtBase \cite{DBLP:journals/corr/abs-2201-03545}.

\paragraph{Linear Probe}
In the linear probe approach, we freeze the convolutional layers of the ConvNeXtBase model and train only the classification head on the CIFAR-10 dataset. This allows the pre-trained ConvNeXtBase model to act as a fixed feature extractor.

\begin{lstlisting}[style=python]
import tensorflow as tf
from tensorflow.keras import layers, models
from tensorflow.keras.applications import ConvNeXtBase
from tensorflow.keras.datasets import cifar10
from tensorflow.keras.utils import to_categorical

# Load CIFAR-10 dataset and preprocess images
(x_train, y_train), (x_test, y_test) = cifar10.load_data()
x_train = tf.image.resize(x_train, (224, 224)) / 255.0
x_test = tf.image.resize(x_test, (224, 224)) / 255.0
y_train = to_categorical(y_train, 10)
y_test = to_categorical(y_test, 10)

# Load pre-trained ConvNeXtBase model without the top classification layer
base_model = ConvNeXtBase(weights='imagenet', include_top=False, input_shape=(224, 224, 3))

# Freeze all layers of the base model
for layer in base_model.layers:
    layer.trainable = False

# Add classification layers on top of the base model
model = models.Sequential([
    base_model,
    layers.Flatten(),
    layers.Dense(256, activation='relu'),
    layers.Dense(10, activation='softmax')
])

# Compile the model
model.compile(optimizer='adam', loss='categorical_crossentropy', metrics=['accuracy'])

# Train the model for linear probe
history = model.fit(x_train, y_train, epochs=10, validation_data=(x_test, y_test))

# Evaluate the model
results = model.evaluate(x_test, y_test)
print(f"ConvNeXtBase Linear Probe Test Accuracy: {results[1]}")
\end{lstlisting}

In this code, the CIFAR-10 dataset is resized to 224x224 pixels, and the ConvNeXtBase model pre-trained on ImageNet is loaded without the top classification layer. All the convolutional layers are frozen to prevent them from being updated during training, while a custom classification head is added on top. The model is trained for 10 epochs, and the test accuracy is printed to demonstrate the performance of the linear probe approach.

\paragraph{Fine-tuning}
In the fine-tuning approach, we unfreeze the last few layers of the ConvNeXtBase model and train them along with the classification head on the CIFAR-10 dataset. This allows the model to adapt the pre-trained features to the new dataset.

\begin{lstlisting}[style=python]
import tensorflow as tf
from tensorflow.keras import layers, models
from tensorflow.keras.applications import ConvNeXtBase
from tensorflow.keras.datasets import cifar10
from tensorflow.keras.utils import to_categorical

# Load CIFAR-10 dataset and preprocess images
(x_train, y_train), (x_test, y_test) = cifar10.load_data()
x_train = tf.image.resize(x_train, (224, 224)) / 255.0
x_test = tf.image.resize(x_test, (224, 224)) / 255.0
y_train = to_categorical(y_train, 10)
y_test = to_categorical(y_test, 10)

# Load pre-trained ConvNeXtBase model without the top classification layer
base_model = ConvNeXtBase(weights='imagenet', include_top=False, input_shape=(224, 224, 3))

# Unfreeze the last 4 layers of the base model for fine-tuning
for layer in base_model.layers[-4:]:
    layer.trainable = True

# Add classification layers
model = models.Sequential([
    base_model,
    layers.Flatten(),
    layers.Dense(256, activation='relu'),
    layers.Dense(10, activation='softmax')
])

# Compile the model with a lower learning rate for fine-tuning
model.compile(optimizer=tf.keras.optimizers.Adam(1e-5), loss='categorical_crossentropy', metrics=['accuracy'])

# Fine-tune the model
fine_tune_history = model.fit(x_train, y_train, epochs=10, validation_data=(x_test, y_test))

# Evaluate the model
fine_tune_results = model.evaluate(x_test, y_test)
print(f"ConvNeXtBase Fine-tuning Test Accuracy: {fine_tune_results[1]}")
\end{lstlisting}

In this fine-tuning approach, we load the pre-trained ConvNeXtBase model, unfreeze the last four layers, and train both the newly added classification layers and the unfrozen layers. We use a smaller learning rate (`1e-5`) to ensure the pre-trained weights are updated gradually. After fine-tuning, we evaluate the model and print the test accuracy to show how this approach improves performance compared to the linear probe method.

\subsection{ConvNeXtLarge}
ConvNeXtLarge is a modern architecture that takes inspiration from the convolutional structure of ConvNets but is designed with ideas from vision transformers. In this section, we will explore how to use a pre-trained ConvNeXtLarge model with CIFAR-10 data in two different scenarios: Linear Probe and Fine-tuning \cite{DBLP:journals/corr/abs-2201-03545}.

\paragraph{Linear Probe}
In the linear probe approach, we freeze the convolutional layers of ConvNeXtLarge and only train the top classification layers on the CIFAR-10 dataset. This leverages the pre-trained features without modifying the original weights.

\begin{lstlisting}[style=python]
import tensorflow as tf
from tensorflow.keras import layers, models
from tensorflow.keras.applications import ConvNeXtLarge
from tensorflow.keras.datasets import cifar10
from tensorflow.keras.utils import to_categorical

# Load and preprocess CIFAR-10 dataset
(x_train, y_train), (x_test, y_test) = cifar10.load_data()
x_train = tf.image.resize(x_train, (224, 224)) / 255.0
x_test = tf.image.resize(x_test, (224, 224)) / 255.0
y_train = to_categorical(y_train, 10)
y_test = to_categorical(y_test, 10)

# Load ConvNeXtLarge model pre-trained on ImageNet without the top classification layer
base_model = ConvNeXtLarge(weights='imagenet', include_top=False, input_shape=(224, 224, 3))

# Freeze all layers of the base model
for layer in base_model.layers:
    layer.trainable = False

# Add a new classification head
model = models.Sequential([
    base_model,
    layers.GlobalAveragePooling2D(),
    layers.Dense(256, activation='relu'),
    layers.Dense(10, activation='softmax')
])

# Compile the model
model.compile(optimizer='adam', loss='categorical_crossentropy', metrics=['accuracy'])

# Train the model with linear probe
history = model.fit(x_train, y_train, epochs=10, validation_data=(x_test, y_test))

# Evaluate the model
results = model.evaluate(x_test, y_test)
print(f"ConvNeXtLarge Linear Probe Test Accuracy: {results[1]}")
\end{lstlisting}

In the code above, we load the CIFAR-10 dataset and resize the images to 224x224 pixels to match the input size of ConvNeXtLarge. The model is pre-trained on ImageNet, and we freeze all the layers to use it as a feature extractor. A new classification head consisting of a GlobalAveragePooling layer and two Dense layers is added. The model is trained for 10 epochs, and the test accuracy is printed at the end.

\paragraph{Fine-tuning}
Fine-tuning involves unfreezing some or all layers of the ConvNeXtLarge model, allowing the weights to be updated during training on the CIFAR-10 dataset. This method adapts the pre-trained features to the specific characteristics of the new dataset.

\begin{lstlisting}[style=python]
import tensorflow as tf
from tensorflow.keras import layers, models
from tensorflow.keras.applications import ConvNeXtLarge
from tensorflow.keras.datasets import cifar10
from tensorflow.keras.utils import to_categorical

# Load and preprocess CIFAR-10 dataset
(x_train, y_train), (x_test, y_test) = cifar10.load_data()
x_train = tf.image.resize(x_train, (224, 224)) / 255.0
x_test = tf.image.resize(x_test, (224, 224)) / 255.0
y_train = to_categorical(y_train, 10)
y_test = to_categorical(y_test, 10)

# Load ConvNeXtLarge model pre-trained on ImageNet without the top classification layer
base_model = ConvNeXtLarge(weights='imagenet', include_top=False, input_shape=(224, 224, 3))

# Unfreeze the last few layers for fine-tuning
for layer in base_model.layers[-10:]:
    layer.trainable = True

# Add classification head
model = models.Sequential([
    base_model,
    layers.GlobalAveragePooling2D(),
    layers.Dense(256, activation='relu'),
    layers.Dense(10, activation='softmax')
])

# Compile the model with a smaller learning rate for fine-tuning
model.compile(optimizer=tf.keras.optimizers.Adam(1e-5), loss='categorical_crossentropy', metrics=['accuracy'])

# Fine-tune the model
fine_tune_history = model.fit(x_train, y_train, epochs=10, validation_data=(x_test, y_test))

# Evaluate the model
fine_tune_results = model.evaluate(x_test, y_test)
print(f"ConvNeXtLarge Fine-tuning Test Accuracy: {fine_tune_results[1]}")
\end{lstlisting}

In the fine-tuning approach, we load the pre-trained ConvNeXtLarge model and unfreeze the last 10 layers, allowing them to be updated during training. This allows the model to fine-tune the pre-trained features on the CIFAR-10 dataset. A smaller learning rate (`1e-5') is used to prevent drastic changes to the pre-trained weights. After training for 10 epochs, we evaluate the model and print the test accuracy to observe the improvement achieved through fine-tuning.

\subsection{ConvNeXtXLarge}
The ConvNeXtXLarge model is a modern convolutional neural network (CNN) architecture that builds upon the ResNet structure and introduces improvements inspired by vision transformers. It is pre-trained on ImageNet and performs well on a variety of image classification tasks. In this section, we will use the ConvNeXtXLarge model for transfer learning on the CIFAR-10 dataset, using both Linear Probe and Fine-tuning approaches. Since CIFAR-10 images are smaller than the input size expected by ConvNeXtXLarge, we resize the images to 224x224 pixels \cite{DBLP:journals/corr/abs-2201-03545}.

\paragraph{Linear Probe}
In the Linear Probe approach, we freeze the pre-trained ConvNeXtXLarge model’s convolutional layers and train only the new classification layers on CIFAR-10. This allows the model to use the pre-trained features without modifying them.

\begin{lstlisting}[style=python]
import tensorflow as tf
from tensorflow.keras import layers, models
from tensorflow.keras.applications import ConvNeXtXLarge
from tensorflow.keras.datasets import cifar10
from tensorflow.keras.utils import to_categorical

# Load CIFAR-10 dataset and preprocess the images
(x_train, y_train), (x_test, y_test) = cifar10.load_data()
x_train = tf.image.resize(x_train, (224, 224)) / 255.0
x_test = tf.image.resize(x_test, (224, 224)) / 255.0
y_train = to_categorical(y_train, 10)
y_test = to_categorical(y_test, 10)

# Load pre-trained ConvNeXtXLarge model without the top classification layers
base_model = ConvNeXtXLarge(weights='imagenet', include_top=False, input_shape=(224, 224, 3))

# Freeze all layers in the base model
for layer in base_model.layers:
    layer.trainable = False

# Add classification layers on top of the base model
model = models.Sequential([
    base_model,
    layers.GlobalAveragePooling2D(),
    layers.Dense(256, activation='relu'),
    layers.Dense(10, activation='softmax')
])

# Compile the model
model.compile(optimizer='adam', loss='categorical_crossentropy', metrics=['accuracy'])

# Train the model (Linear Probe)
history = model.fit(x_train, y_train, epochs=10, validation_data=(x_test, y_test))

# Evaluate the model
results = model.evaluate(x_test, y_test)
print(f"ConvNeXtXLarge Linear Probe Test Accuracy: {results[1]}")
\end{lstlisting}

In this code, we load the CIFAR-10 dataset, resize the images to 224x224 pixels, and normalize them. We then load the ConvNeXtXLarge model without its top layers and freeze all the pre-trained layers. A new classification head is added, which includes a global average pooling layer and two dense layers. The model is trained for 10 epochs with only the new layers being updated, and the test accuracy is printed at the end.

\paragraph{Fine-tuning}
In the Fine-tuning approach, we unfreeze a few layers of the ConvNeXtXLarge model and allow them to be updated during training, along with the classification layers. This approach helps adapt the pre-trained features more effectively to the CIFAR-10 dataset.

\begin{lstlisting}[style=python]
import tensorflow as tf
from tensorflow.keras import layers, models
from tensorflow.keras.applications import ConvNeXtXLarge
from tensorflow.keras.datasets import cifar10
from tensorflow.keras.utils import to_categorical

# Load CIFAR-10 dataset and preprocess the images
(x_train, y_train), (x_test, y_test) = cifar10.load_data()
x_train = tf.image.resize(x_train, (224, 224)) / 255.0
x_test = tf.image.resize(x_test, (224, 224)) / 255.0
y_train = to_categorical(y_train, 10)
y_test = to_categorical(y_test, 10)

# Load pre-trained ConvNeXtXLarge model without the top classification layers
base_model = ConvNeXtXLarge(weights='imagenet', include_top=False, input_shape=(224, 224, 3))

# Unfreeze the last 4 layers of the base model for fine-tuning
for layer in base_model.layers[-4:]:
    layer.trainable = True

# Add classification layers on top of the base model
model = models.Sequential([
    base_model,
    layers.GlobalAveragePooling2D(),
    layers.Dense(256, activation='relu'),
    layers.Dense(10, activation='softmax')
])

# Compile the model with a lower learning rate for fine-tuning
model.compile(optimizer=tf.keras.optimizers.Adam(1e-5), loss='categorical_crossentropy', metrics=['accuracy'])

# Fine-tune the model
fine_tune_history = model.fit(x_train, y_train, epochs=10, validation_data=(x_test, y_test))

# Evaluate the model
fine_tune_results = model.evaluate(x_test, y_test)
print(f"ConvNeXtXLarge Fine-tuning Test Accuracy: {fine_tune_results[1]}")
\end{lstlisting}

In this code, we again load the CIFAR-10 dataset and resize the images. We load the ConvNeXtXLarge model pre-trained on ImageNet and unfreeze the last four layers, allowing them to be fine-tuned. The model is compiled with a smaller learning rate to prevent overfitting during the fine-tuning process. The model is trained for 10 epochs, and the test accuracy is printed, showing the benefits of fine-tuning.
\section{Multimodal}

Multimodal deep learning models represent a powerful advancement in machine learning, designed to process and integrate information from multiple modalities, such as text, images, audio, and video. Unlike traditional models that typically focus on one type of data, multimodal deep learning seeks to understand and combine diverse forms of input to improve predictive accuracy and create richer, more holistic models of the world \cite{niu2024large,Peng_2024,peng2024securinglargelanguagemodels}.

{\color{codegreen} In previous sections, we have thoroughly covered TensorFlow’s default pre-trained models, including VGG, ResNet, Inception, and EfficientNet, etc. It is \textbf{not recommended} for users to introduce other pre-trained models into TensorFlow, as this could lead to various unexpected errors within TensorFlow. The introduction of alternative models will be discussed in the chapters on \textbf{PyTorch} and \textbf{HuggingFace}.}

\subsection{CLIP: Connecting text and images (2021)}

OpenAI CLIP (Contrastive Language-Image Pre-training) is a powerful multimodal model designed to understand and link visual and textual information. Developed by OpenAI, CLIP learns by associating images with their corresponding descriptions using a contrastive learning technique. This allows the model to perform a wide range of tasks, even without task-specific training, by understanding visual data in the context of natural language \cite{radford2021learningtransferablevisualmodels}.

CLIP is one of the most widely used models today and also one of the largest models trained in terms of dataset size. It was trained on a dataset of 400 million image-text pairs collected from the internet, providing a diverse and rich source of visual and linguistic data. For training the ResNet-50 (ResNet-50x64) model, OpenAI used 592 V100 GPUs over the course of 18 days, while the Vision Transformer (ViT-B/32) model was trained using 256 V100 GPUs over 12 days. These two architectures—the ResNet, a convolutional neural network, and the ViT, which uses transformers to capture long-range dependencies in images—were selected to explore different approaches in multimodal understanding.

Such a massive dataset, even before training, requires substantial effort just to collect, clean, and process. The complexity of filtering and curating hundreds of millions of image-text pairs is an enormous task in itself—one that is far beyond the reach of most individual researchers or small labs. In addition, the computational resources necessary to train these models, such as hundreds of high-performance GPUs running for weeks, are simply inaccessible to the majority of researchers. Therefore, it's impractical for most people to attempt to train these models themselves. Fortunately, OpenAI has released the pre-trained weights for CLIP, allowing the broader research community to use these models without needing to replicate the immense data collection and computation processes.

Pre-trained models like CLIP exist precisely for this reason. In today’s AI landscape, the scale of data and computation required to train large models has become unattainable for most. Instead, researchers and developers rely on models trained by large organizations or universities. Using these pre-trained weights allows them to bypass the challenges of data collection and training, and still access state-of-the-art performance.

Yann LeCun once used the analogy of a cake to describe the different forms of learning: reinforcement learning (RL) is the cherry on top, supervised learning is the icing, and the cake itself is made of unsupervised learning. This analogy emphasizes the growing importance of unsupervised learning. Supervised learning relies on labeled data, which is expensive and time-consuming to collect. Unsupervised learning, however, can harness vast amounts of unlabeled data, making it far more scalable. By focusing on unsupervised learning, models like CLIP can generalize better across a wide range of tasks, relying on the massive amounts of raw data available in the world.

\paragraph{Key Concepts of CLIP}
Contrastive Learning: CLIP is trained on a large dataset containing images and their associated text descriptions. During training, the model is taught to maximize the similarity between the correct image-text pairs while minimizing the similarity between incorrect ones. This contrastive learning method enables CLIP to learn relationships between images and text in a highly generalizable way.

\paragraph{Zero-shot Learning} One of CLIP's standout features is its ability to perform "zero-shot" classification. After pre-training, CLIP can recognize new image categories without additional task-specific training. For example, you can provide CLIP with labels or descriptions for novel concepts, and it will match the appropriate image to that label based on its learned understanding of language and images.

{\color{codegreen} Users should avoid fine-tuning CLIP’s image or text encoders, as it can degrade performance. This happens because fine-tuning without using \textbf{contrastive learning} can disrupt the balance between CLIP’s joint understanding of both modalities (text and image), which is crucial for its success in zero-shot tasks. Contrastive learning is key to maintaining this synergy by ensuring that the model learns robust associations between images and their textual descriptions.}

\paragraph{Multimodal Understanding} CLIP combines language models with image understanding, making it versatile in tasks like image classification, object detection, and even generating captions for images. It connects what it "sees" (image content) to what it "reads" (text), which opens up new possibilities in the realm of human-AI interaction.

\paragraph{Robust Performance} The model’s training on a large and diverse dataset gives it strong performance across various domains and contexts. It can classify images in many different styles, such as sketches, cartoons, or photographs, as long as it can match them with an appropriate textual description.

\paragraph{Applications} CLIP is being used in areas such as image search, content moderation, and art generation. It is also useful for building interfaces that rely on both visual and textual data, like improving the capabilities of virtual assistants or enabling more intuitive human-computer interaction in creative fields.

\paragraph{How CLIP Works}
CLIP consists of two primary components: an image encoder and a text encoder. The image encoder processes the visual data, typically using a variant of a convolutional neural network (CNN), while the text encoder uses a transformer architecture to process natural language descriptions. Both encoders project their inputs into a shared latent space, where the contrastive learning happens. CLIP then learns to align similar image-text pairs in this shared space.

\paragraph{Code Example}
The following code example uses the CLIP model provided by Hugging Face to calculate image-text similarity scores (Hugging Face, n.d.) \cite{huggingface2024clip}.

\begin{lstlisting}[style=python]
# Importing necessary libraries
import tensorflow as tf  # TensorFlow for deep learning computations
from PIL import Image  # Python Imaging Library (PIL) for image processing
import requests  # For making HTTP requests to download an image
from transformers import AutoProcessor, TFCLIPModel  # Transformers library components for processing and model

# Load the pre-trained CLIP model from Hugging Face (OpenAI CLIP: Vision Transformer (ViT) Base, Patch32)
# CLIP model is used for image-text alignment, comparing how similar an image is to given text descriptions.
model = TFCLIPModel.from_pretrained("openai/clip-vit-base-patch32")

# Load the pre-trained processor from Hugging Face (AutoProcessor for CLIP)
# This processor is responsible for processing both the image and the text inputs for the CLIP model.
processor = AutoProcessor.from_pretrained("openai/clip-vit-base-patch32")

# Specify the URL of an image (from COCO dataset in this case)
url = "http://images.cocodataset.org/val2017/000000039769.jpg"

# Download the image from the URL and open it using PIL's Image class
image = Image.open(requests.get(url, stream=True).raw)

# Process the inputs (both text and image) with the processor
# We pass two text descriptions: "a photo of a cat" and "a photo of a dog"
# The image and text are processed and returned as tensors ready for the model
inputs = processor(
    text=["a photo of a cat", "a photo of a dog"],  # Text input to compare with the image
    images=image,  # Image input
    return_tensors="tf",  # Return TensorFlow tensors (since we're using the TF version of CLIP)
    padding=True  # Apply padding to ensure consistent tensor shapes
)

# Run the inputs through the model to get output
# The model outputs the logits for image-text similarity
outputs = model(**inputs)

# Extract logits (raw scores before applying any activation function) for image-text similarity
# The logits_per_image gives the similarity between the image and the text descriptions.
logits_per_image = outputs.logits_per_image  

# Apply softmax to convert the raw logits into probabilities
# This gives us a probability distribution for the similarity of the image with each text description.
probs = tf.nn.softmax(logits_per_image, axis=1)  # We use softmax to get probabilities over the labels

\end{lstlisting}

\section{Compare and Summarize}

In this section, we provide a comparison and summary of the prominent deep learning models, which are all the pretrained-models in \textbf{tensorflow}, specifically focusing on their architectures, parameter sizes, and use cases. Each model family is designed to tackle different challenges in computer vision tasks, such as image classification, object detection, and feature extraction. Below is an overview of these models.

\subsection{VGG (2014)}
The VGG models, VGG16 and VGG19, are known for their simplicity and depth. Both architectures consist of small 3x3 convolutional filters applied in a sequential manner, followed by fully connected layers. While these models were state-of-the-art in 2014, their major drawback is the large number of parameters, which leads to high computational costs and slow training. Despite this, they serve as a strong baseline for deep learning benchmarks and are still widely used for transfer learning \cite{simonyan2015deepconvolutionalnetworkslargescale}.

\subsection{Inception (2015)}
Inception models introduce the concept of "Inception modules," which allow for multi-scale feature extraction within a single layer. InceptionV3 is a highly efficient model in terms of computation, while InceptionResNet integrates residual connections, improving the model's ability to handle deeper architectures. InceptionResNetV2 further enhances this by improving accuracy while maintaining efficient computation. These models are well-suited for tasks requiring efficiency without compromising accuracy \cite{szegedy2015rethinkinginceptionarchitecturecomputer}.

\subsection{ResNet (2015)}
ResNet models revolutionized deep learning with the introduction of residual connections, solving the vanishing gradient problem in deep networks. ResNet50, ResNet101, and ResNet152 (V1 and V2 variants) vary in depth and complexity, with ResNetV2 showing slight improvements over V1 in terms of optimization. These models are highly effective for deep tasks, such as image classification and object detection, due to their ability to train deeper networks without degradation in performance \cite{he2015deepresiduallearningimage}.

\subsection{MobileNet (2017)}
MobileNet models are optimized for mobile and embedded devices, prioritizing computational efficiency. MobileNetV1 uses depthwise separable convolutions to reduce the number of parameters, while MobileNetV2 introduces inverted residual blocks and linear bottlenecks for better performance. MobileNetV3 (both Large and Small variants) further refines this with additional optimizations from neural architecture search (NAS). These models are ideal for real-time applications on resource-constrained devices \cite{howard2017mobilenetsefficientconvolutionalneural}.

\subsection{Xception (2017)}
Xception extends the idea of depthwise separable convolutions by fully decoupling the spatial convolutions from the channel-wise ones. This results in a more efficient architecture that can achieve higher accuracy with fewer parameters compared to traditional convolutional layers. Xception is highly competitive with Inception models and is often used for high-performance image classification tasks \cite{chollet2016xception}.

\subsection{NASNet (2018)}
NASNet models are designed using Neural Architecture Search, which automates the process of discovering the optimal architecture. NASNetLarge is built for high accuracy, while NASNetMobile focuses on efficiency for mobile devices. These models represent a balance between automation in architecture design and performance, with NASNetMobile offering an excellent trade-off between accuracy and efficiency on smaller devices.

\subsection{DenseNet (2017)}
DenseNet architectures introduce dense connections between layers, ensuring maximum information flow between them. DenseNet121, DenseNet169, and DenseNet201 differ in terms of depth, but all focus on alleviating the vanishing gradient problem and improving feature reuse. These models are computationally efficient and have fewer parameters compared to other architectures of similar depth, making them well-suited for tasks like image segmentation \cite{huang2018denselyconnectedconvolutionalnetworks}.

\subsection{EfficientNet (2019)}
EfficientNet models are built upon a scaling strategy that balances network depth, width, and resolution. EfficientNetV1 offers multiple variants (B0 to B7), providing different levels of accuracy and computational cost. EfficientNetV2 introduces further optimizations for faster training and inference times. These models are designed for achieving high accuracy with fewer parameters and are frequently used in tasks that require a good trade-off between efficiency and performance \cite{tan2021efficientnetv2smallermodelsfaster}.

\subsection{ConvNeXt (2022)}
ConvNeXt is a modernized version of the traditional convolutional neural network (CNN) design, incorporating advancements from Transformer-based models. ConvNeXtTiny, Small, Base, Large, and XLarge offer varying levels of depth and computational complexity. ConvNeXt models are built for high performance in image classification tasks and have shown strong competitive results against Transformer models while maintaining the simplicity of CNNs \cite{DBLP:journals/corr/abs-2201-03545}.

\subsection{OpenAI - CLIP: Connecting text and images (2021)}
OpenAI's CLIP (Contrastive Language–Image Pretraining) model is a transformative approach that bridges the gap between vision and language. CLIP is trained using a massive dataset of images paired with their corresponding textual descriptions, allowing it to understand and generate meaningful associations between visual data and natural language. Unlike traditional models that require fine-tuning for specific tasks, CLIP can perform zero-shot learning—interpreting new tasks without additional training—by leveraging its ability to match text and images directly. This versatility makes CLIP particularly useful for a wide range of applications, from image classification and generation to more complex multi-modal tasks like visual question answering and image captioning. Its capability to align both image and text representations in a shared latent space marks a significant advancement in AI's understanding of complex, real-world data \cite{radford2021learningtransferablevisualmodels}.

\subsection{Summary}
In summary, each model family presents unique strengths and trade-offs. VGG models are easy to understand but computationally expensive. Inception models excel in efficiency and multi-scale feature extraction. ResNet models introduced residual connections, allowing for very deep networks to be trained effectively. MobileNet and NASNet models focus on computational efficiency, especially for mobile applications, while DenseNet improves feature reuse through dense connections. EfficientNet balances performance and computational cost through a compound scaling strategy. Finally, ConvNeXt bridges the gap between CNNs and Transformers, leveraging modern architectural advances to compete with state-of-the-art models.

The choice of model depends on the specific use case: for mobile devices, MobileNet and NASNetMobile are preferable, while for highly accurate image classification tasks, models like EfficientNetB7, DenseNet201, or ConvNeXtXLarge are more suitable. ResNet and InceptionResNet are well-rounded choices for deep learning tasks requiring a balance of depth and performance.

\begin{landscape}
\begin{longtable}{|p{3cm}|p{1.0cm}|p{3cm}|p{1.5cm}|p{4.5cm}|p{4.5cm}|}
    \caption{Comparison of Pre-trained Models in TensorFlow} \\
    \hline
    \textbf{Model} & \textbf{Release Year} & \textbf{Model Size} & \textbf{Speed} & \textbf{Pros} & \textbf{Cons} \\
    \hline
    \endfirsthead
    
    \hline
    \textbf{Model} & \textbf{Release Year} & \textbf{Model Size} & \textbf{Speed} & \textbf{Pros} & \textbf{Cons} \\
    \hline
    \endhead
    
    \hline
    \multicolumn{6}{|r|}{Continued on next page} \\
    \hline
    \endfoot
    
    \hline
    \endlastfoot
    
    VGG16 & 2014 & 138M parameters & Slow & Simple architecture, good for transfer learning & Large model, high computational cost \\
    VGG19 & 2014 & 143M parameters & Slow & Similar to VGG16 but deeper & Same as VGG16, even higher computational cost \\
    \hline
    InceptionV3 & 2015 & 23M parameters & Fast & Efficient multi-scale feature extraction & Complex architecture, hard to implement manually \\
    InceptionResNetV2 & 2016 & 56M parameters & Moderate & Combines benefits of Inception and ResNet & Higher complexity, more parameters \\
    \hline
    ResNet50 & 2015 & 25.6M parameters & Fast & Solves vanishing gradient problem with residuals & Large memory footprint \\
    ResNet101 & 2015 & 44.5M parameters & Moderate & Good performance for deep learning tasks & Computationally expensive \\
    ResNet152 & 2015 & 60.2M parameters & Slow & Excellent for very deep tasks & Very slow, large memory usage \\
    ResNet50V2 & 2016 & 25.6M parameters & Fast & Improved optimization over ResNetV1 & Still large in size \\
    \hline
    MobileNetV1 & 2017 & 4.2M parameters & Very Fast & Extremely efficient for mobile devices & Lower accuracy compared to larger models \\
    MobileNetV2 & 2018 & 3.4M parameters & Very Fast & Improved efficiency, better accuracy than V1 & Still not as accurate as larger models \\
    MobileNetV3Large & 2019 & 5.4M parameters & Very Fast & Optimized through NAS for better performance & Trade-off between accuracy and efficiency \\
    MobileNetV3Small & 2019 & 2.9M parameters & Very Fast & Great for resource-constrained environments & Lower accuracy \\
    \hline
    Xception & 2017 & 22.9M parameters & Moderate & Higher accuracy with fewer parameters than Inception & Complex architecture \\
    \hline
    NASNetLarge & 2018 & 88.9M parameters & Slow & Optimized architecture through NAS for accuracy & Large and slow, not suitable for real-time tasks \\
    NASNetMobile & 2018 & 5.3M parameters & Fast & Optimized for mobile devices with good accuracy & Lower accuracy than larger models \\
    \hline
    DenseNet121 & 2017 & 8M parameters & Moderate & Dense connections improve feature reuse & More memory and computation needed \\
    DenseNet169 & 2017 & 14.3M parameters & Moderate & Balanced trade-off between depth and performance & More complex than simpler architectures \\
    DenseNet201 & 2017 & 20M parameters & Slow & High accuracy, great for deep tasks & High memory consumption \\
    \hline
    EfficientNetB0 & 2019 & 5.3M parameters & Fast & Good trade-off between accuracy and efficiency & Less accurate than larger variants \\
    EfficientNetB7 & 2019 & 66M parameters & Slow & State-of-the-art accuracy & Very large and slow \\
    EfficientNetV2S & 2021 & 24M parameters & Fast & Faster training and inference & Requires more computation \\
    EfficientNetV2L & 2021 & 121M parameters & Slow & Very high accuracy & Very large model \\
    \hline
    ConvNeXtTiny & 2022 & 28.6M parameters & Fast & Modernized CNN design with high performance & Still larger than MobileNets \\
    ConvNeXtLarge & 2022 & 198M parameters & Slow & Competitive with Transformer-based models & Large and computationally heavy \\

    OpenAI Clip & 2021 & 428M parameters & Moderate & Multimodal learning for both image and text, good at zero-shot learning and transfer learning across different tasks. & Large model size and high computational cost, requires significant resources for training and inference. \\
    \hline
\end{longtable}
\end{landscape}

\subsection{Mind map of this Chapter}

In this chapter, we generated a mind map using the Python script. The mind map illustrates the main branches of TensorFlow pretrained models and their evolution over different versions, ranging from VGG to the latest ConvNeXt networks. The models are categorized by year and visualized accordingly, with each model and its submodels marked using distinct colors.

We employed the `NetworkX' library to construct the graph structure and used `Matplotlib' for visualization. The central node represents "TensorFlow Pretrained Models", from which branches extend to various models. For instance, the VGG series appeared in 2013, the Inception series in 2014, and the ResNet series in 2015. The nodes and their respective subnodes are colored based on their year, using a `jet' colormap for differentiation.

Additionally, we used a radial layout to organize the graph, positioning different hierarchical levels of models as concentric layers radiating from the center. To enhance readability, we adjusted the font color and formatting of node labels, using white text for dark background nodes and black text for light background nodes.

After running the Python script, the resulting mind map was saved as `TensorFlow Pretrained Models.pdf':

\begin{figure}[H]
    \centering
    \includegraphics[width=1.0\textwidth]{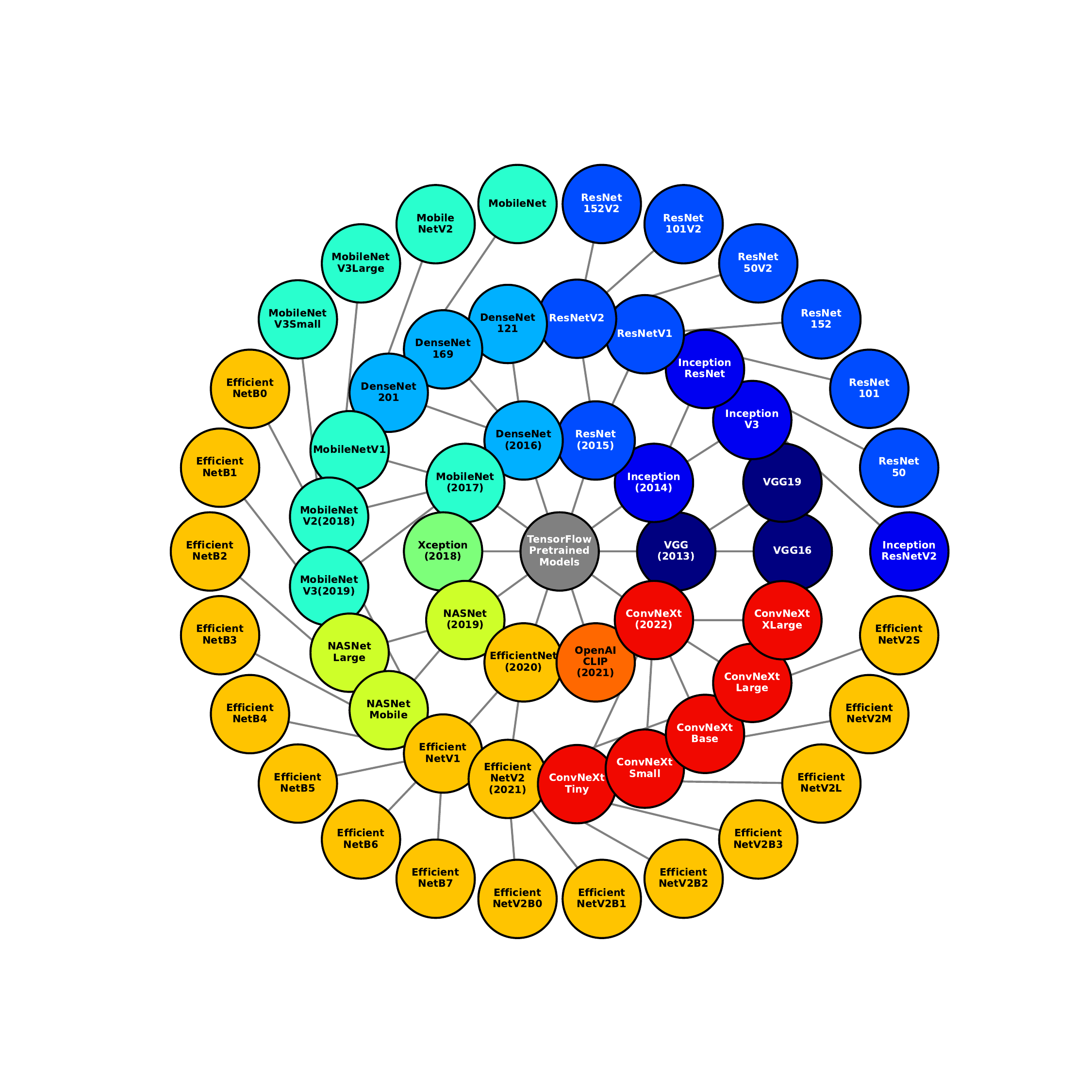}
    \caption{Mind map of TensorFlow Pretrained Models}
    \label{fig:mindmap}
\end{figure}

\subsection{Python Script for Mind Map Generation}

The following code was used to generate the mind map:

\begin{lstlisting}[style=python]
import numpy as np
import networkx as nx

import matplotlib.cm as cm
import matplotlib.pyplot as plt


# Create a Graph
G = nx.Graph()

# Central node
G.add_node("TensorFlow Pretrained Models")

# Adding branches based on the provided structure with years
models = {
    "VGG (2013)": ["VGG16", "VGG19"],
    "Inception (2014)": {
        "Inception V3": [], 
        "Inception ResNet": ["Inception ResNetV2"]
    },
    "ResNet (2015)": {
        "ResNetV1": ["ResNet 50", "ResNet 101", "ResNet 152"],
        "ResNetV2": ["ResNet 50V2", "ResNet 101V2", "ResNet 152V2"]
    },
    "DenseNet (2016)": ["DenseNet 121", "DenseNet 169", "DenseNet 201"],
    "MobileNet (2017)": {
        "MobileNetV1": ["MobileNet"],
        "MobileNet V2(2018)": ["Mobile NetV2"],
        "MobileNet V3(2019)": ["MobileNet V3Large", "MobileNet V3Small"]
    },
    "Xception (2018)": [],
    "NASNet (2019)": ["NASNet Large", "NASNet Mobile"],
    "EfficientNet (2020)": {
        "Efficient NetV1": [
            "Efficient NetB0", "Efficient NetB1", "Efficient NetB2", 
            "Efficient NetB3", "Efficient NetB4", "Efficient NetB5", 
            "Efficient NetB6", "Efficient NetB7"
        ],
        "Efficient NetV2 (2021)": [
            "Efficient NetV2B0", "Efficient NetV2B1", "Efficient NetV2B2",
            "Efficient NetV2B3", "Efficient NetV2L", "Efficient NetV2M", 
            "Efficient NetV2S"
        ]
    },
    "OpenAI CLIP (2021)": [],
    "ConvNeXt (2022)": [
        "ConvNeXt Tiny", "ConvNeXt Small", "ConvNeXt Base", 
        "ConvNeXt Large", "ConvNeXt XLarge"
    ],
}

# Adding the nodes to the tree and year-based categorization for colors
year_to_models = {
    2013: ["VGG (2013)"],
    2014: ["Inception (2014)"],
    2015: ["ResNet (2015)"],
    2016: ["DenseNet (2016)"],
    2017: ["MobileNet (2017)"],
    2018: ["Xception (2018)"],
    2019: ["NASNet (2019)"],
    2020: ["EfficientNet (2020)"],
    2021: ["OpenAI CLIP (2021)"],
    2022: ["ConvNeXt (2022)"],
}

# Map years to colors using a colormap (jet)
cmap = cm.get_cmap('jet')
years = sorted(year_to_models.keys())
year_colors = {year: cmap(i / len(years)) for i, year in enumerate(years)}

# Assign edges and nodes for the models
node_colors = {}
for model, submodels in models.items():
    # Get the year color for the current model
    for year, model_list in year_to_models.items():
        if model in model_list:
            color = year_colors[year]
            break
    node_colors[model] = color
    G.add_edge("TensorFlow Pretrained Models", model)
    if isinstance(submodels, list):
        for submodel in submodels:
            G.add_edge(model, submodel)
            node_colors[submodel] = color  # Same color for submodels
    elif isinstance(submodels, dict):
        for submodel, subsubmodels in submodels.items():
            G.add_edge(model, submodel)
            node_colors[submodel] = color  # Same color for submodels
            for subsubmodel in subsubmodels:
                G.add_edge(submodel, subsubmodel)
                node_colors[subsubmodel] = color  # Same color for subsubmodels

# Assign a neutral color for the central node
node_colors["TensorFlow Pretrained Models"] = (0.5, 0.5, 0.5, 1)

# Function to get radial layout where nodes are positioned radially based on layers from the root
def radial_layout(G, center_node, scale):
    """
    Create a radial layout for a graph where each layer of nodes is placed farther from the center node.
    The scale parameter controls how much space is added between layers.
    """
    layers = {}
    visited = {center_node: 0}
    queue = [center_node]

    # Perform BFS to determine layers from the center node
    while queue:
        node = queue.pop(0)
        layer = visited[node]
        if layer not in layers:
            layers[layer] = []
        layers[layer].append(node)

        for neighbor in G.neighbors(node):
            if neighbor not in visited:
                visited[neighbor] = layer + 1
                queue.append(neighbor)

    # Assign positions to each node
    pos = {}
    max_nodes_in_layer = max(len(nodes) for nodes in layers.values())
    for layer, nodes in layers.items():
        radius = layer * scale  # Increase radius with layer number
        angle_step = 2 * np.pi / len(nodes)
        for i, node in enumerate(nodes):
            angle = i * angle_step
            pos[node] = (radius * np.cos(angle), radius * np.sin(angle))

    return pos

# Apply radial layout to the graph
pos = radial_layout(G, "TensorFlow Pretrained Models", scale=2.5)

# Adjust label colors and text based on background color
def adjust_label_color_and_format(node_label, color):
    """
    Adjust the text color for labels and format long text with line breaks.
    
    Dark node background -> white text, Light node background -> black text
    Also format labels to add line breaks if text is too long.
    """
    # Determine if the label color should be white or black based on node color brightness
    r, g, b, _ = color
    brightness = (r * 0.299 + g * 0.587 + b * 0.114)  # Perceived brightness
    text_color = 'white' if brightness < 0.5 else 'black'
    
    # Break label text into multiple lines if it's too long
    node_label = "\n".join(node_label.split())
        
    return node_label, text_color

# Prepare node labels with adjusted text color and formatting
labels_with_colors = {}
for node in G.nodes():
    formatted_label, text_color = adjust_label_color_and_format(node, node_colors[node])
    labels_with_colors[node] = (formatted_label, text_color)

# Extract node colors in the correct order for plotting
node_color_list = [node_colors[node] for node in G.nodes()]

# Draw the graph with radial layout and adjusted label formatting
plt.figure(figsize=(15, 15))

# Draw the nodes with outlines (using edgecolors) and node size
nx.draw_networkx_nodes(G, pos, node_color=node_color_list, node_size=6000, edgecolors='black', linewidths=2)

# Draw the labels with proper formatting and color
for node, (label, color) in labels_with_colors.items():
    x, y = pos[node]
    plt.text(x, y, label, fontsize=10, fontweight='bold', horizontalalignment='center', verticalalignment='center', color=color)

# Draw curved edges for the graph
nx.draw_networkx_edges(G, pos, connectionstyle="arc3,rad=0.2", edge_color="gray", width=2)

# remove frame
ax = plt.gca()
ax.spines['top'].set_visible(False)
ax.spines['right'].set_visible(False)
ax.spines['left'].set_visible(False)
ax.spines['bottom'].set_visible(False)
# remove axis
ax.get_xaxis().set_visible(False)
ax.get_yaxis().set_visible(False)

# Show the plot
# plt.show()
plt.savefig('TensorFlow Pretrained Models.pdf')
plt.close()
\end{lstlisting}

This script uses `NetworkX' and `Matplotlib' to build and render the mind map, employing a radial layout to arrange the nodes, reflecting the hierarchical structure radiating from the central node of TensorFlow Pretrained Models.
  % Example from the first code

\bibliographystyle{ieeetr}
\bibliography{sample}

\end{document}